\pdfoutput=1

\documentclass[11pt]{article}

\usepackage[final]{acl}

\usepackage{times}
\usepackage{latexsym}

\usepackage[T1]{fontenc}

\usepackage{amsmath,amsfonts,amssymb}
\usepackage{amsthm}
\usepackage{graphicx}
\usepackage{subcaption}
\usepackage{bm}
\usepackage{cases}
\usepackage{multirow}
\usepackage{booktabs}
\usepackage{mathtools}
\usepackage{color}
\usepackage{stmaryrd}
\usepackage{pifont}
\usepackage{changes}
\usepackage{enumitem}
\usepackage{siunitx}

\usepackage[utf8]{inputenc}

\usepackage{microtype}

\usepackage{inconsolata}

\usepackage{graphicx}
\usepackage{placeins}

\title{From Signal Degradation to Computation Collapse: Uncovering the Two Failure Modes of LLM Quantization}

\author{
  Chenxi Zhou\textsuperscript{1,2},
  Pengfei Cao\textsuperscript{2,3}\thanks{Corresponding authors},
  Jiang Li\textsuperscript{4},
  Bohan Yu\textsuperscript{1,2},
  Jinyu Ye\textsuperscript{2},
  Jun Zhao\textsuperscript{2,3},
  Kang Liu\textsuperscript{2,3}\footnotemark[1]
  \\
  \\
  \textsuperscript{1}School of Advanced Interdisciplinary Sciences, University of Chinese Academy of Sciences \\
  \textsuperscript{2}The Key Laboratory of Cognition and Decision Intelligence for Complex Systems, \\ Institute of Automation, Chinese Academy of Sciences \\
  \textsuperscript{3}School of Artificial Intelligence, University of Chinese Academy of Sciences \\
  \textsuperscript{4}College of Computer Science, Inner Mongolia University \\
  \texttt{zhouchenxi2025@ia.ac.cn, \{pengfei.cao, jzhao, kliu\}@nlpr.ia.ac.cn} \\[1.5em]
}

\begin{document}
\maketitle
\begin{abstract}

Post-Training Quantization (PTQ) is critical for the efficient deployment of Large Language Models (LLMs). While 4-bit quantization is widely regarded as an optimal trade-off, reducing the precision to 2-bit usually triggers a catastrophic ``performance cliff.'' It remains unclear whether the underlying mechanisms differ fundamentally. Consequently, we conduct a systematic mechanistic analysis, revealing two qualitatively distinct failure modes: Signal Degradation, where the computational patterns remain intact but information precision is impaired by cumulative error; and Computation Collapse, where key components fail to function, preventing correct information processing and destroying the signal in the early layers. Guided by this diagnosis, we conduct mechanism-aware interventions, demonstrating that targeted, training-free repair can mitigate Signal Degradation, but remains ineffective for Computation Collapse. Our findings provide a systematic diagnostic framework for PTQ failures and suggest that addressing Computation Collapse requires structural reconstruction rather than mere compensation.


\end{abstract}

\section{Introduction}
\label{sec:1_intro}

Post-Training Quantization (PTQ) has emerged as a crucial technique for efficient Large Language Model (LLM) deployment. In practice, 4-bit quantization is often regarded as an optimal trade-off~\cite{jin_acl_2024}, achieving significant compression with acceptable performance loss. However, reducing the precision to 2-bit with common methods (e.g., GPTQ~\cite{frantar_iclr_2023}) usually triggers a catastrophic ``performance cliff,'' particularly in tasks requiring precise factual knowledge. Since factual recall forms the foundation of LLM capabilities, this collapse signals a fundamental breakdown that requires deep investigation.

Existing research on PTQ spans three primary directions. 
\textbf{The first focuses on macroscopic evaluation}, measuring how much performance drops on diverse downstream tasks~\cite{li_icml_2024, jin_acl_2024, liu_colm_2025}. 
\textbf{The second direction pursues algorithmic refinement}, employing numerical optimization strategies such as outlier suppression~\cite{lin_mlsys_2024} or rotation matrices~\cite{tseng_quip_2024} to reduce errors. 
However, these two directions share a common limitation. They primarily focus on quantifying the performance degradation or minimizing numerical error, but overlook why the model's internal mechanism fails. They treat the quantization damage as a numerical issue rather than investigating the disruption of knowledge storage and recall.

\textbf{The third stream involves preliminary mechanistic exploration}. Common approaches identify critical modules by analyzing layer or component sensitivity~\cite{namburi_nips_2023, zhang_towards_2025, xiao_exploring_2025, dumitru_acl_2025}, while deeper studies attribute failures to the ``RMSNorm Reversal'' effect~\cite{chang_emnlp_2025}. However, these insights remain fragmented, lacking a systematic mechanistic interpretation of the failure modes.
Despite these efforts, we still cannot explain why the ``performance cliff'' exists: \emph{Is the catastrophic failure under common 2-bit merely a quantitative aggravation of 4-bit degradation, or does it mark a qualitative shift to a fundamentally distinct mechanism?}

To answer this, we conduct an in-depth mechanistic analysis. We first trace the layer-wise information flow and causal pathways to investigate whether the knowledge signal exists and propagates correctly. Based on these observations, we reveal two qualitatively distinct PTQ failures. Using standard PTQ settings as representative cases, we propose the \textbf{Two Failure Modes Hypothesis}:
\begin{itemize}[leftmargin=*]
    \setlength\itemsep{0em}
    \item \textbf{Failure Mode I: Signal Degradation.} The model's computational patterns remain largely intact. Quantization error acts as cumulative noise that impairs information precision.
    \item \textbf{Failure Mode II: Computation Collapse.} The quantization error is severe enough to fundamentally damage the functionality of key components. Information cannot be processed correctly and is completely destroyed in the early layers.
\end{itemize}

We validate this hypothesis through a systematic analysis. We examine the functionality of critical components and analyze the internal structure of the representation space. This analysis confirms that Signal Degradation involves functional but impaired components, whereas Computation Collapse stems from a fundamental structural breakdown.

Finally, guided by the diagnosis, we design targeted intervention experiments. We demonstrate that Signal Degradation can be repaired by targeted, training-free strategies. In contrast, Computation Collapse is systemic, where even advanced low-rank compensation remains ineffective, necessitating structural reconstruction (e.g., fine-tuning).


Overall, the main contributions of this work can be summarized as follows:
\begin{itemize} [topsep=3pt, itemsep=3pt, parsep=1pt]
    \item We propose a systematic interpretability analysis framework, providing a general approach to diagnose performance decline under quantization.
    \item We identify two distinct failure modes, Signal Degradation and Computation Collapse, demonstrating that they differ qualitatively rather than merely in severity.
    \item We clarify the optimization strategies for different failure modes, suggesting that while degradation benefits from targeted repair, collapse requires structural reconstruction rather than mere compensation.
\end{itemize}


\section{Related Work}
\label{sec:related_work}

\subsection{Post-Training Quantization}
Post-Training Quantization (PTQ) compresses LLMs efficiently, but the primary challenge lies in handling activation outliers. To mitigate this, methods have evolved from simple rounding to sophisticated numerical transformations. Early weight-only methods like GPTQ~\cite{frantar_iclr_2023} minimize reconstruction error using Hessian information. Techniques like AWQ~\cite{lin_mlsys_2024} and SmoothQuant~\cite{xiao_icml_2023} perform channel-wise scaling to suppress outliers, while recent approaches such as QuIP\#~\cite{tseng_quip_2024} and SpinQuant~\cite{liu_iclr_2025} employ rotation matrices to flatten activation distributions. 

Despite their success in reducing statistical errors like MSE, these methods remain limited to a numerical perspective. By focusing strictly on aligning the output distribution with the full-precision baseline, they overlook internal behaviors and fail to explain how the underlying computational mechanisms change under quantization.

\subsection{Mechanistic Analysis of Quantization}
Mechanistic interpretability offers tools to reverse-engineer model behaviors, such as decoding hidden states via Logit Lens~\cite{nostalgebraist_interpreting_2020} or locating knowledge via Causal Tracing~\cite{meng_nips_2022}. However, the application of these powerful diagnostic tools to investigate the internal mechanics of quantized models remains preliminary.

Prior work in quantization analysis has largely focused on component sensitivity, identifying fragile layers or modules based on Hessian spectra or weight magnitudes~\cite{zhang_towards_2025, dong_hawq-v2_2020}. More recently, researchers have extended mechanistic analysis to specific model capabilities, such as analyzing the compromise of refusal mechanisms~\cite{chhabra_towards_2025}, shifts in truthfulness~\cite{fu_emnlp_2025}, or the unintended recovery of unlearned knowledge~\cite{zhang_iclr_2025}. However, these studies remain fragmented, focusing on isolated tasks or behaviors. Our work aims to provide a systematic mechanistic explanation for quantization failures.


\section{Two Failure Modes Hypothesis}
\label{sec:3_preliminary_experiment}

\subsection{Experimental Setup}

\noindent\textbf{Models and Quantization.} We conduct our primary analysis on Llama-3.1-8B~\cite{grattafiori_llama_2024}. To ensure generalizability, we validate findings on Qwen3-8B~\cite{yang_qwen3_2025}, Mistral-7B-Instruct-v0.3~\cite{jiang_mistral_2023}, and Gemma-2-9B-it~\cite{team_gemma_2024}. We select GPTQ~\cite{frantar_iclr_2023} as the primary baseline as it is the most widely adopted weight-only PTQ method. We contrast 4-bit (the PTQ sweet-spot) and 2-bit (typically unusable) to investigate their fundamentally distinct degradation behaviors, providing 8-bit and 3-bit results for context. Algorithmic generalizability is further validated using AWQ in Appendix~\ref{sec:app_awq}.

\noindent\textbf{Datasets and Task.} We evaluate factual knowledge recall using Pararel~\cite{elazar_measuring_2021} (39 relation types). It is deliberately selected because factual recall is a foundational capability, and its strict \texttt{<subject>-<relation>-<target>} structure provides fixed token positions, facilitating precise mechanistic diagnosis. Relations are mapped to standardized templates for next-token prediction (Appendix~\ref{sec:app_prompts}). Generalizability to broader tasks (MMLU, GSM8K) is verified in Appendix~\ref{sec:app_general_tasks}.

\noindent\textbf{Analysis Subsets.} To specifically investigate quantization-induced failures, we partition the dataset for each model based on FP16 and 4-bit performance into two core subsets: the Robust Subset (\texttt{fp\_and\_4bit\_correct}) and the Failure Subset (\texttt{fp\_correct\_4bit\_wrong}). We do not partition for 2-bit models as they universally fail. All subsequent mechanistic comparisons are performed on these subsets.


\subsection{Phenomenological Evidence}
\label{sec:3.2_phenomenological}

\begin{figure}[t]
  \centering
  \includegraphics[width=0.98\columnwidth]{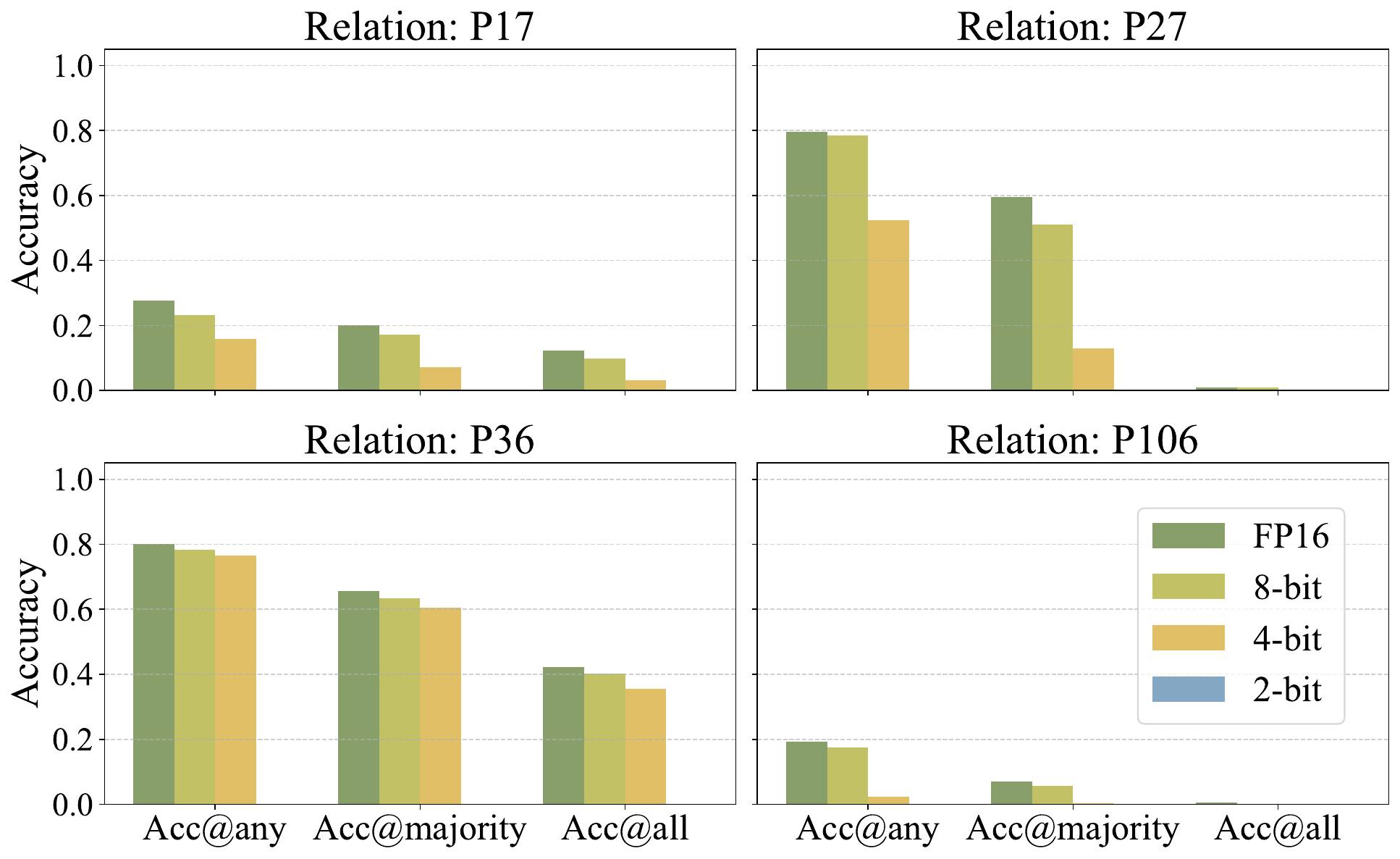}
  \caption{Multi-prompt factual recall accuracy of Llama3.1-8B under different quantization levels on four Pararel relations. We report Accuracy@any ($\ge$1 correct), @majority ($>$50\%), and @all (100\%).}
  \label{fig:robust_eval_4rels}
\end{figure}

\noindent \textbf{Performance Cliff.} 
We conduct a multi-prompt robustness evaluation (see Appendix~\ref{sec:app_prompts}) on the factual recall task. Figure~\ref{fig:robust_eval_4rels} illustrates a pronounced ``performance cliff''. The degradation from FP16 to 4-bit is gradual, maintaining usability. Conversely, the transition to 2-bit triggers a catastrophic collapse where accuracy plummets to zero. This sharp discontinuity suggests that 2-bit quantization represents a distinct failure state rather than a mere lower-precision version of 4-bit.

\begin{figure}[t]
  \centering
  \includegraphics[width=0.98\columnwidth]{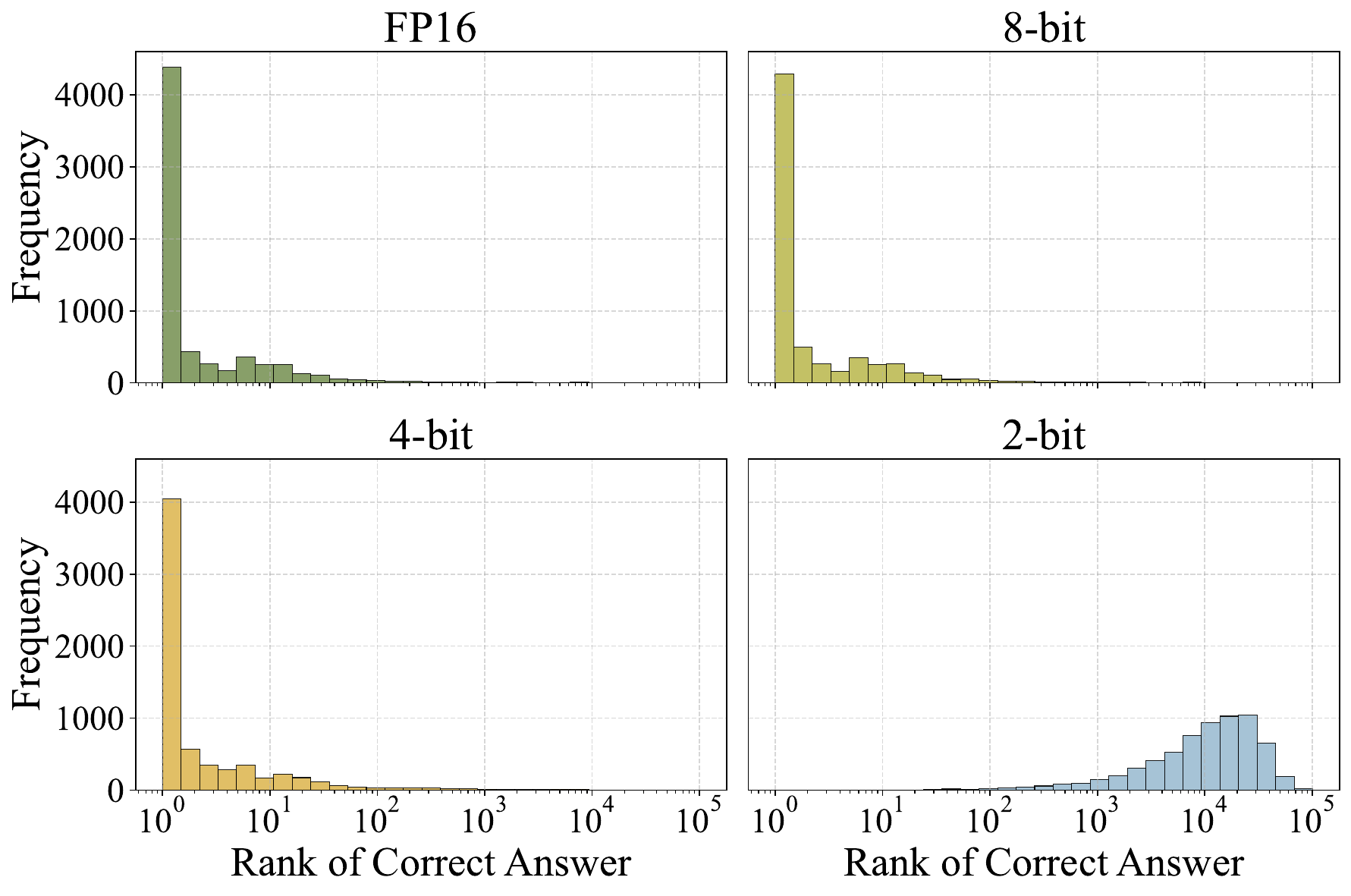}
  \caption{Distribution of the rank of the correct answer  for Llama3.1-8B under different quantization levels on four Pararel relations (P17, P27, P36, P106).}
  \label{fig:Llama3_P36_rank_distribution}
\end{figure}

\noindent \textbf{Rank Drop vs. Collapse.} 
To analyze the nature of these errors, we examine the rank of the correct answer in the final output distribution (Figure~\ref{fig:Llama3_P36_rank_distribution}). 4-bit primarily leads to an ``answer rank drop'', where the correct answer shifts downward but typically remains within the top tier (e.g., Top-5). This indicates that the model retains the correct information despite reduced confidence.
In contrast, 2-bit results in an ``answer rank collapse''. The rank falls to thousands, almost random guessing. Qualitatively, 2-bit models collapse into generating high-frequency stop words (e.g., ``the'', ``.''), reflecting a complete failure in knowledge recall.


\subsection{Layer-wise Knowledge Probing}
\label{sec:3.3_layer_probing}

\begin{figure}[t]
    \centering
    \includegraphics[width=\linewidth]{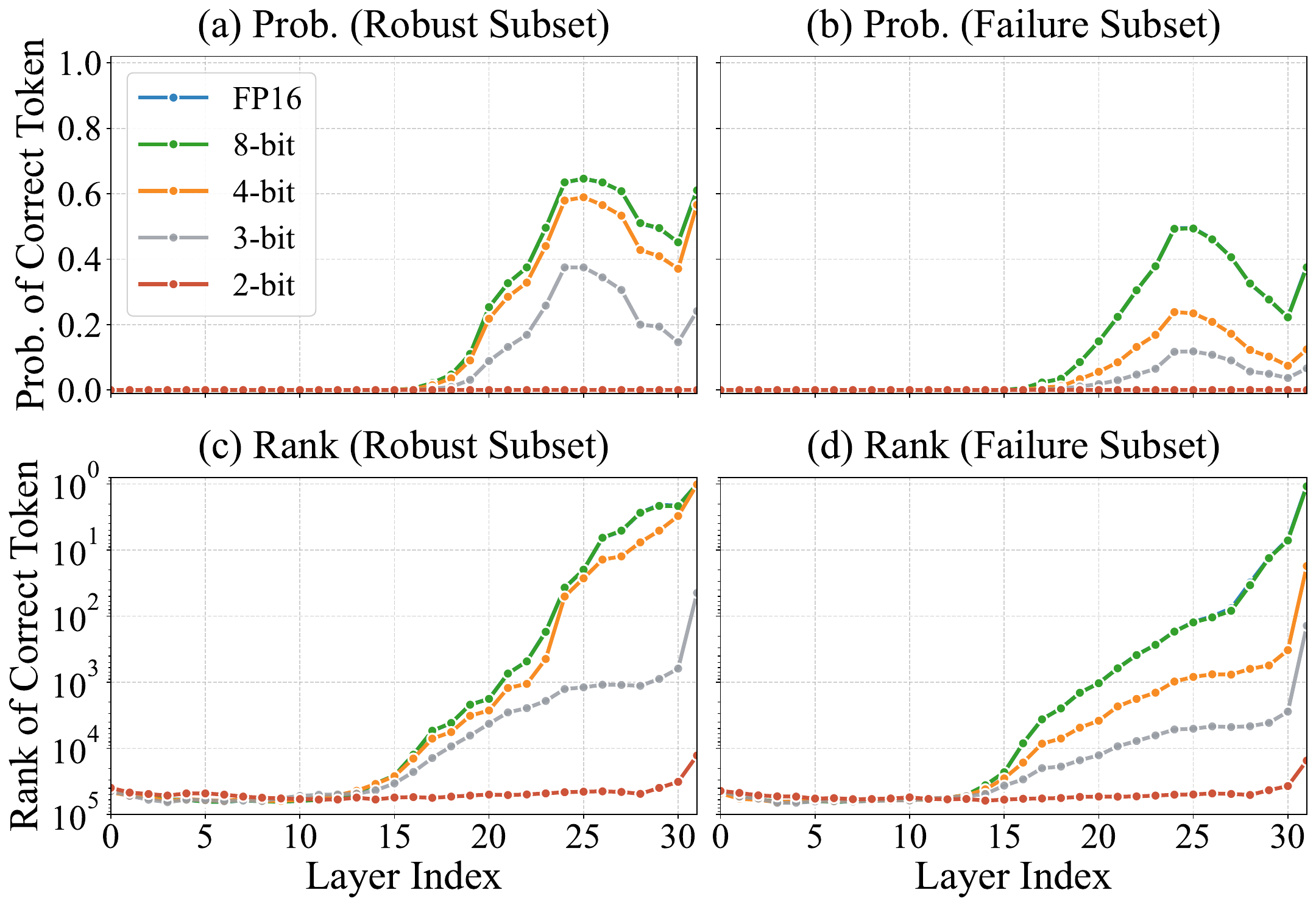}
    \caption{Layer-wise change of probability and rank. FP16 nearly overlaps with 8-bit.}
    \label{fig:probing_prob_and_rank_layer_by_layer}
\end{figure}

To investigate the internal status underlying these macroscopic differences, we examine whether a decodable knowledge signal exists within the intermediate states. We employ the logit lens~\cite{nostalgebraist_interpreting_2020} to project the hidden state $h^{(l)}$ at layer $l$ directly into the vocabulary space via the unembedding matrix $W_U$.
Figure~\ref{fig:probing_prob_and_rank_layer_by_layer} traces the layer-wise change of the correct token's probability and rank, revealing distinct dynamics.

\noindent\textbf{Signal Absence.} 
The 2-bit models exhibit a consistent failure to form an effective knowledge signal.
As shown in the red curves in Figure~\ref{fig:probing_prob_and_rank_layer_by_layer}, the probability of the correct answer remains near zero throughout all layers, and its rank stays extremely low (in the tens of thousands). This indicates that the knowledge signal is never successfully generated during the computation process.

\noindent\textbf{Signal Degradation.} 
In contrast, 4-bit models demonstrate an observable knowledge signal. In the Robust Subset (Fig.~\ref{fig:probing_prob_and_rank_layer_by_layer}a, c), the signal closely tracks the FP16 baseline.
Even in the Failure Subset (Fig.~\ref{fig:probing_prob_and_rank_layer_by_layer}b, d) where the model ultimately fails, the signal still emerges in mid-to-late layers but with reduced intensity.
The probability curve shows lower confidence, and the rank improves more slowly than in FP16. This characterizes 4-bit failure as signal degradation, where the correct signal is present but ultimately overtaken by the noise, unlike the complete absence seen in 2-bit models.


\subsection{Causal Analysis of Information Flow}
\label{sec:3.4_causal_analysis}

While Section~\ref{sec:3.3_layer_probing} analyzes the existence of knowledge signals, it remains unclear whether the causal mechanism for processing them is intact. To distinguish whether the information flow is merely impaired or fundamentally broken, we employ causal activation patching~\cite{heimersheim_how_2024} to assess the integrity of the information pathway.

\paragraph{(1) Cross-Model Repair (Sufficiency).} 
We adapt Causal Tracing~\cite{meng_nips_2022} to test signal sufficiency. We replace the residual stream state $h_{Q}^{(l,t)}$ (i.e., the layer output at layer $l$, token position $t$) in the quantized model with the corresponding ``clean'' activation $h_{FP}^{(l,t)}$ from the FP16 model. If this injection restores the correct prediction, it proves that the injected signal is sufficient to restore the output and the downstream pathway remains functional.

\paragraph{(2) Zeroing Ablation (Necessity).} 
We perform zeroing ablation to test node necessity~\cite{heimersheim_how_2024}. We set activations at specific positions $h^{(l,t)}$ to zero to identify critical nodes. A sharp drop in the probability of the correct answer indicates that the ablated state is necessary for the computation.

\begin{figure}[t]
    \centering
    \begin{subfigure}{\linewidth}
        \includegraphics[width=\linewidth]{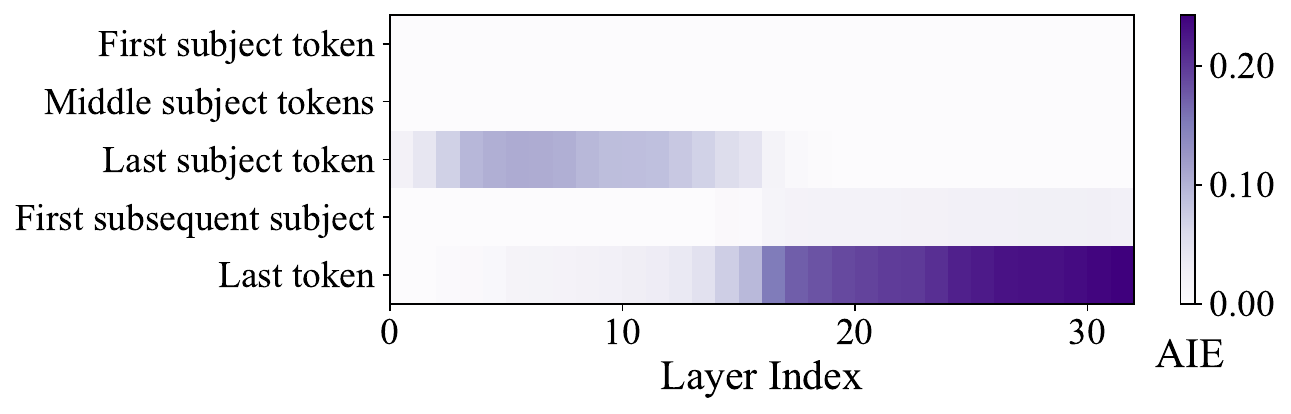}
        \subcaption{4-bit Model Repaired with FP Activations}
        \label{fig:repair_4bit}
    \end{subfigure}
    \vspace{1mm}
    \begin{subfigure}{\linewidth}
        \includegraphics[width=\linewidth]{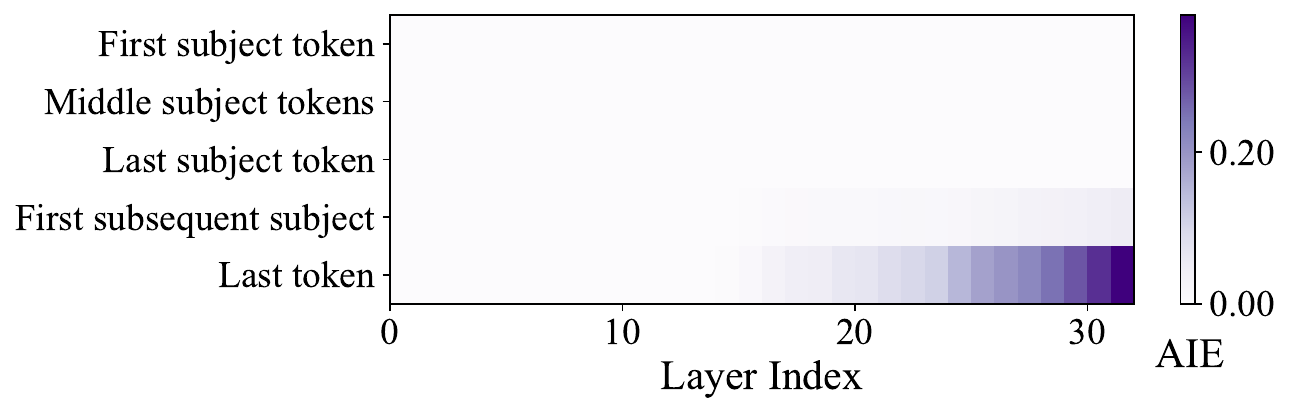}
        \subcaption{2-bit Model Repaired with FP Activations}
        \label{fig:repair_2bit}
    \end{subfigure}
    \caption{Cross-model activation repair on the Failure Subset. The heatmap values represent the Average Indirect Effect (AIE), defined as the increase in the correct token's prediction probability.}
    \label{fig:activation_repair}
\end{figure}


\paragraph{Repair Results (Sufficiency).} 
Figure~\ref{fig:activation_repair} displays the impact of patching clean signals. The 4-bit model (Fig.~\ref{fig:repair_4bit}) shows clear hotspots at the last subject token in early layers. This position is critical for accessing factual knowledge~\cite{meng_nips_2022}. Injecting clean signals here significantly restores prediction performance, proving the connection to the final output is intact. In contrast, the 2-bit model (Fig.~\ref{fig:repair_2bit}) is unresponsive to subject patching. This implies that the computational pathway is broken. The layers fail to pass the information forward, even given correct inputs.

\begin{figure*}[!t]
    \centering
    \begin{subfigure}[t]{0.365\textwidth} 
        \includegraphics[width=\linewidth]{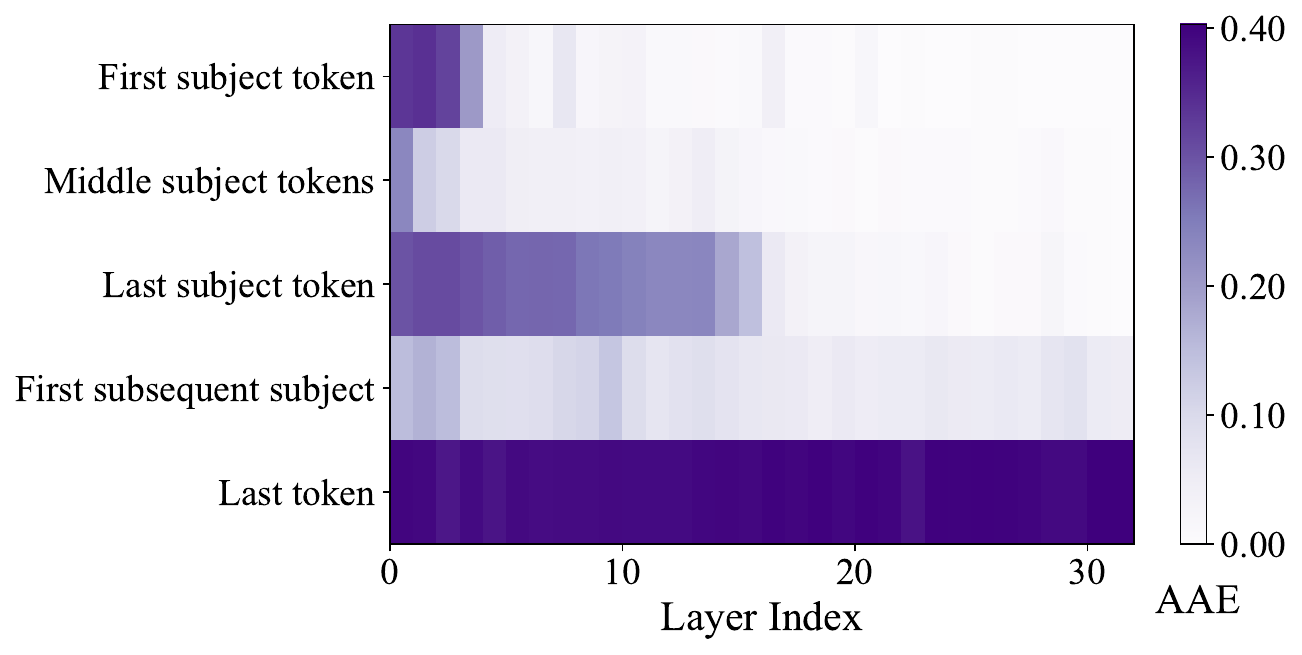} 
        \subcaption{FP Model}
        \label{fig:ablation_fp_wide}
    \end{subfigure}
    \hfill 
    \begin{subfigure}[t]{0.31\textwidth} 
        \includegraphics[width=\linewidth]{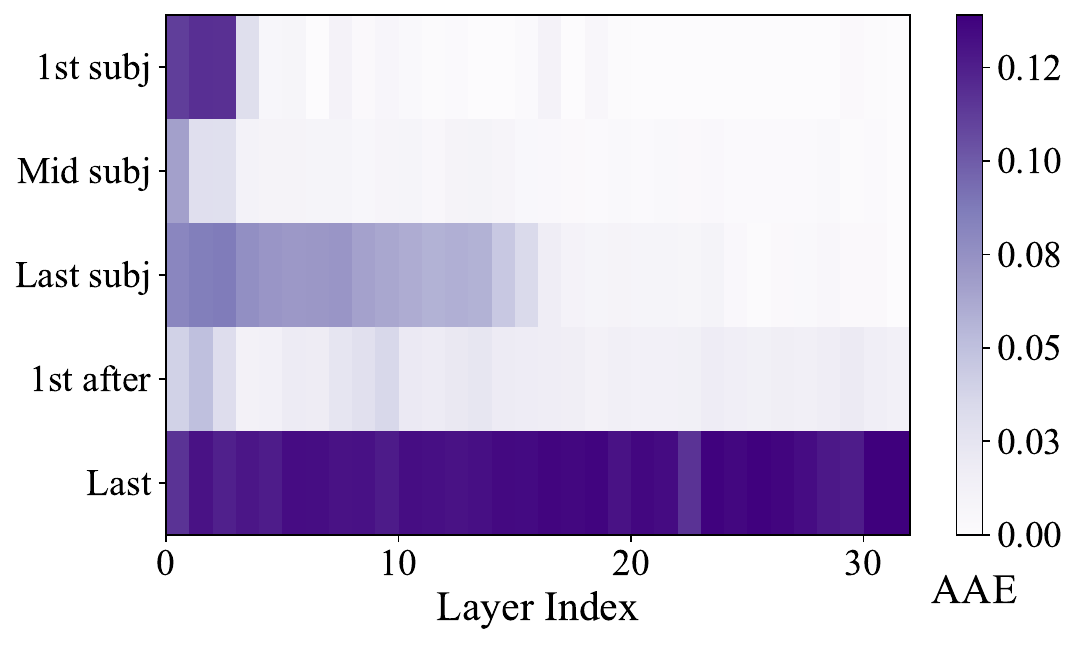} 
        \subcaption{4-bit Model}
        \label{fig:ablation_4bit_wide}
    \end{subfigure}
    \hfill 
    \begin{subfigure}[t]{0.31\textwidth}
        \includegraphics[width=\linewidth]{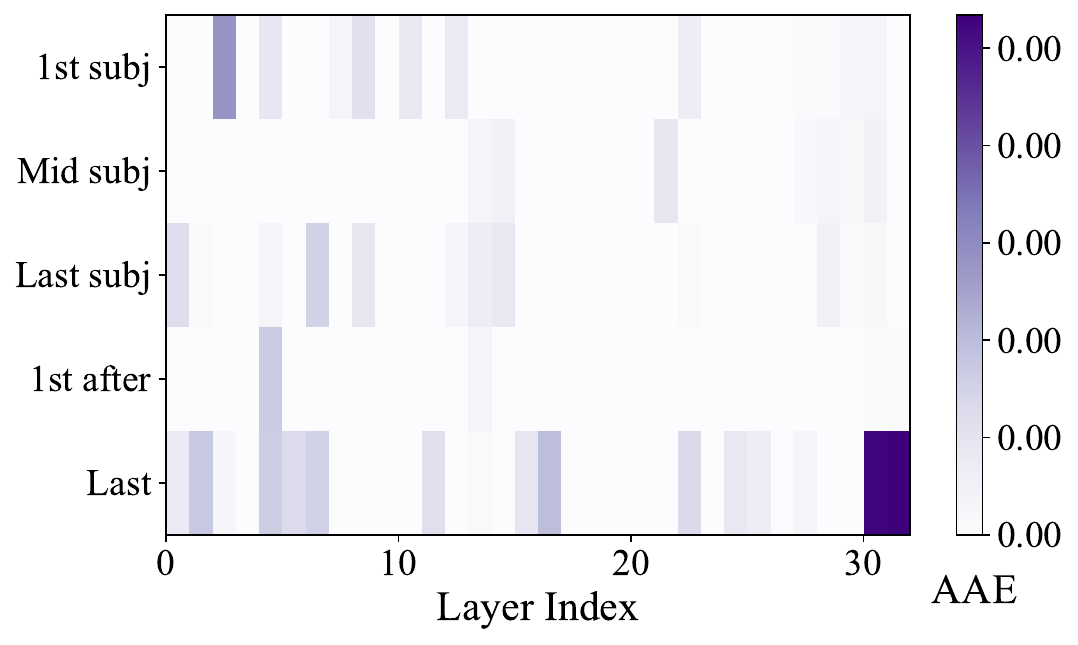}
        \subcaption{2-bit Model}
        \label{fig:ablation_2bit_wide}
    \end{subfigure}
    \caption{Zeroing ablation analysis on the Failure Subset. The heatmap values represent the Average Ablation Effect (AAE), defined as the decrease in the correct token's prediction probability.}
    \label{fig:zero_ablation_wide}
\end{figure*}

\paragraph{Ablation Results (Necessity).}
Figure~\ref{fig:zero_ablation_wide} identifies the critical causal dependencies. The 4-bit model (Fig.~\ref{fig:ablation_4bit_wide}) closely mirrors the FP16 baseline (Fig.~\ref{fig:ablation_fp_wide}), relying on the same last subject token and layers even when the final prediction fails. The reduced intensity suggests that these states are less precise but still functionally necessary. Conversely, the 2-bit model (Fig.~\ref{fig:ablation_2bit_wide}) exhibits a diffuse and unstructured pattern. It loses the concentrated critical nodes seen in FP16. This absence of identifiable dependencies indicates a breakdown of the information processing. 


\paragraph{Hypothesis Formulation.}
Combining the macroscopic (Sec.~\ref{sec:3.2_phenomenological}), layer-wise (Sec.~\ref{sec:3.3_layer_probing}), and causal (Sec.~\ref{sec:3.4_causal_analysis}) evidence, we formulate the two failure modes hypothesis:
\begin{itemize}[leftmargin=*]
    \setlength\itemsep{0em}
    \item \textbf{Failure Mode I: Signal Degradation.} The model's computational patterns remain largely intact. Quantization error acts as cumulative noise that impairs information precision.
    \item \textbf{Failure Mode II: Computation Collapse.} The quantization error is severe enough to fundamentally damage the functionality of key components. Information cannot be processed correctly and is completely destroyed in the early layers.
\end{itemize}



\section{Mechanistic Validation and Targeted Intervention}
\label{sec:4_validation_and_intervention}


\subsection{Analysis of Component-level Impairment}
\label{sec:4.1_component}

\subsubsection{Attention Patterns} 
\label{sec:4.1.1_attn_mechanism}

A functional attention mechanism should be both focused and accurate; we verify this with normalized attention entropy and focus divergence.

\paragraph{Global Concentration (Entropy).} 
First, we measure the normalized attention entropy to assess if the model can concentrate its attention. For an attention head $h$ at token $t$, we calculate its Shannon entropy $H(A_{h,t})$ and normalize it by the maximum possible entropy: $E_{norm}(h, t) = H(A_{h,t}) / \log_2(t+1)$. We average this across all heads to detect systematic uncertainty.

\paragraph{Focus Divergence (JSD).} 
Entropy alone is insufficient because a model might confidently focus on the wrong token. To measure this deviation, we calculate the Jensen--Shannon divergence (JSD) between the quantized attention distribution ($P_Q$) and the FP16 baseline ($P_{FP}$) at the critical last subject token: $JSD(P_{FP} || P_Q) = \frac{1}{2} D_{KL}(P_{FP} || M) + \frac{1}{2} D_{KL}(P_Q || M)$, where $M$ is the average distribution. A high JSD indicates the focus has shifted significantly.

\begin{figure}[!t]
  \centering
  \begin{subfigure}[b]{0.49\columnwidth}
    \centering
    \includegraphics[width=\textwidth]{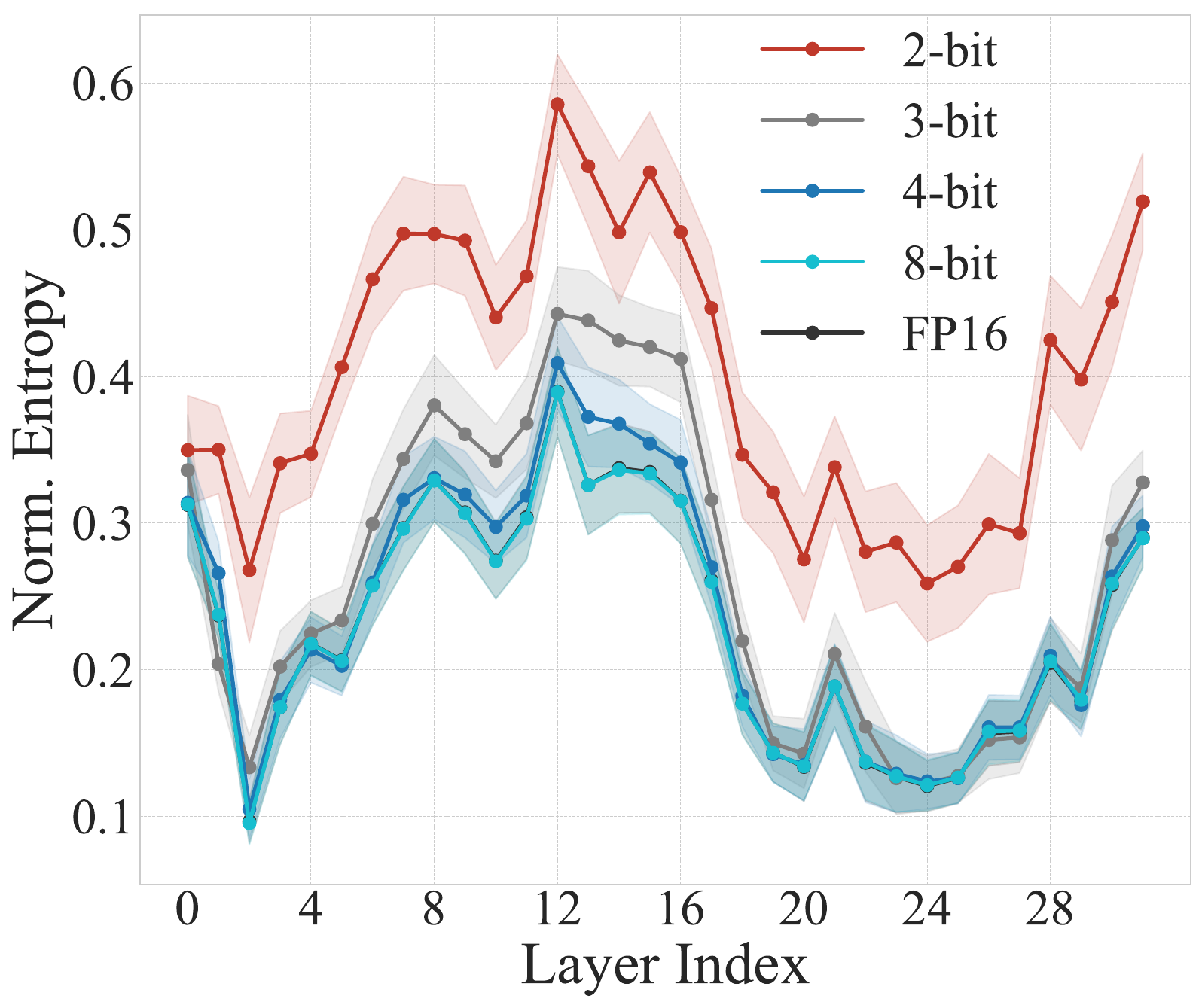} 
    \caption{Global Entropy}
    \label{fig:main_entropy}
  \end{subfigure}
  \hfill
  \begin{subfigure}[b]{0.495\columnwidth}
    \centering
    \includegraphics[width=\textwidth]{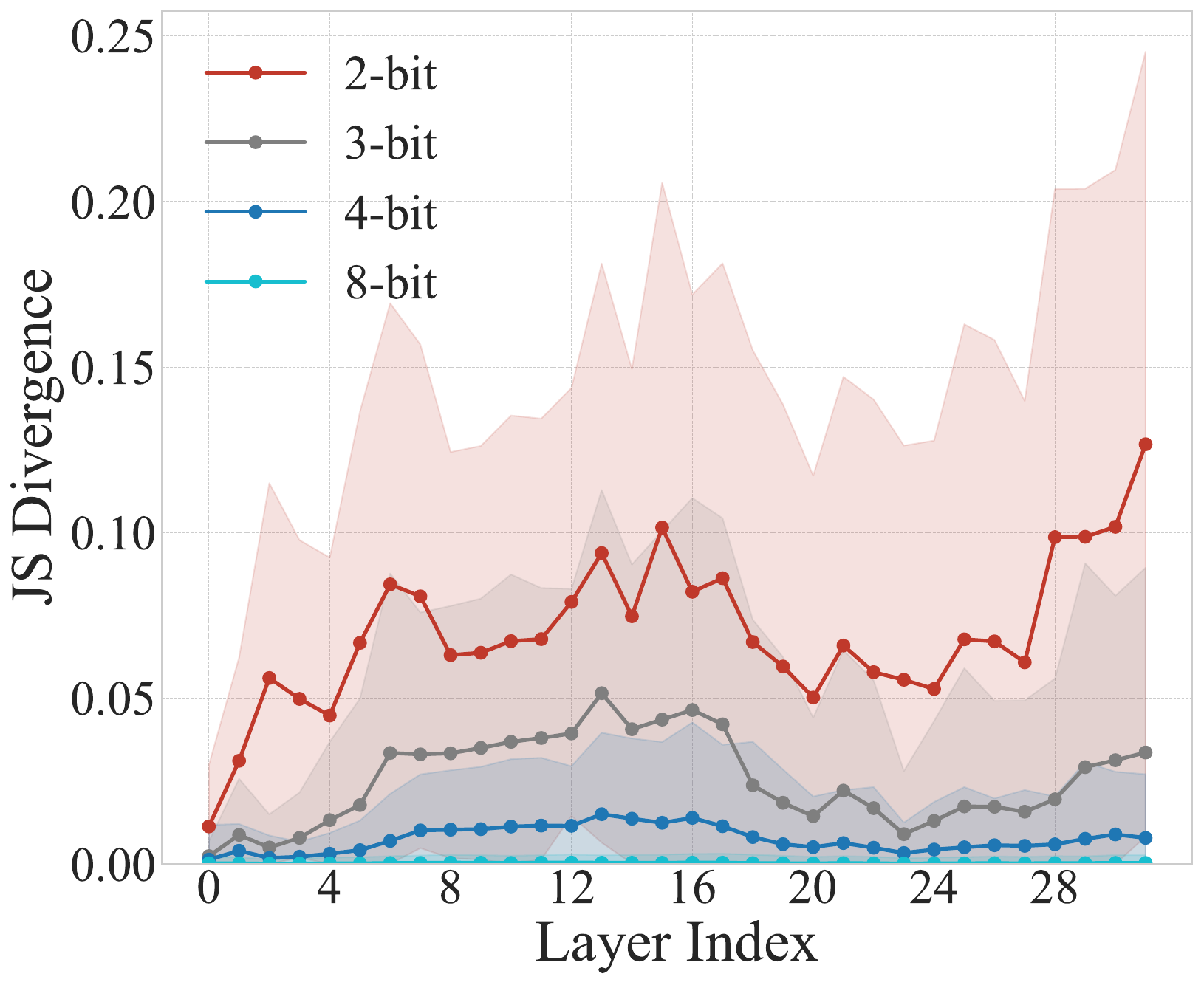} 
    \caption{Focus Divergence}
    \label{fig:main_jsd}
  \end{subfigure}
  \caption{Analysis of attention mechanisms on the Failure Subset. (a) Normalized Attention Entropy (all tokens). (b) Jensen-Shannon Divergence from the FP16 baseline (last subject token).}
  \label{fig:attn_analysis_main} 
\end{figure}

As illustrated in Figure~\ref{fig:attn_analysis_main}, the 4-bit model generally follows the FP16 trend with only slightly increased entropy. In contrast, the 2-bit model exhibits high entropy across all layers, indicating a global failure to concentrate. Meanwhile, its JSD surges significantly, proving that the attention focus deviates fundamentally.

\subsubsection{FFN Key-Value Memory}
\label{sec:4.1.2_ffn_mechanism}


FFN layers function as key-value memories~\cite{geva_emnlp_2021}. For Llama models, the intermediate activation $h_{key} = \text{SiLU}(W_{gate}x) \odot (W_{up}x)$ acts as the ``key'' to select specific expert neurons. We examine the integrity of this key at the last subject token with two metrics.

\paragraph{Gating Consistency (Sign Flip Rate).}
First, we measure the sign flip rate (SFR) of the gate input ($W_{gate}x$). Since the SwiGLU activation depends on the sign, a noise-caused flip ($\text{sign}(x_Q) \neq \text{sign}(x_{FP})$) can fundamentally reverse the neuron's logical state (active vs. suppressed).

\paragraph{Retrieval Accuracy (Jaccard Index).}
Second, we use the Jaccard index to check the Top-1\% activated neurons in $h_{key}$. This measures if the model activates the same neurons as the FP16.

\begin{figure*}[t] 
  \centering
  \begin{subfigure}[b]{0.32\textwidth}
    \centering
    \includegraphics[width=\textwidth]{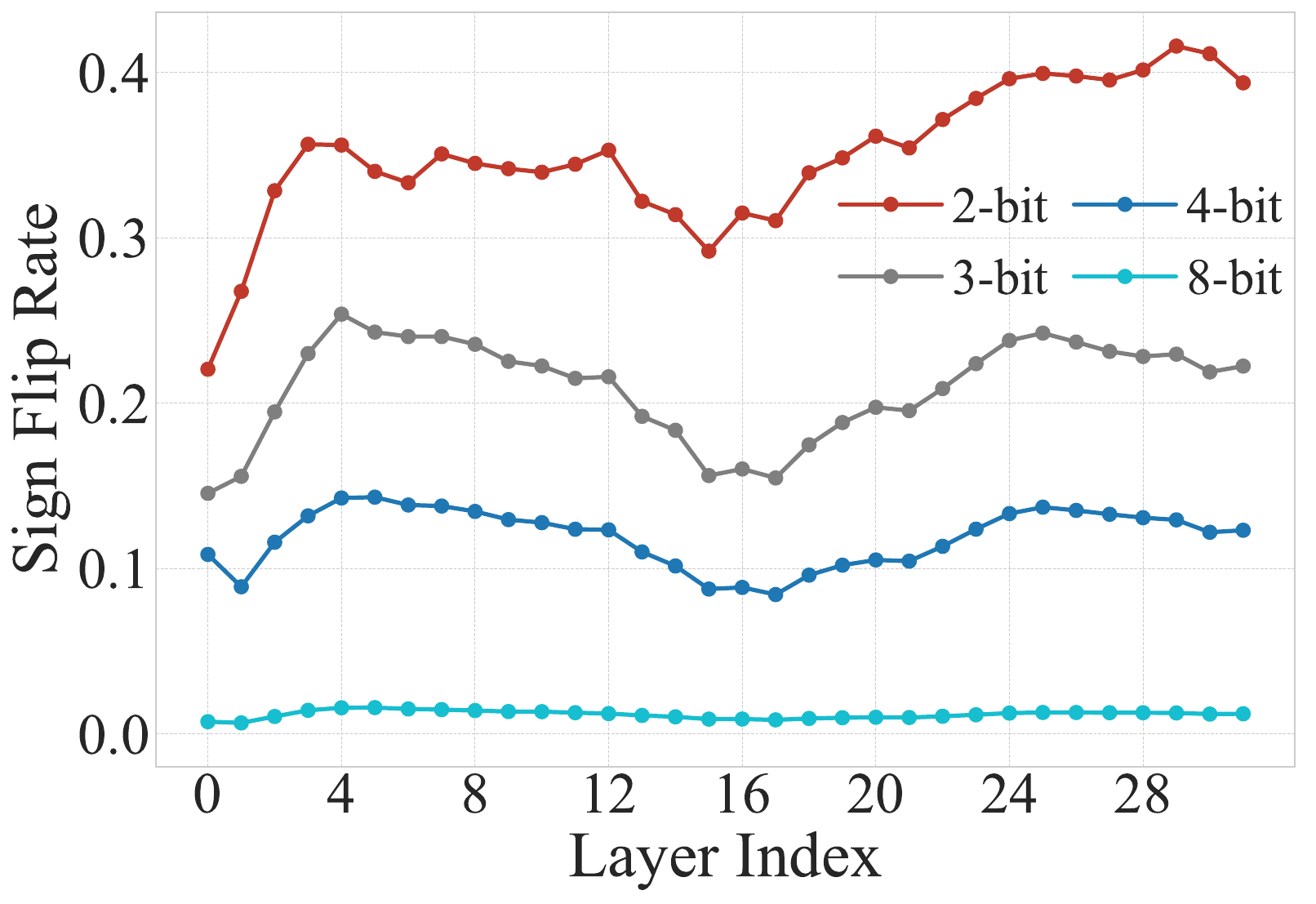} 
    \caption{Gate Sign Flip Rate}
    \label{fig:ffn_sfr}
  \end{subfigure}
  \hfill
  \begin{subfigure}[b]{0.32\textwidth}
    \centering
    \includegraphics[width=\textwidth]{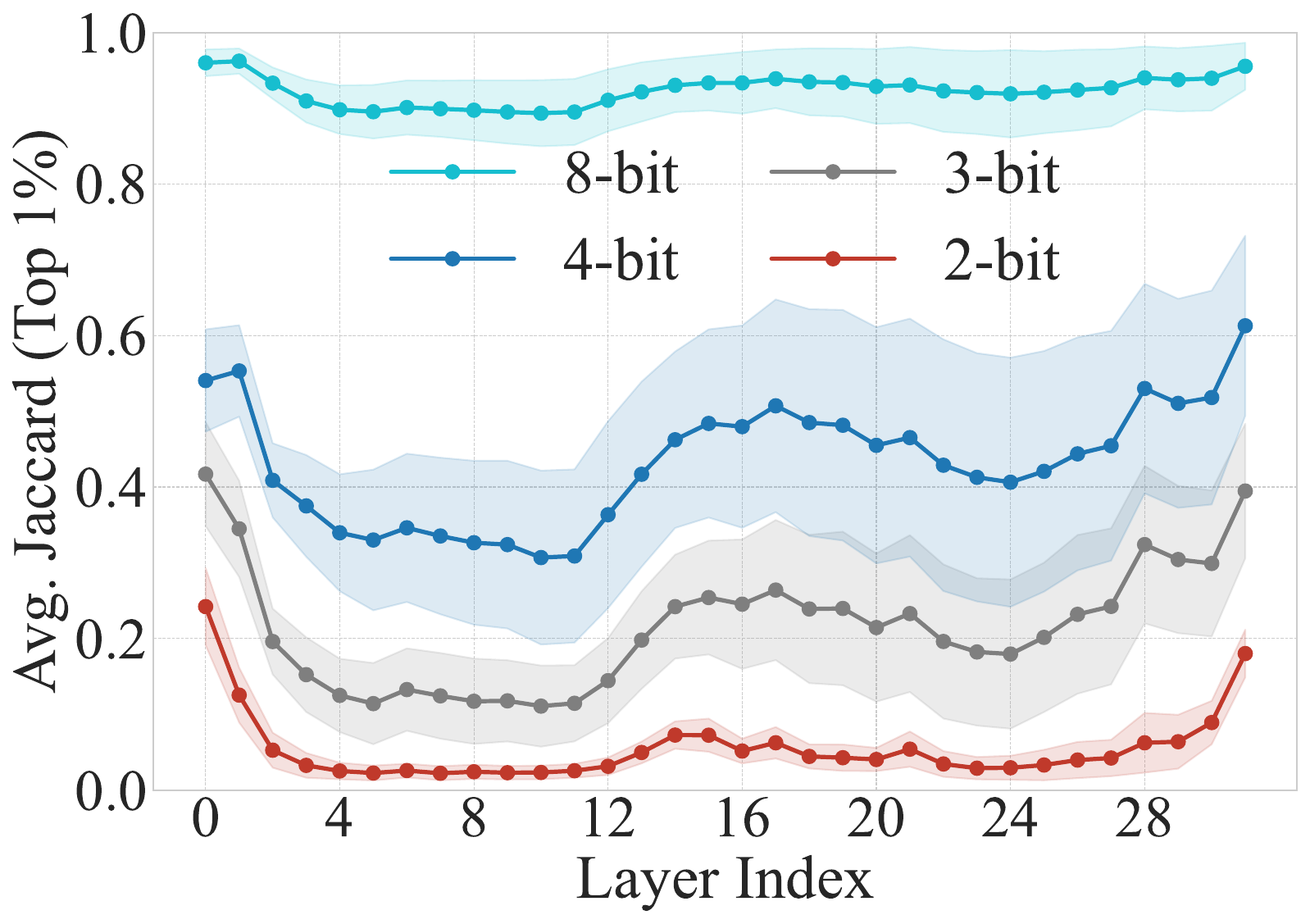}
    \caption{Expert Jaccard Similarity (Top-1\%)}
    \label{fig:ffn_jaccard}
  \end{subfigure}
  \hfill
  \begin{subfigure}[b]{0.32\textwidth}
    \centering
    \includegraphics[width=\textwidth]{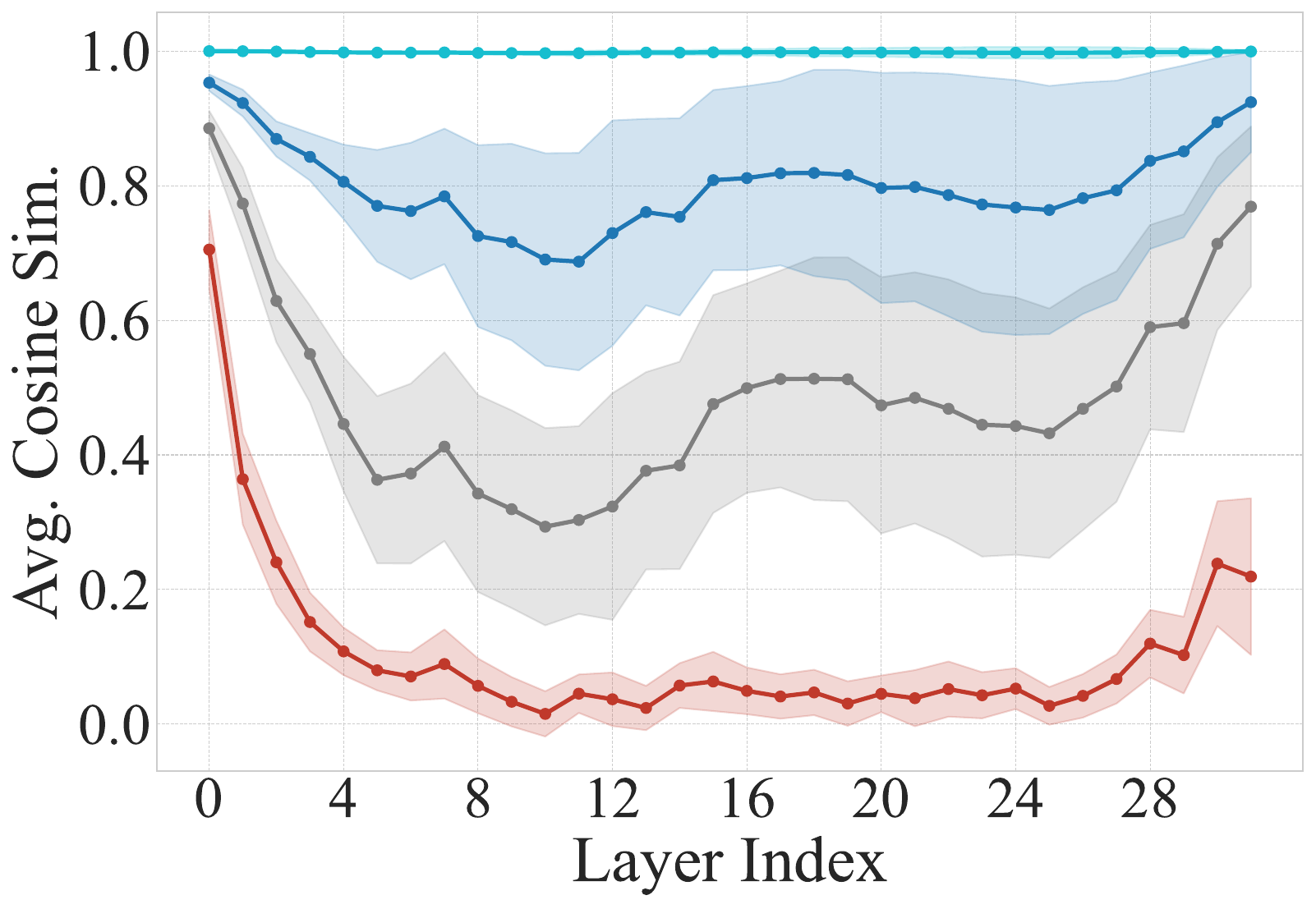}
    \caption{Value Similarity (Cosine)}
    \label{fig:ffn_value_analysis}
  \end{subfigure}
  \caption{Analysis of FFN Key-Value Memory at the last subject token on the Failure Subset. 
  The 2-bit exhibits high gating instability (a) and low expert overlap (b), leading to semantic collapse (c).
  }
  \label{fig:ffn_analysis_combined} 
\end{figure*}

As shown in Figure~\ref{fig:ffn_sfr}, the 2-bit model exhibits a high sign flip rate ($>30\%$), indicating quantization noise is large enough to reverse the gate direction. Consequently, the Jaccard Index drops to $\approx 0.1$ (Figure~\ref{fig:ffn_jaccard}), confirming the model activates the wrong neurons. In contrast, the 4-bit model maintains high gating consistency and retrieval overlap.

\paragraph{Analysis of Values (Semantic Direction).}

Finally, we check the output quality by measuring the cosine similarity between the quantized FFN output ($h_{value}= W_{down} h_{key}$) and the FP16 baseline. This tells us if the retrieved information has the correct semantic direction.

Figure~\ref{fig:ffn_value_analysis} confirms the contrast. The 4-bit model maintains high similarity ($\approx 0.8$) even when it fails, implying it retrieves the correct concept but with precision errors. In contrast, the 2-bit model drops to near-zero immediately, confirming the retrieved information is completely unrelated to the target. Similar patterns were observed on the Robust Subset, see Appendix~\ref{sec:appendix_component}.

\subsection{Analysis of Representation-level Deviation}
\label{sec:4.2_representation}

Building on the component-level findings, we now examine whether the quantization noise merely blurs the signal or fundamentally destroys the structural integrity of the representation space.

\subsubsection{Analysis of Representational Topology}
\label{sec:4.2.1_cka_topology}

We employ linear centered kernel alignment (CKA)~\cite{kornblith_icml_2019} to analyze the structural correspondence between the activation matrices of quantized and FP16 models.


\begin{figure}[!t]
  \centering
  \includegraphics[width=\columnwidth]{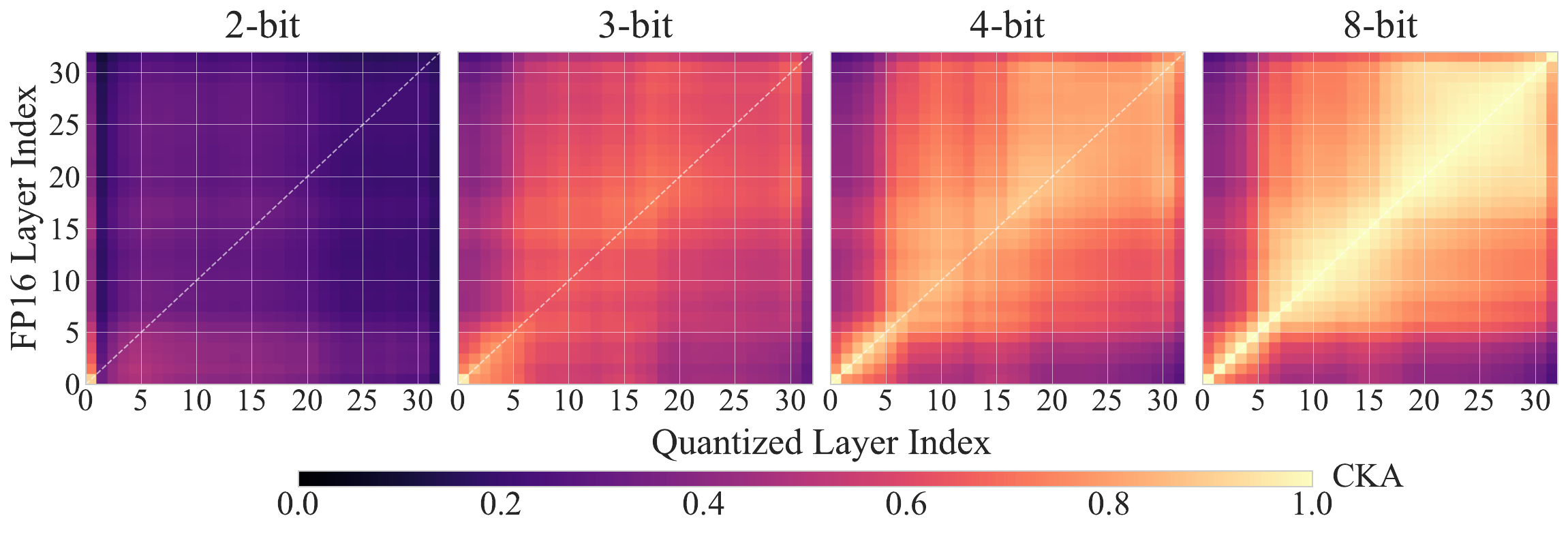}
  \caption{CKA heatmaps of hidden states at the last subject token. 
  }
  \label{fig:cka_analysis_main} 
\end{figure}

Figure~\ref{fig:cka_analysis_main} visualizes results at the last subject token, the critical site for knowledge extraction~\cite{meng_nips_2022} validated by our earlier causal tracing. The diagonal line represents layer-wise correspondence, indicating behavioral similarity to the FP16 model at the same layer. We observe a sharp contrast between the two failure modes. The 4-bit model retains a bright diagonal and block structure similar to 8-bit, only with slightly reduced intensity. This confirms that the global representational structure is preserved. Conversely, the 2-bit model appears almost entirely dark purple. The absence of diagonal structure indicates a ``Structural Collapse,'' where the representational spaces are totally different. Component-wise breakdowns and positional validation are shown in Appendix~\ref{sec:appendix_representation}.

\subsubsection{Analysis of Semantic Subspace}
\label{sec:semantic_subspace}

While CKA analyzes the global topology, we use singular value decomposition (SVD) to inspect the internal structure of the activation matrices ($A$). We conduct two complementary analyses on the Failure Subset.

\paragraph{Activation Subspace Alignment.}

First, we check if the quantized models utilize the same semantic directions as the FP16 model. We compare the top-$k$ principal directions (columns of $V$, where $A = USV^T$). We set $k=50$ (capturing $>90\%$ of spectral energy) to isolate core semantics from long-tail noise. Let $V_{fp, k}$ and $V_{q, k}$ be the subspaces of the FP16 and quantized models. We calculate their similarity as:
\begin{equation} \label{eq:subspace_sim}
Sim(V_{fp}, V_q) = \frac{1}{k} \sum_{i=1}^{k} \sigma_i(V_{fp, k}^T V_{q, k})^2
\end{equation}

\begin{figure}[!t]
    \centering
    \begin{subfigure}[b]{0.48\columnwidth}
        \centering
        \includegraphics[width=\textwidth]{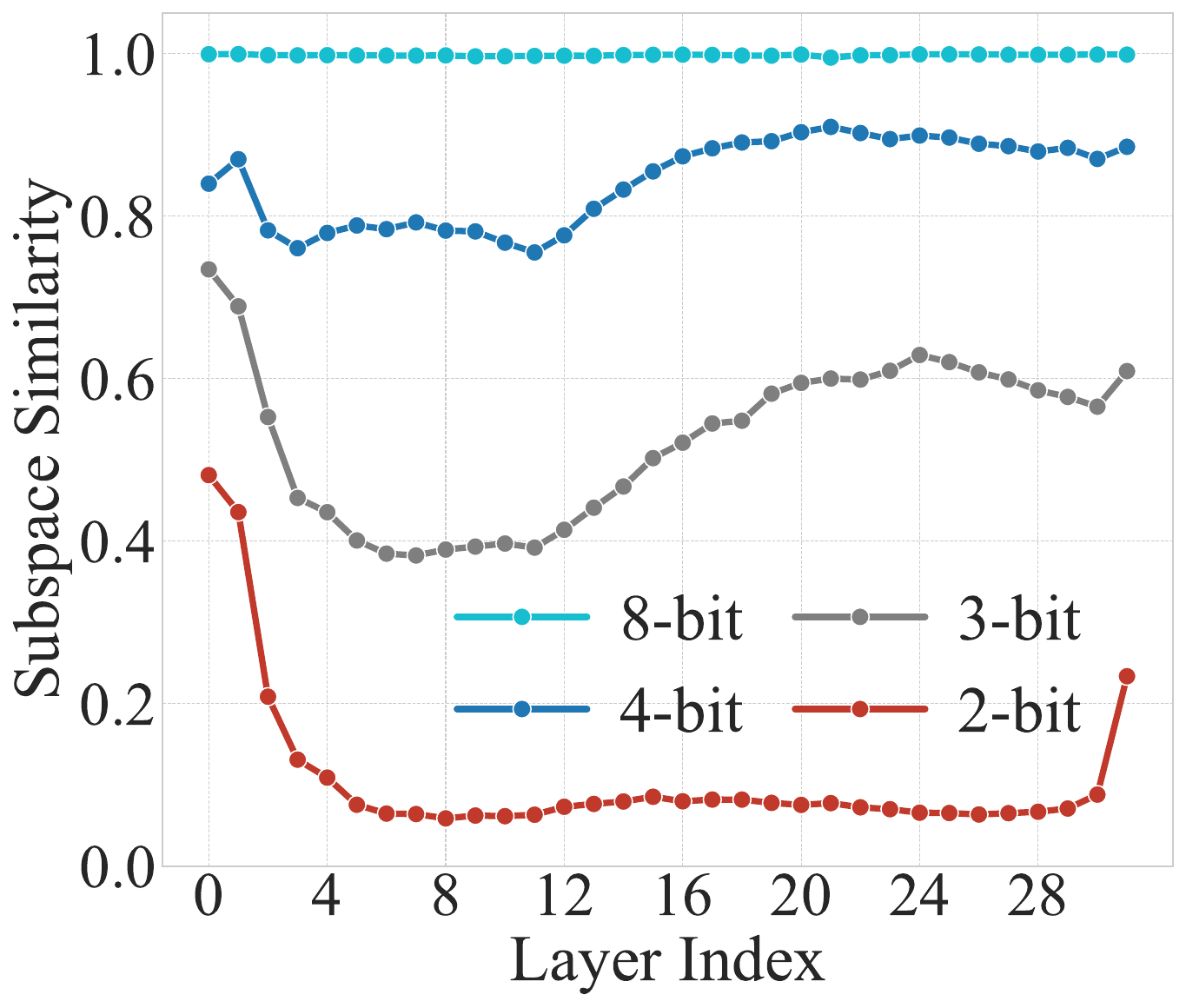} %
        \caption{Activation Subspace}
        \label{fig:svd_activation_subspace}
    \end{subfigure}
    \hfill
    \begin{subfigure}[b]{0.48\columnwidth}
        \centering
        \includegraphics[width=\textwidth]{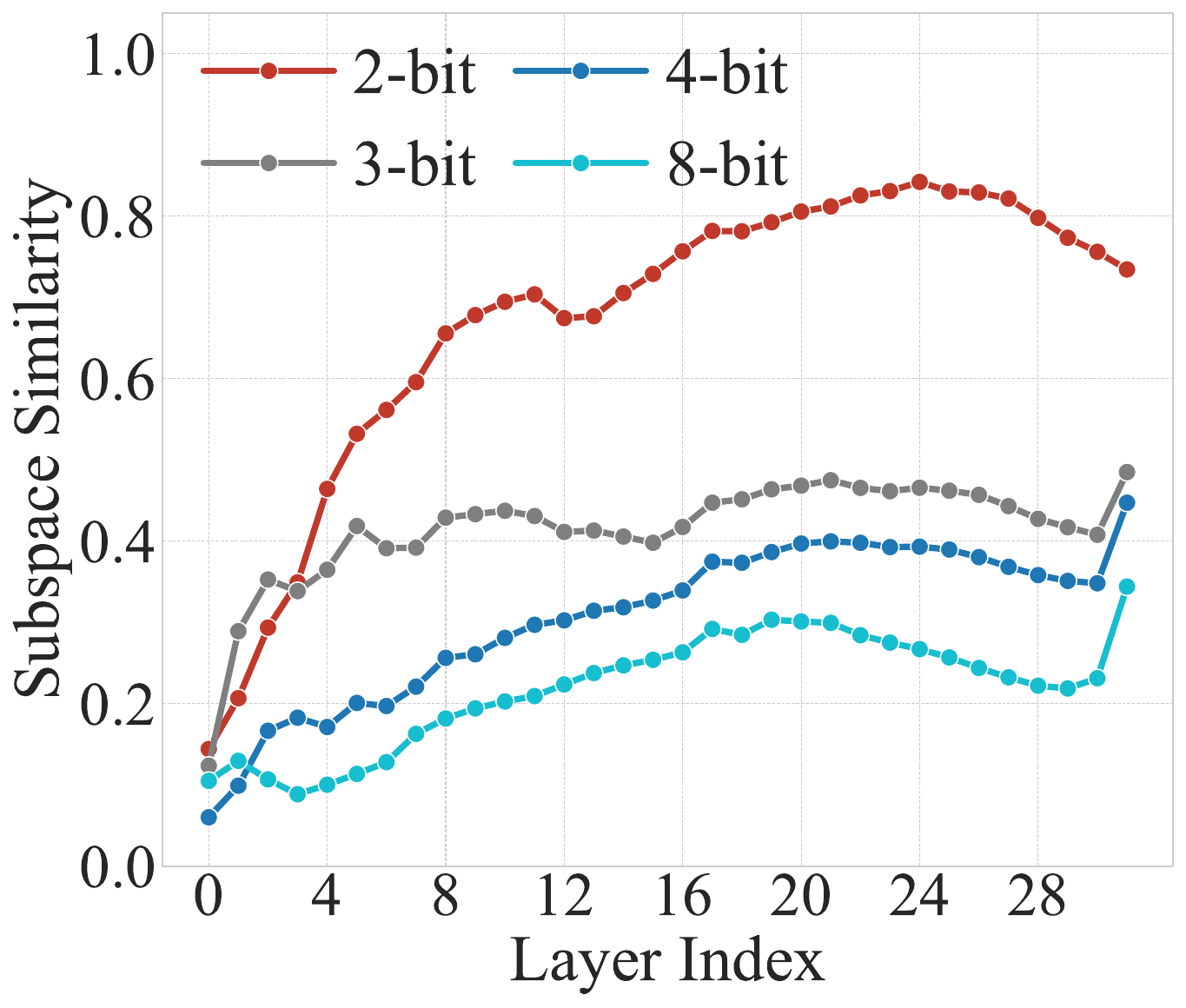} 
        \caption{Error Subspace}
        \label{fig:svd_error_subspace}
    \end{subfigure}
    \caption{Layer-wise SVD analysis (Top-50 dimensions) on the Failure Subset. (a) Similarity of activation subspaces to FP16. (b) Alignment between quantization error and FP16 subspaces.}
    \label{fig:svd_analysis}
\end{figure}

Figure~\ref{fig:svd_analysis}(a) shows that the 4-bit model maintains high similarity ($>0.8$) to FP16, confirming its core computational directions remain largely intact even when the model fails. In contrast, the 2-bit model drops to near-zero similarity, indicating a complete loss of the original semantic directions.

\paragraph{Error Subspace Analysis.}
While activation subspace analysis confirms the deviation of representation directions, it doesn't explain whether the error aligns with the original signal. Consequently, we decompose the error matrix ($E = A_q - A_{fp}$) and measure the alignment between principal error directions ($V_{err}$) and original signal directions ($V_{fp}$).

Figure~\ref{fig:svd_analysis}(b) reveals a critical difference. The 2-bit error is highly aligned with the signal subspace (similarity $\approx 0.8$). This means the quantization error is not random noise but directly interferes with the model's primary features. Conversely, the 4-bit error is much less alignment ($\approx 0.3$), resembling random noise that affects precision without destroying signal structure.
Results on the Robust Subset are consistent and shown in Appendix~\ref{sec:appendix_representation}.


\noindent \textbf{Summary of Diagnosis.}
Combining the component-level and representation-level evidence, we confirm the existence of two distinct failure modes. The 4-bit models exhibit Signal Degradation, where representations are impaired but structurally intact. Conversely, standard 2-bit models exemplify Computation Collapse, where both component functionality and semantic structure are fundamentally destroyed. 
Crucially, these failures are not strictly tied to specific bit-widths, but reflect the distinct nature of the damage.

\subsection{Mechanism-Aware Interventions}
\label{sec:4.3_intervention}

Guided by the mechanistic diagnosis, we now demonstrate that the Signal Degradation mode (typical in 4-bit) is localizable and repairable, whereas the Computation Collapse mode (observed in 2-bit) proves systemic and irreversible without retraining.

\subsubsection{Signal Degradation: Localization and Repair}
\label{sec:4.3.1_intervention_4bit}

The Signal Degradation hypothesis implies that the impairment is not structural but cumulative. We validate this by locating the degradation source and designing a targeted repair.

\begin{figure}[!t]
    \centering
    \includegraphics[width=\columnwidth]{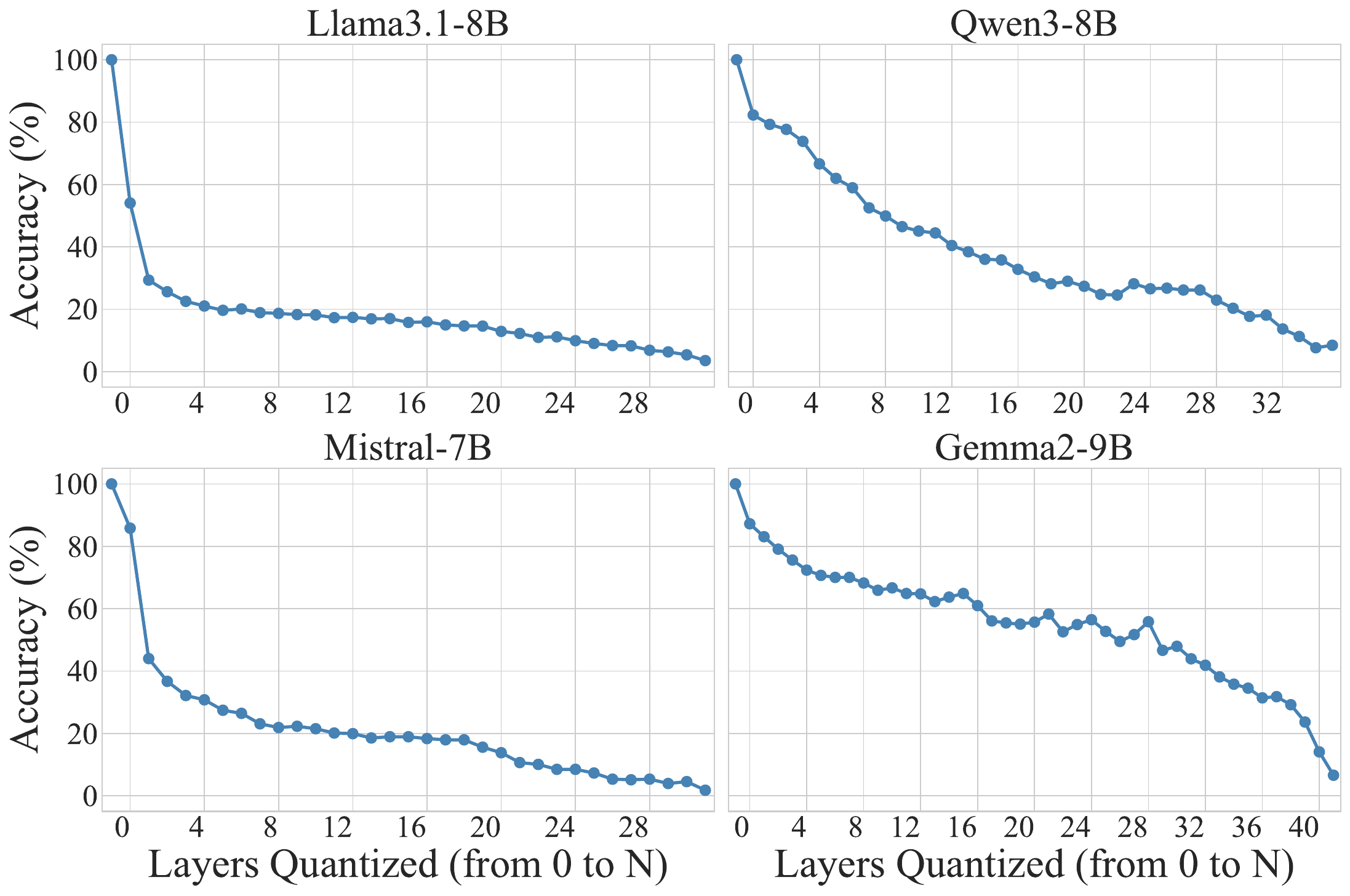} %
    \caption{Progressive 4-bit quantization (``domino effect'') analysis on the Failure Subset.}
    \label{fig:domino_4bit}
\end{figure}

\paragraph{Localization: The ``First Domino'' Test.} 
To locate failure origins, we conduct a ``domino effect'' experiment by progressively quantizing the model from layer 0 to $k$ in 4-bit, keeping subsequent layers in FP16. Figure~\ref{fig:domino_4bit} reveals two distinct, architecture-dependent degradation patterns: 
\textbf{(1) Early Representation Bottleneck} (Llama3.1, Mistral): Accuracy drops sharply when quantizing only the first few layers.
\textbf{(2) Uniform Degradation} (Qwen3, Gemma2): Performance declines smoothly across all layers.
Complementary single-layer quantization and component-level sensitivity analysis are provided in Appendix~\ref{sec:appendix_interntion_4bit}.


\paragraph{Intervention: A Two-Stage Repair Strategy.} 
Guided by the localization, we design a two-stage intervention to recover the degraded signal.


\noindent\textbf{(1) Source Protection.} We first apply targeted protection to mitigate error at its primary sources. For Llama/Mistral, we apply early-layer protection, retaining the first two layers in 8-bit (~4.25 avg. bits). For Qwen/Gemma, where sensitivity is distributed, we apply kurtosis-based protection (~4.1 avg. bits), preserving high-kurtosis weights that are most vulnerable. This aligns with mixed-precision methods like SPQR~\cite{dettmers_iclr_2023}, which keep sensitive weights in high precision. It supports our diagnosis that protecting critical components effectively prevents degradation.

\begin{figure}[!t]
    \centering
    \includegraphics[width=\columnwidth]{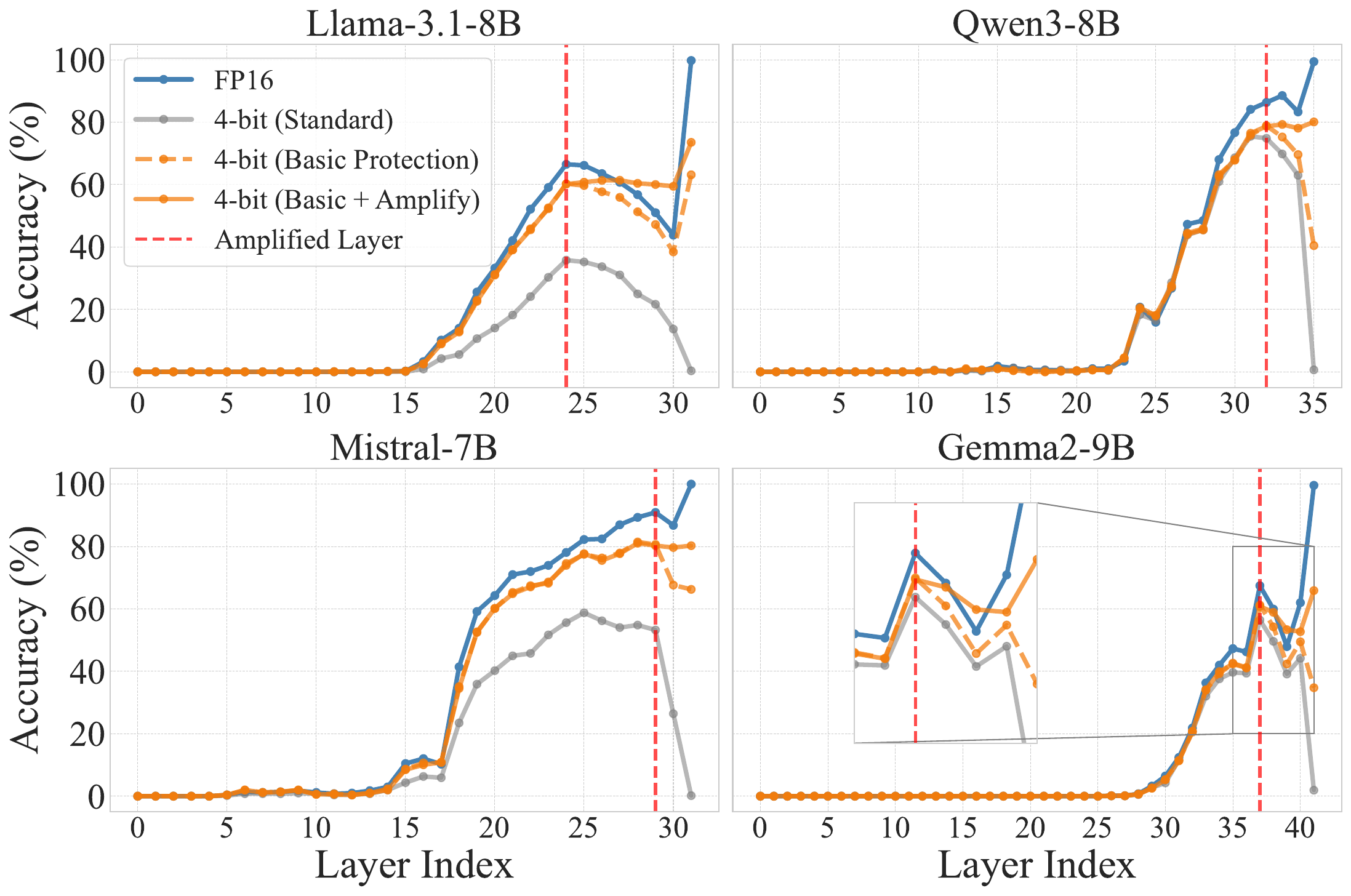}
    \caption{Logit Lens accuracy on the Failure Subset. Our two-stage strategy (orange lines) restores the degraded baseline toward FP16.}
    \label{fig:intervention_logit_lens}
\end{figure}

Figure~\ref{fig:intervention_logit_lens} (dashed orange) shows this basic protection improves internal signal quality over the baseline (gray). However, final-layer accuracy still lags as cumulative errors weaken the signal until it is surpassed by linguistic noise.

\noindent\textbf{(2) Signal Restoration.} To counteract the late-stage competition failure, we introduce peak signal amplification. We identify the layer with the highest confidence (lowest entropy) and amplify its output logits by a factor $\alpha > 1$. As shown in Figure~\ref{fig:intervention_logit_lens} (solid orange), this corrects the late-stage drop and restores the trajectory close to FP16.

As summarized in Table~\ref{tab:4bit_intervention_summary}, this combined strategy yields substantial gains across all models, confirming that 4-bit failure is a recoverable impairment of signal intensity.

\begin{table}[!t]
\centering
\small
\begin{tabular}{@{}lccc@{}}
\toprule
\textbf{Model} & \textbf{Baseline} & \textbf{+ Basic} & \textbf{+ Basic \& Amplify} \\
& \textbf{(4-bit)} & \textbf{Repair} & \textbf{(Final)} \\
\midrule
Llama3.1-8B & 0.00\% & 67.91\% & \textbf{75.19\%} ($\alpha$=3) \\
Mistral-7B    & 0.00\% & 66.86\% & \textbf{81.26\%} ($\alpha$=9) \\
Qwen3-8B      & 0.00\% & 40.24\% & \textbf{79.88\%} ($\alpha$=7) \\
Gemma2-9B     & 0.00\% & 33.85\% & \textbf{64.08\%} ($\alpha$=2) \\
\bottomrule
\end{tabular}
\caption{4-bit intervention results on the Failure Subset.}
\label{tab:4bit_intervention_summary}
\end{table}

\subsubsection{Computation Collapse: Systemic Irreversibility}
\label{sec:4.3.2_intervention_2bit}

In contrast, we posit that Computation Collapse is a systemic processing failure. We validate its irreversibility under training-free interventions through three complementary analyses.

\paragraph{(1) Irreversibility of Damage.}
We apply the same ``domino'' test to 2-bit models. Table~\ref{tab:domino_2bit} shows catastrophic results: for Llama3, quantizing just the first two layers ($k=1$) causes accuracy to plummet from 100\% to 41.65\%. This proves that 2-bit damage is instantaneous and irreversible, where the signal is destroyed at the source, and even 30 subsequent FP16 layers cannot recover it.

\begin{table}[!t]
\centering
\small
\setlength{\tabcolsep}{13pt}
\begin{tabular}{@{} l S[table-format=3.2] S[table-format=2.2] @{}}
\toprule
& \multicolumn{2}{c}{\textbf{Accuracy on Subsets (\%)}} \\
\cmidrule(lr){2-3}
\textbf{Quantized Layers} & \textbf{Robust} & \textbf{Failure} \\
\midrule
None (FP16)        & 100.00 & 100.00 \\
$k=0$ (Layer 0)      & 65.47  & 15.03  \\
$k=1$ (Layers 0-1)   & 41.65  & 5.29   \\
$k=2$ (Layers 0-2)   & 24.66  & 2.50   \\
$k=3$ (Layers 0-3)   & 10.72  & 1.04   \\
$k=5$ (Layers 0-5)   & 2.51   & 0.38   \\
\bottomrule
\end{tabular}
\caption{The ``domino effect'' of 2-bit damage on Llama3.1-8B. Models are quantized from layer 0 to $k$ in 2-bit, with subsequent layers remaining FP16.}
\label{tab:domino_2bit}
\end{table}

\begin{figure}[!t]
    \centering
    \begin{subfigure}[b]{0.48\columnwidth}
        \includegraphics[width=\textwidth]{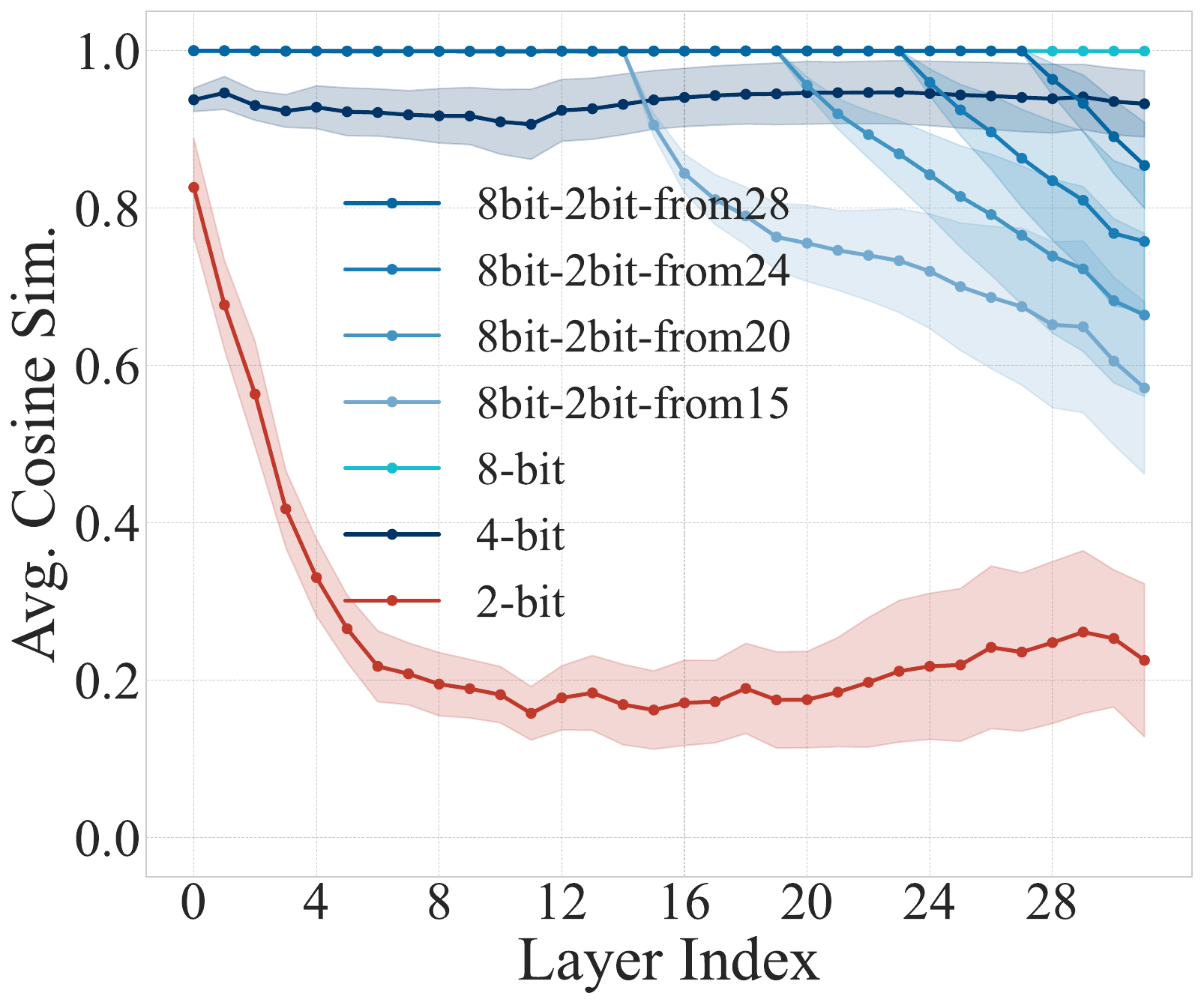} %
        \caption{8-bit to 2-bit}
        \label{fig:signal_protect_8bit}
    \end{subfigure}
    \hfill
    \begin{subfigure}[b]{0.48\columnwidth}
        \includegraphics[width=\textwidth]{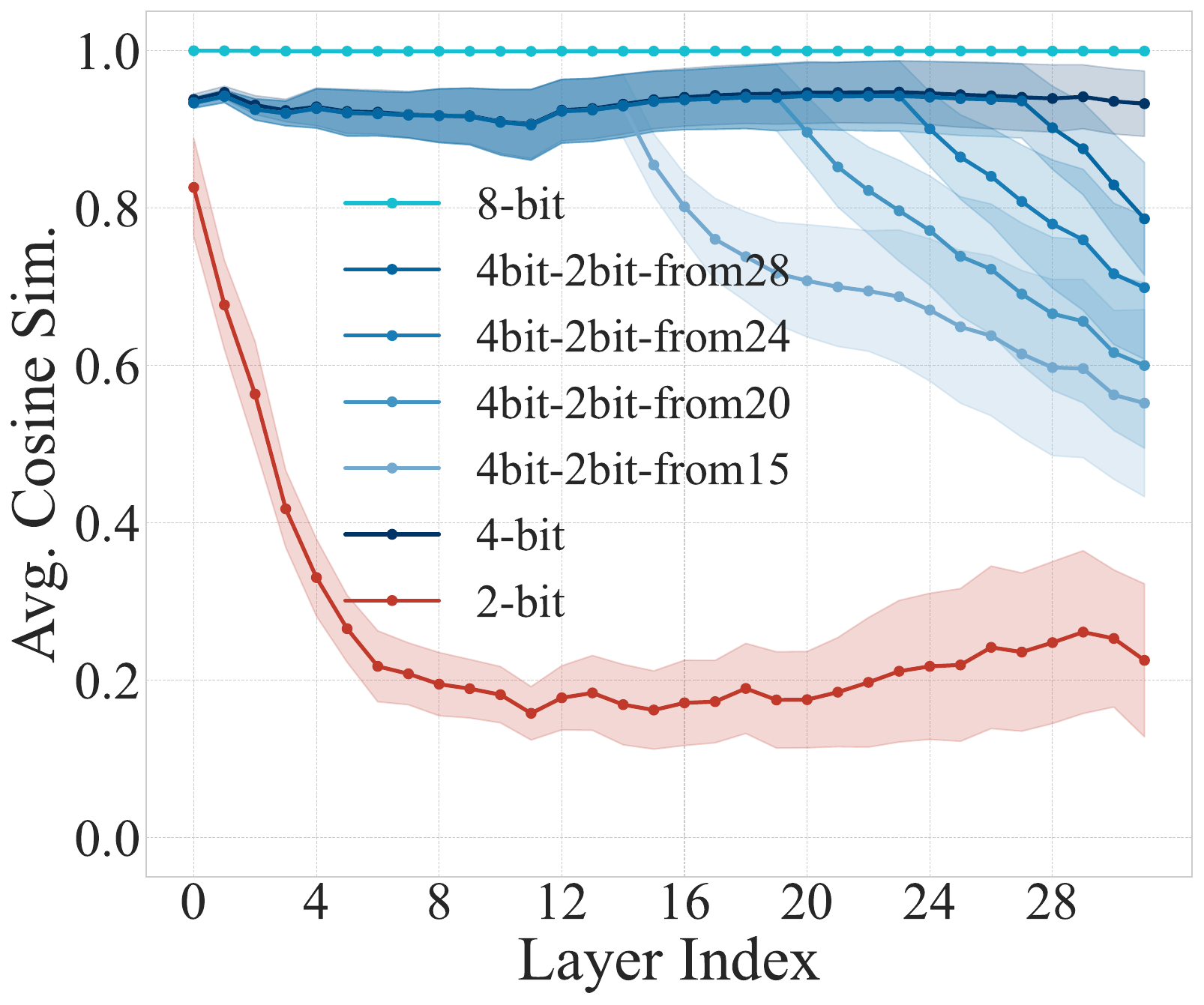} %
        \caption{4-bit to 2-bit}
        \label{fig:signal_protect_4bit}
    \end{subfigure}
    \caption{Layer output cosine similarity under high-precision signal injection on the Robust Subset.}
    \label{fig:signal_protect_test}
\end{figure}

\paragraph{(2) Failure to Process High-Precision Signals.}
We further test if 2-bit components can function when provided with high-quality signal. We keep the first $k$ layers at high precision (8/4-bit) and quantize subsequent layers to 2-bit.
As Figure~\ref{fig:signal_protect_test} shows, cosine similarity remains high ($>0.9$) initially but collapses immediately upon entering the 2-bit layers. This confirms 2-bit components are computationally non-functional, failing to sustain information even given perfect input. Component-level analysis is shown in Appendix~\ref{sec:appendix_interventions}.

\paragraph{(3) Failure of Mere Compensation.}
We attempt to recover performance using both our protection strategies (highly effective against Signal Degradation) and EORA~\cite{liu_eora_2024}, an advanced low-rank compensation method. However, the 2-bit collapse resists all such interventions. This confirms that the failure stems from a fundamental component malfunction rather than localized precision loss, necessitating structural reconstruction (e.g., fine-tuning) rather than mere compensation.


\section{Conclusion}

In this work, we bridge the macroscopic performance cliff with microscopic mechanistic failures. We propose and validate the Two Failure Modes Hypothesis, distinguishing between Signal Degradation and Computation Collapse. Our analysis reveals a qualitative shift from Signal Degradation (impaired but functional) to Computation Collapse (fundamental component malfunction). Crucially, the distinct repairability of these modes implies that the collapse necessitates reconstructing computational functionality rather than simple compensation. This work offers a diagnostic foundation for future principled quantization.

\section*{Limitations}
Our investigation currently focuses on weight-only quantization across representative model families. Consequently, extending these findings to other paradigms, such as activation quantization, remains a direction for future work. Additionally, our evaluation anchors on factual knowledge recall; how the identified failure modes manifest in complex reasoning tasks deserves separate investigation.

\section*{Acknowledgments}
This work was supported by Beijing Natural Science Foundation (L243006), the National Natural Science Foundation of China (No.62406321), the independent research project of the Key Laboratory of Cognition and Decision Intelligence for Complex Systems and CIPS-SMP-Zhipu Large Model Fund.



\bibliography{main}

\clearpage

\appendix

\section{Experimental Details}
\label{sec:app_setup}

\subsection{Quantization Configuration}

We use GPTQModel for post-training quantization with group size 128. Calibration is performed on 128 randomly sampled C4 sequences of length 2048~\cite{raffel_exploring_2020}. All subsequent evaluations use greedy decoding (temperature = 0) to ensure deterministic inference.

\subsection{Prompt Templates}
\label{sec:app_prompts}

\paragraph{Primary Templates (Mechanistic Analysis).}
For the primary mechanistic analysis, we select one specific template per relation that naturally ends with the object to facilitate next-token probing. Table~\ref{tab:app_prompts} provides examples of the templates used for different relation types.

\begin{table}[h]
    \centering
    \small
    \begin{tabular}{l|l}
        \toprule
        \textbf{Relation ID} & \textbf{Template} \\
        \midrule
        P19 (Place of Birth) & [X] was born in \\
        P27 (Country of Citizenship) & [X] is a citizen of \\
        P36 (Capital) & The capital of [X] is \\
        P106 (Profession) & The profession of [X] is \\
        \bottomrule
    \end{tabular}
    \caption{Examples of standardized templates used for mechanistic analysis.}
    \label{tab:app_prompts}
\end{table}

\paragraph{Robustness Templates (Phenomenological Check).}
For the robustness evaluation in Figure~\ref{fig:robust_eval_4rels}, we utilize the full set of Pararel paraphrases. To handle varying target positions [Y] across patterns (e.g., ``[X]'s capital is [Y]'', ``[Y] is the capital of [X]''), we standardize the input by wrapping statements into an instruction: 
\emph{Based on your knowledge, complete the following sentence by filling in the blank: '\{cloze\_statement\}' The missing word is:} 
This ensures the model generates the target entity as the immediate completion, regardless of the original sentence structure.

\subsection{Dataset Partition Statistics}

Table~\ref{tab:app_subsets} details the sample counts for the Robust Subset (\texttt{fp\_and\_4bit\_correct}) and the Failure Subset (\texttt{fp\_correct\_4bit\_wrong}) across all evaluated models.

\begin{table}[t]
    \centering
    \small
    \begin{tabular*}{0.9\linewidth}{l@{\extracolsep{\fill}}ccc}
        \toprule
        \textbf{Model} & \textbf{Total} & \textbf{Robust} & \textbf{Failure} \\
        \midrule
        Llama-3.1-8B & 7,777 & 5,661 & 2,116 \\
        Qwen3-8B   & 4,055 & 3,558 & 497 \\
        Mistral-7B   & 4,294 & 3,787 & 507 \\
        Gemma-2-9B   & 6,375 & 5,601 & 774 \\
        \bottomrule
    \end{tabular*}
    \caption{Sample counts for analysis subsets.}
    \label{tab:app_subsets}
\end{table}

\section{Supplementary Mechanistic Validation}
\label{sec:appendix_mechanism}

\subsection{Component-level Impairment}
\label{sec:appendix_component}

\begin{figure*}[!t]
    \centering
    \begin{subfigure}[b]{0.32\textwidth}
        \centering
        \includegraphics[width=\textwidth]{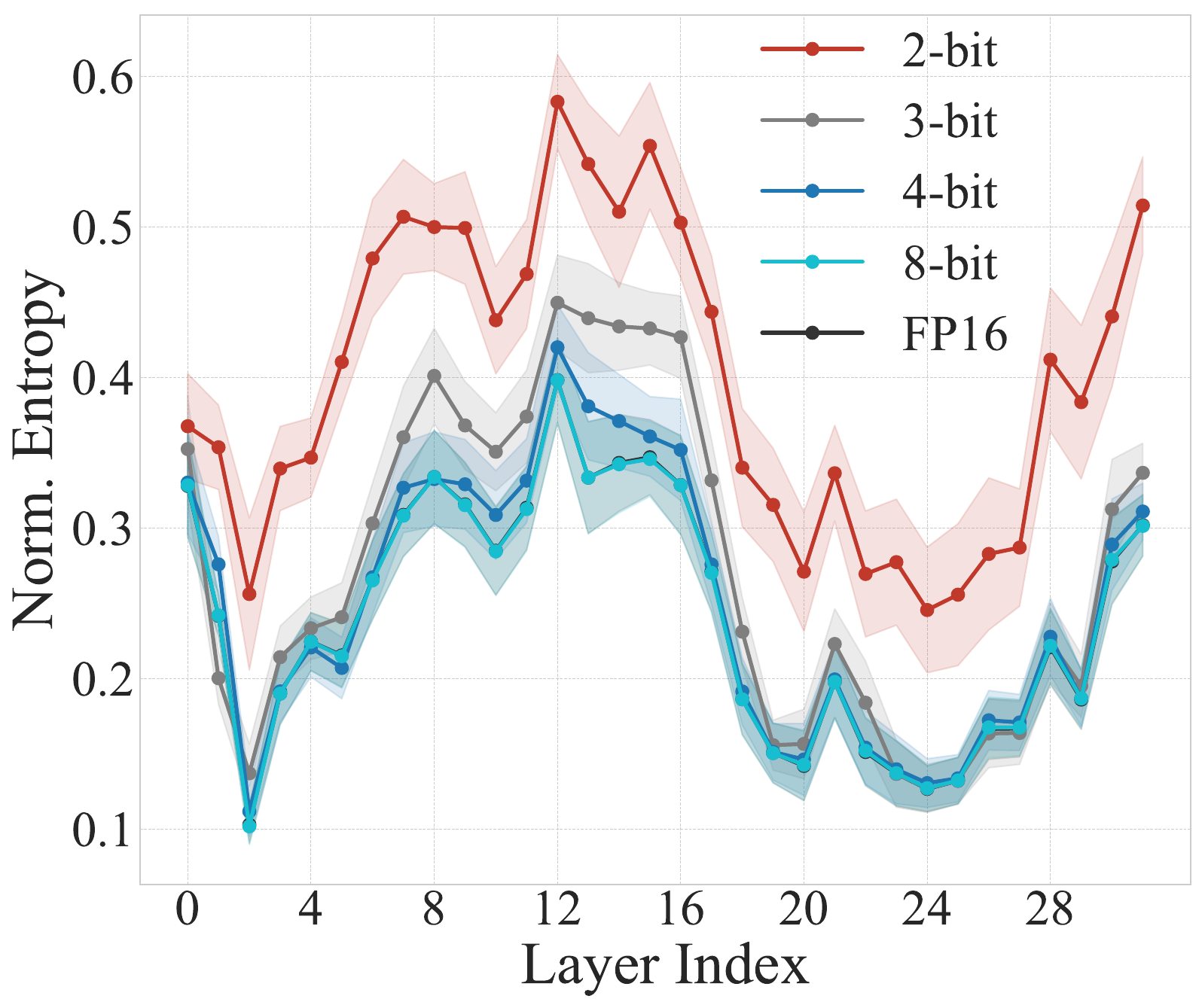}
        \caption{All tokens on the Robust Subset}
        \label{fig:app_entropy_correct_all}
    \end{subfigure}
    \hfill
    \begin{subfigure}[b]{0.32\textwidth}
        \centering
        \includegraphics[width=\textwidth]{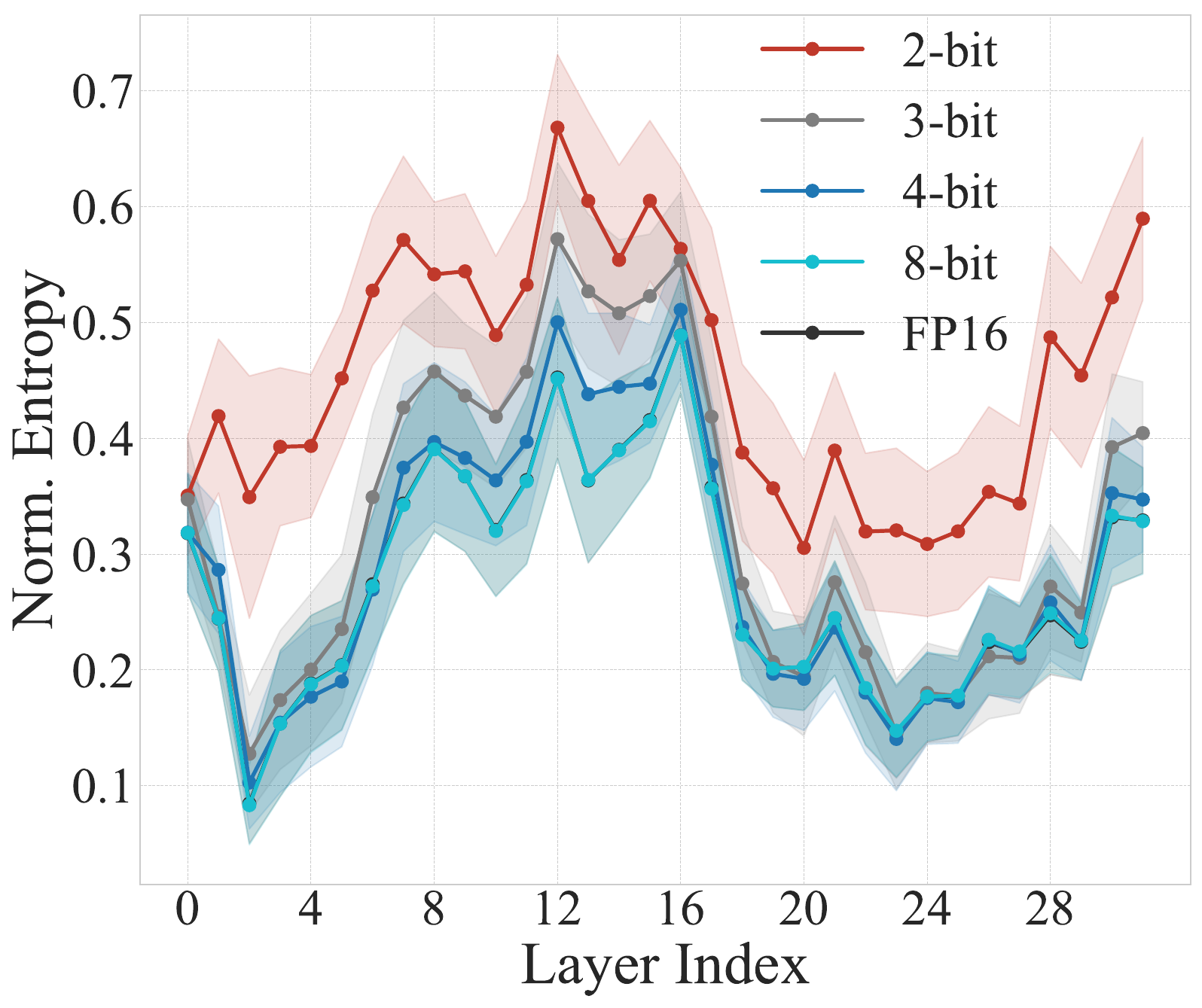}
        \caption{Last subject token on the Failure}
        \label{fig:app_entropy_wrong_subject}
    \end{subfigure}
    \hfill
    \begin{subfigure}[b]{0.32\textwidth}
        \centering
        \includegraphics[width=\textwidth]{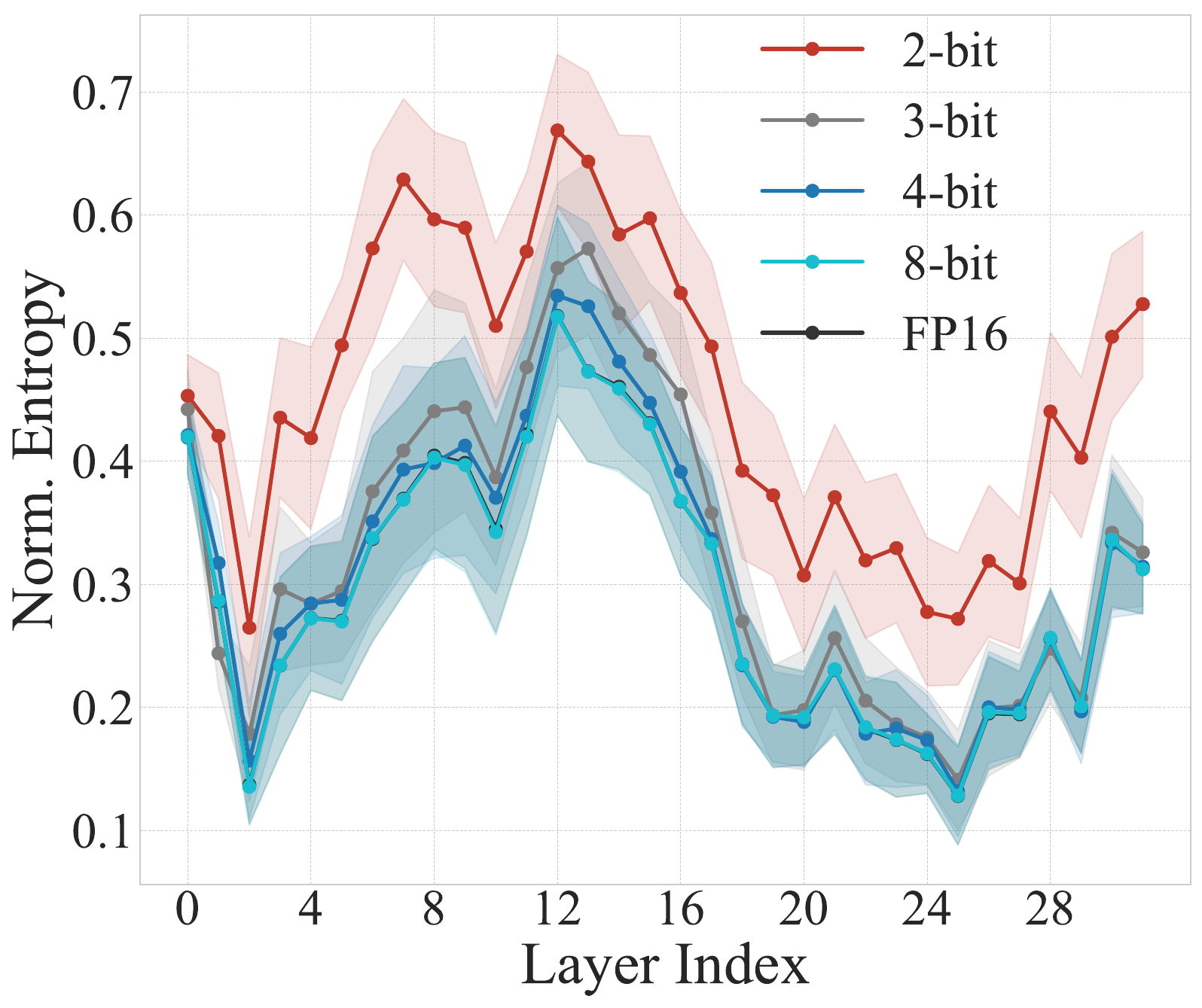}
        \caption{Last token on the Failure Subset}
        \label{fig:app_entropy_wrong_last}
    \end{subfigure}
    \caption{Supplementary results for Normalized Attention Entropy.}
    \label{fig:app_attn_entropy}
\end{figure*}

\begin{figure*}[!t]
    \centering
    \begin{subfigure}[b]{0.45\textwidth}
        \centering
        \includegraphics[width=\textwidth]{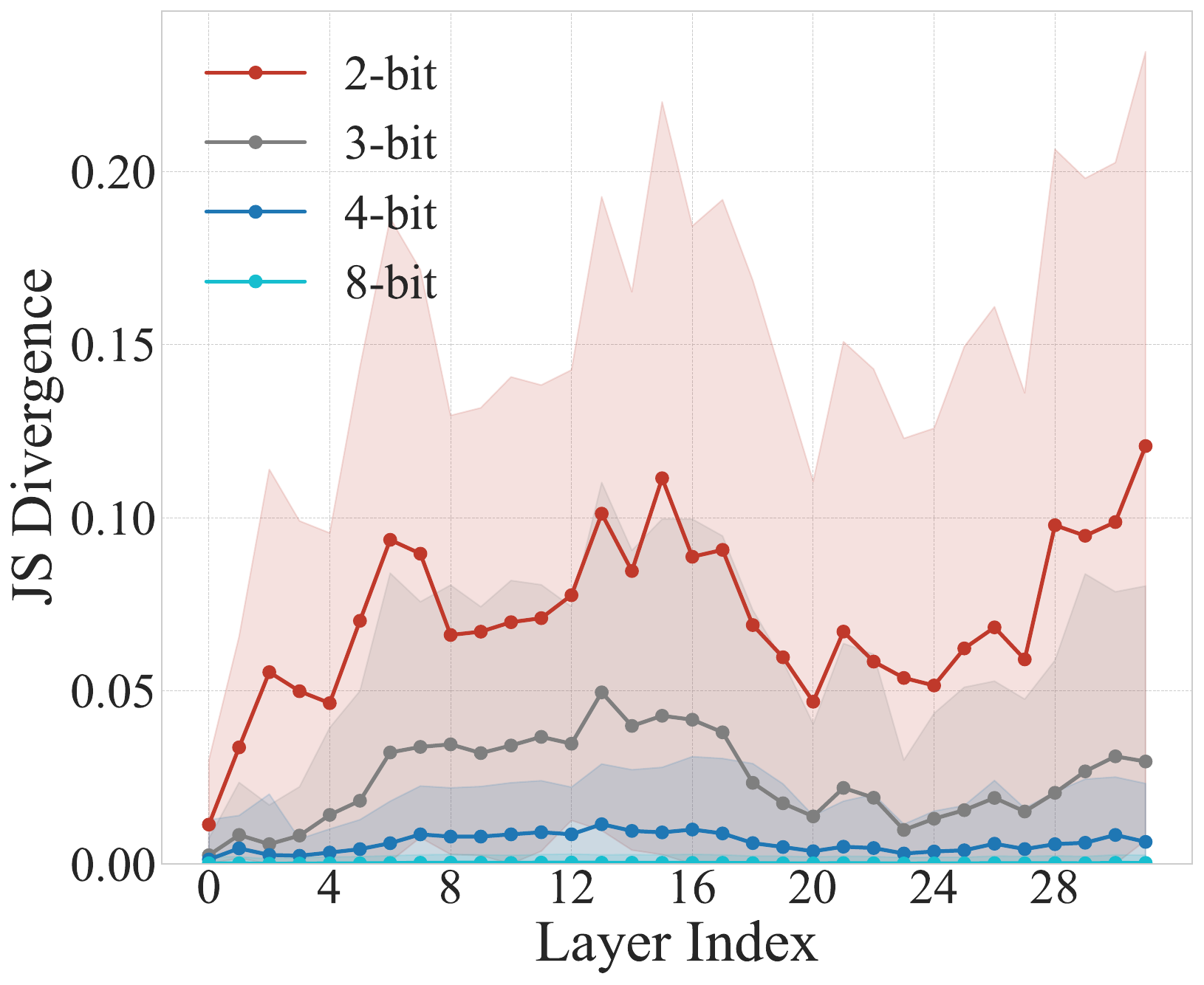} 
        \caption{Last subject token on the Robust Subset}
        \label{fig:app_jsd_correct}
    \end{subfigure}
    \hspace{0.02\textwidth}
    \begin{subfigure}[b]{0.45\textwidth}
        \centering
        \includegraphics[width=\textwidth]{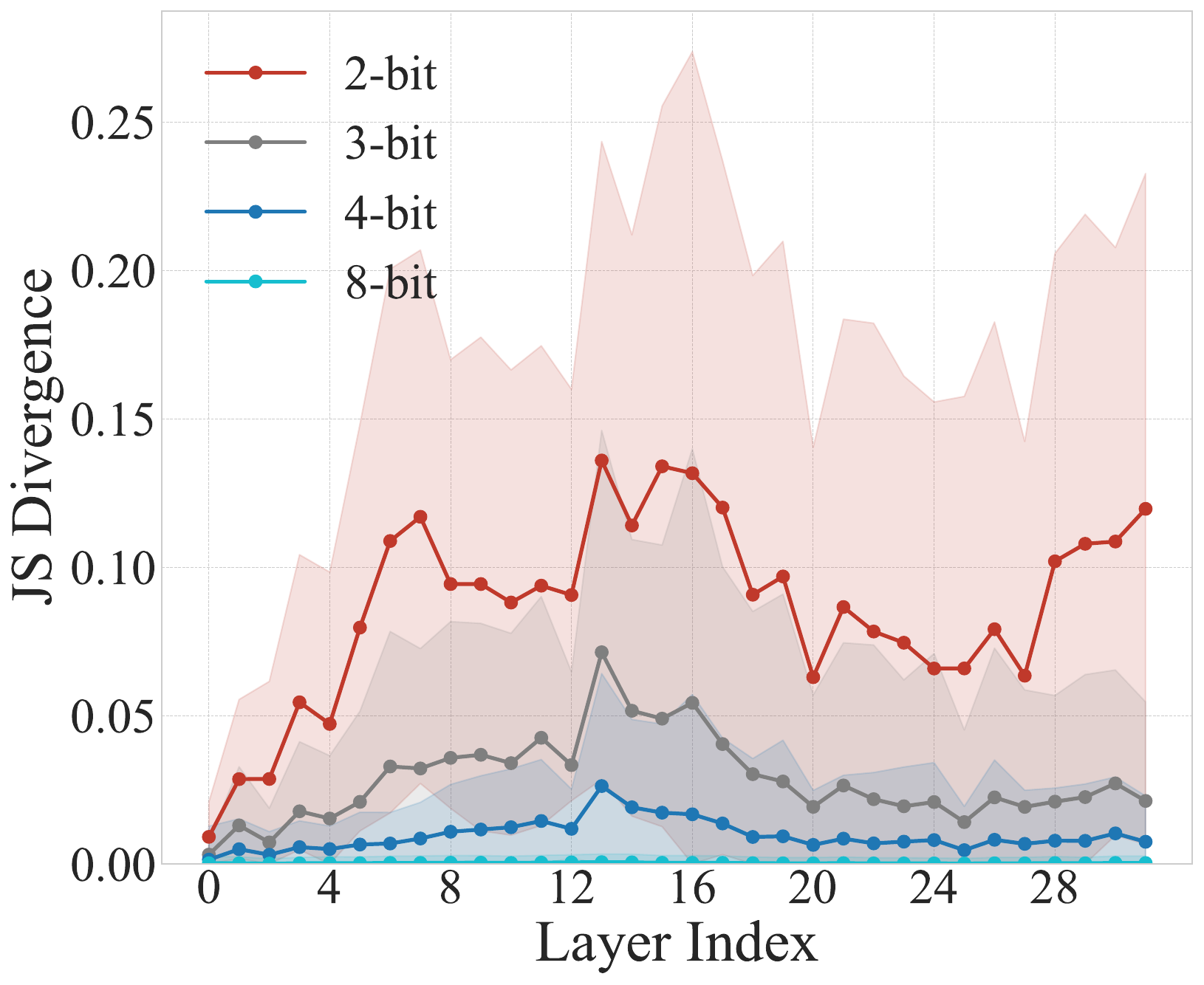} 
        \caption{Last token on the Failure Subset}
        \label{fig:app_jsd_last}
    \end{subfigure}
    \caption{Supplementary JSD Analysis. (a) 4-bit models maintain high alignment on the Robust Subset, while 2-bit models show instability. (b) Divergence persists at the last token.}
    \label{fig:app_attn_jsd}
\end{figure*}

\paragraph{Attention Pattern (Entropy \& JSD).} 
Figures~\ref{fig:app_attn_entropy} and \ref{fig:app_attn_jsd} confirm that the high uncertainty and attention divergence in 2-bit models are universal across datasets and token positions. Notably, while 4-bit models show tighter alignment on the Robust Subset (Fig.~\ref{fig:app_attn_jsd}a), 2-bit models consistently exhibit significant divergence.

\begin{figure*}[t]
  \centering
  \begin{subfigure}[b]{0.32\textwidth}
    \centering
    \includegraphics[width=\textwidth]{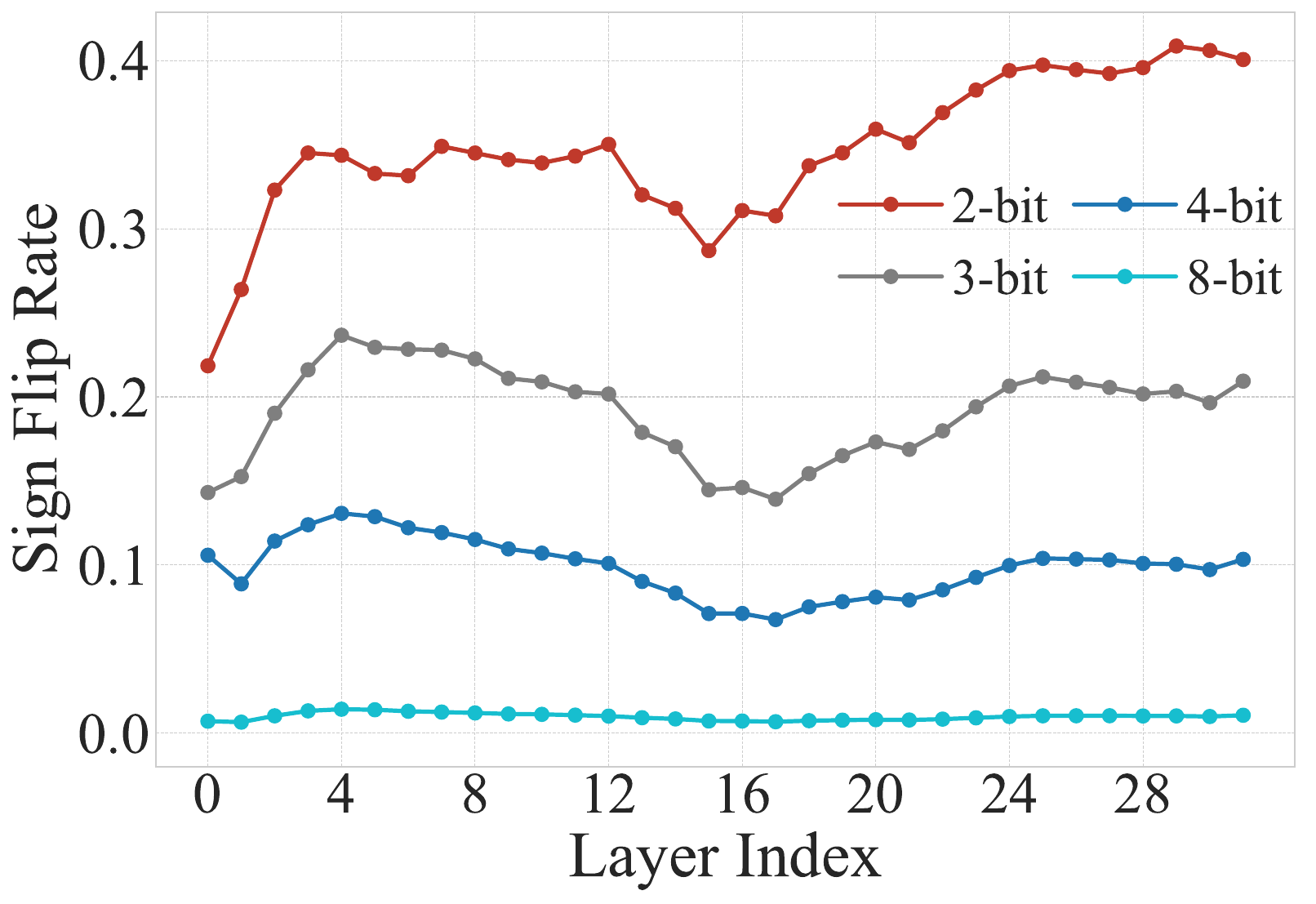} 
    \caption{Gate Sign Flip Rate}
    \label{fig:app_sfr_correct}
  \end{subfigure}
  \hfill
  \begin{subfigure}[b]{0.32\textwidth}
    \centering
    \includegraphics[width=\textwidth]{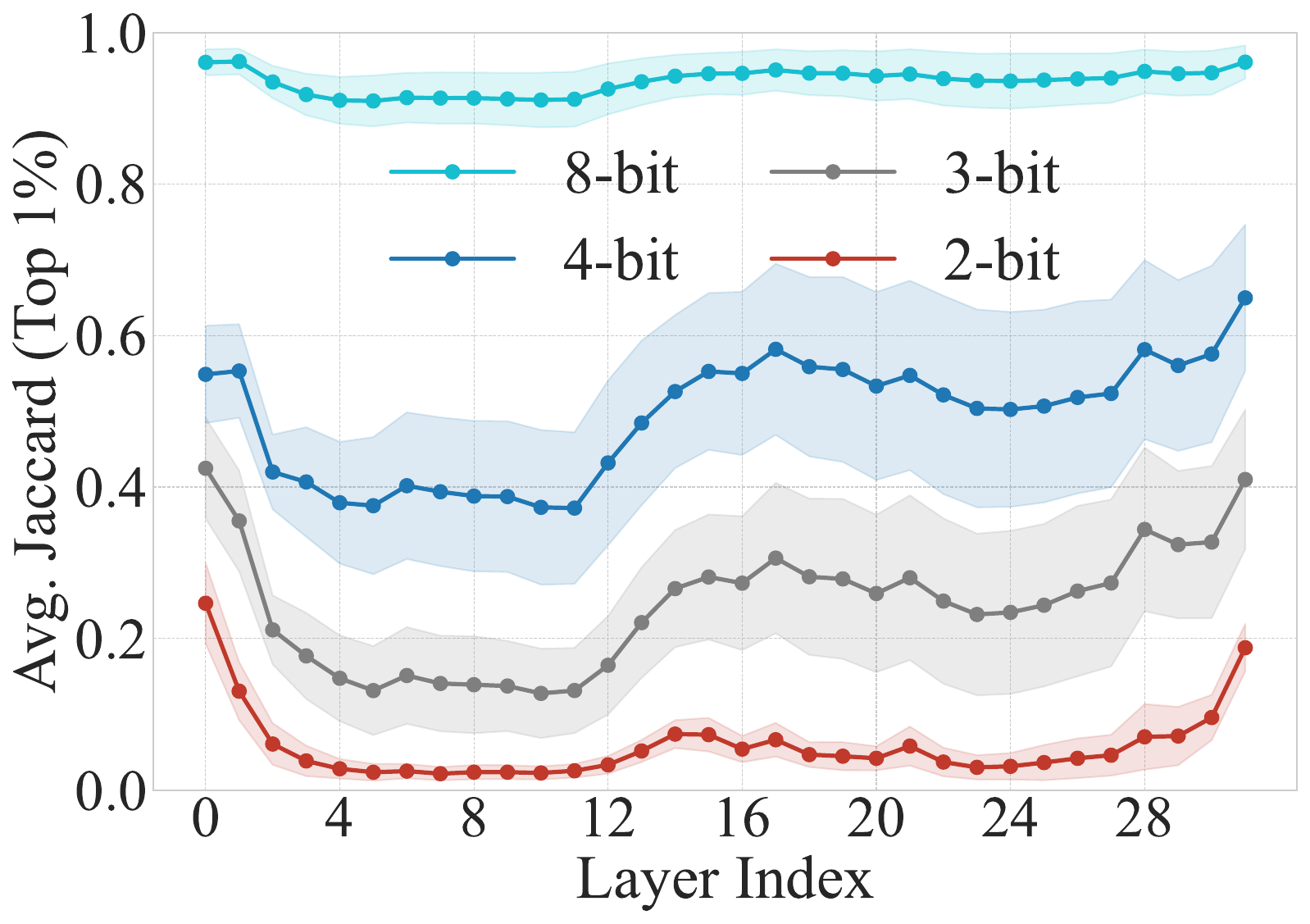} 
    \caption{Expert Jaccard Similarity}
    \label{fig:app_jaccard_correct}
  \end{subfigure}
  \hfill
  \begin{subfigure}[b]{0.32\textwidth}
    \centering
    \includegraphics[width=\textwidth]{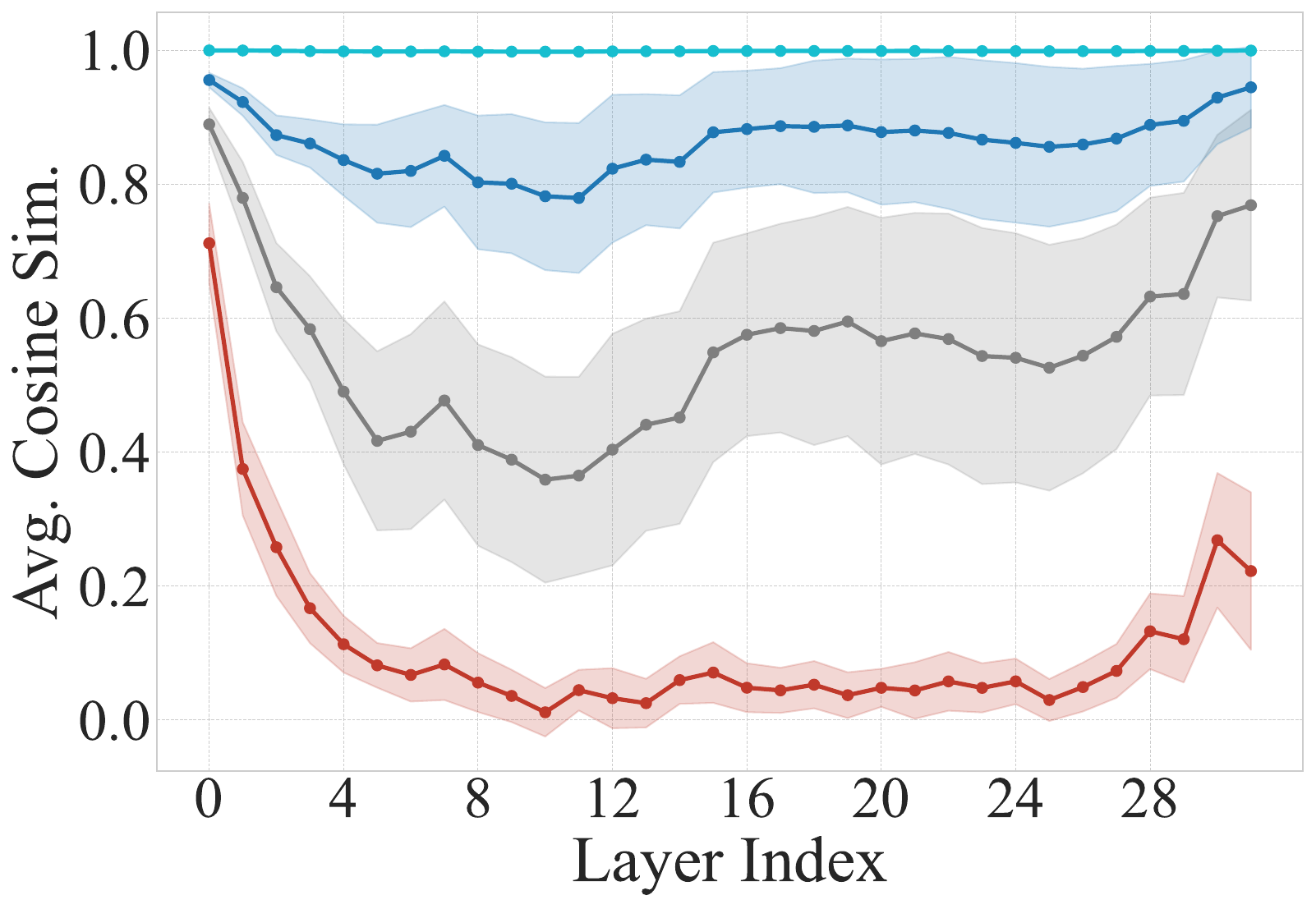} 
    \caption{Value Similarity (Cosine)}
    \label{fig:app_value_correct}
  \end{subfigure}
  \caption{Supplementary FFN Analysis on the Robust Subset at the last subject token. Even on easier samples, 2-bit models show internal instability (a, b) and output degradation (c), while 4-bit models remain healthy.}
  \label{fig:app_ffn_analysis_correct} 
\end{figure*}

\begin{figure*}[t]
  \centering
  \begin{subfigure}[b]{0.32\textwidth}
    \centering
    \includegraphics[width=\textwidth]{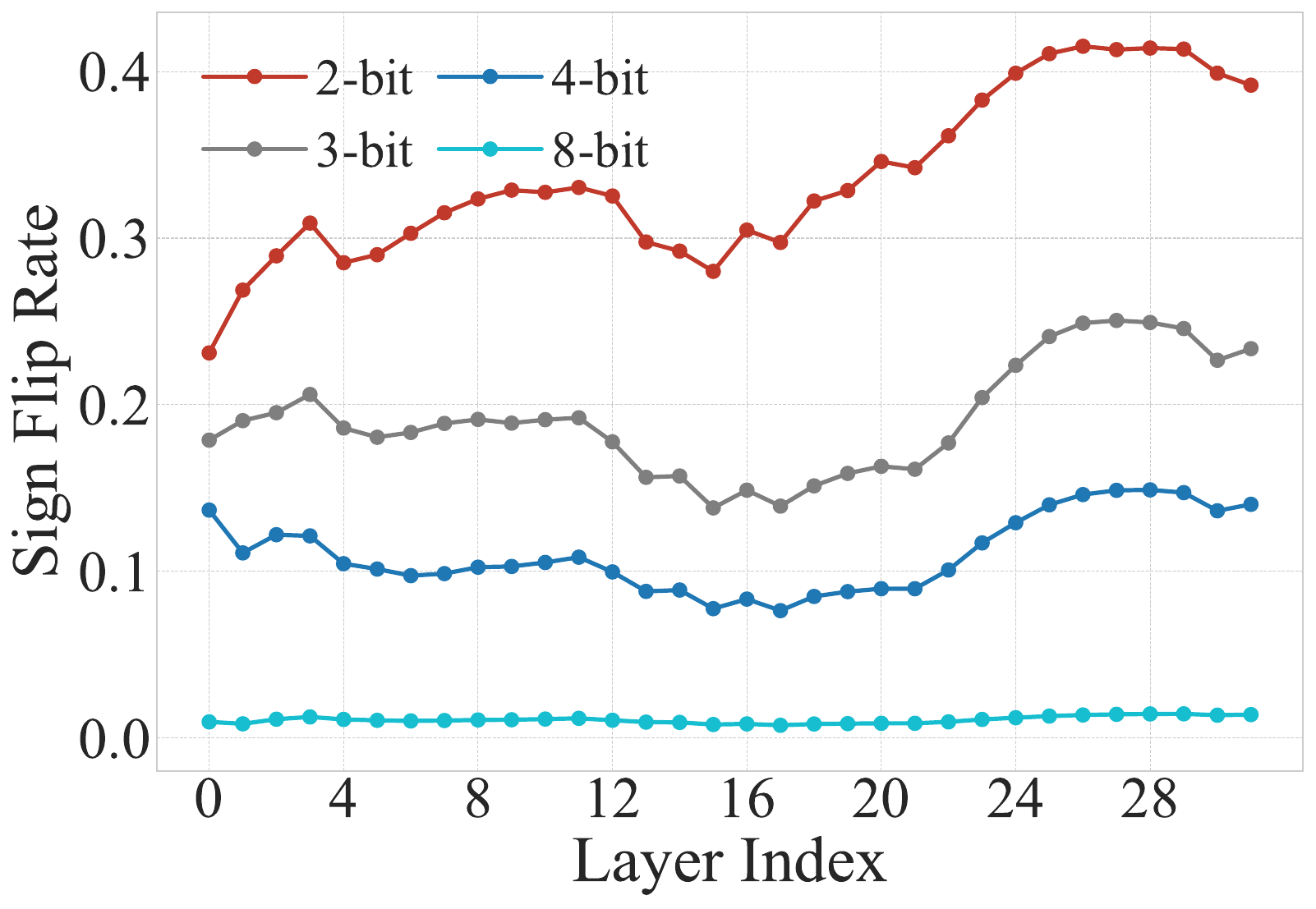} 
    \caption{Gate Sign Flip Rate}
    \label{fig:app_sfr_last}
  \end{subfigure}
  \hfill
  \begin{subfigure}[b]{0.32\textwidth}
    \centering
    \includegraphics[width=\textwidth]{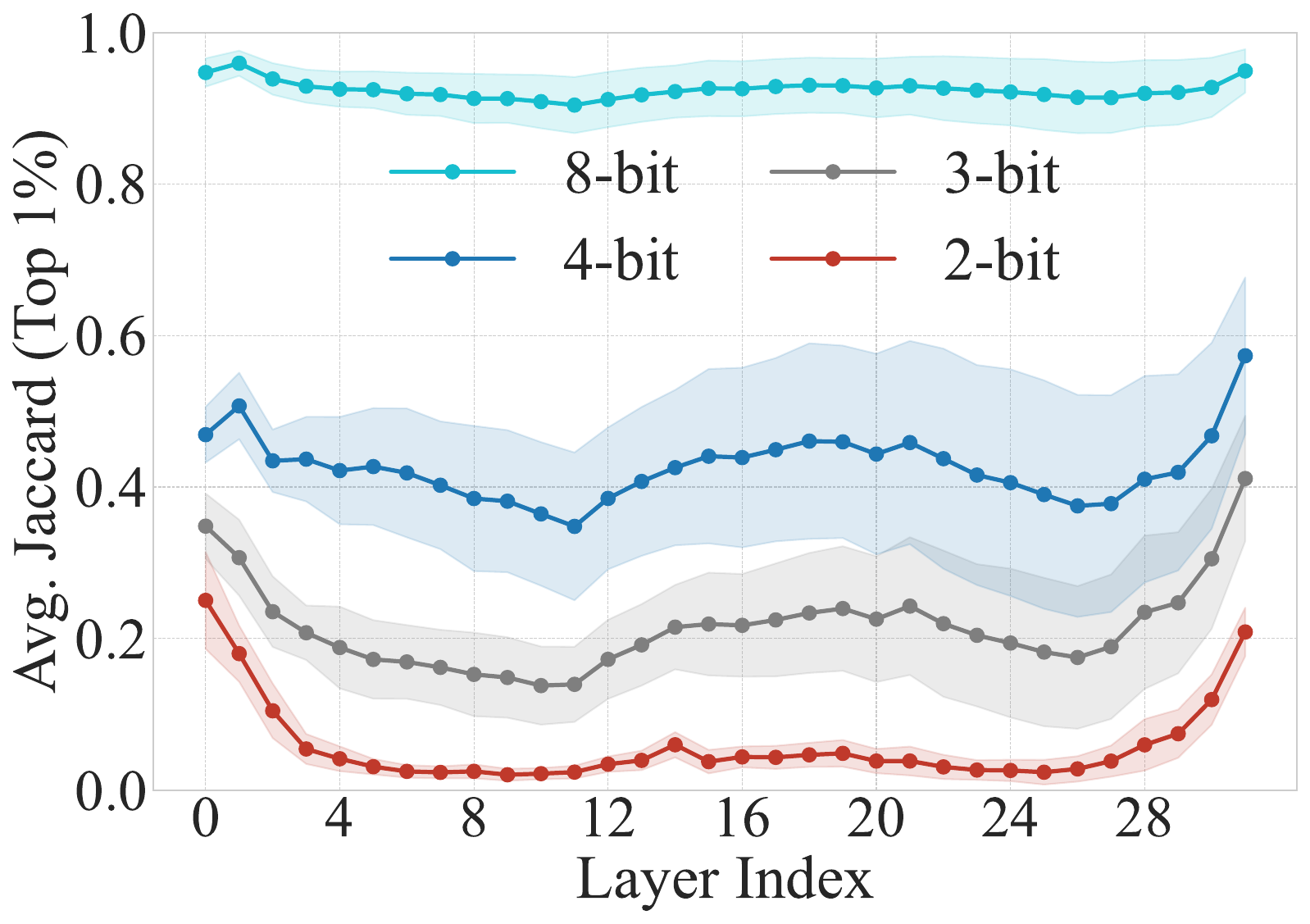} 
    \caption{Expert Jaccard Similarity}
    \label{fig:app_jaccard_last}
  \end{subfigure}
  \hfill
  \begin{subfigure}[b]{0.32\textwidth}
    \centering
    \includegraphics[width=\textwidth]{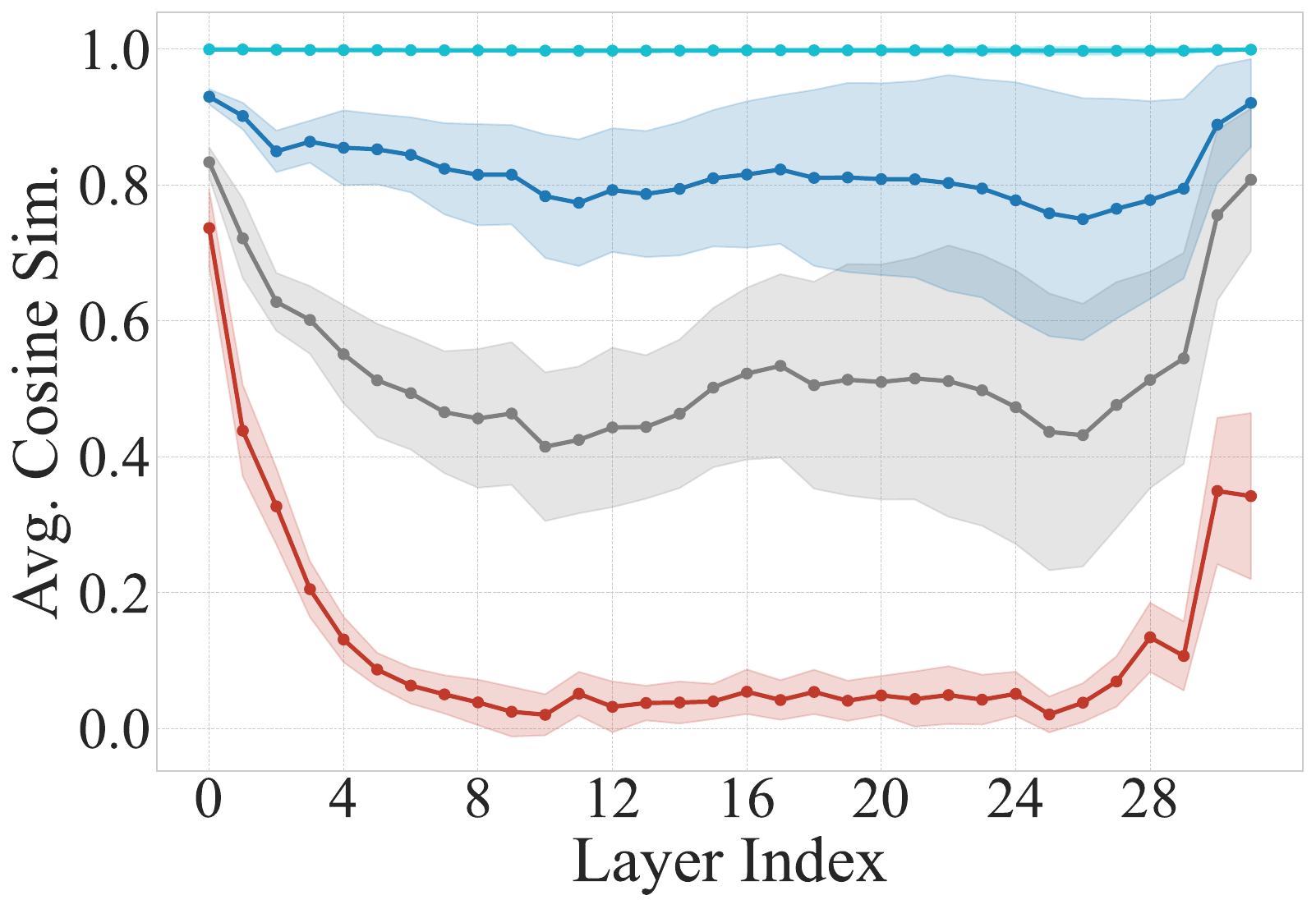} 
    \caption{Value Similarity (Cosine)}
    \label{fig:app_value_last}
  \end{subfigure}
  \caption{Supplementary FFN Analysis at the last token on the Failure Subset. The failure mode is consistent across positions: 2-bit causes gating collapse (a) and retrieval failure (b), destroying the final representation (c).}
  \label{fig:app_ffn_analysis_last} 
\end{figure*}

\paragraph{FFN Key-Value Memory.}
Figures~\ref{fig:app_ffn_analysis_correct} and \ref{fig:app_ffn_analysis_last} confirm that the collapse in 2-bit is universal, regardless of task difficulty or token position. 
Specifically, 2-bit models consistently exhibit extreme sign flip rates and near-zero Jaccard scores (Panels a \& b), indicating a complete breakdown in expert selection. This leads to a semantic collapse in the Value outputs (Panel c), where similarity drops to near-zero. In contrast, 4-bit models maintain strong alignment, exhibiting higher mean similarity and lower variance compared to the difficult subset.

\subsection{Representational Topology}
\label{sec:appendix_representation}

\begin{figure*}[!t]
    \centering
    \begin{subfigure}[b]{0.99\textwidth}
        \centering
        \includegraphics[width=\textwidth]{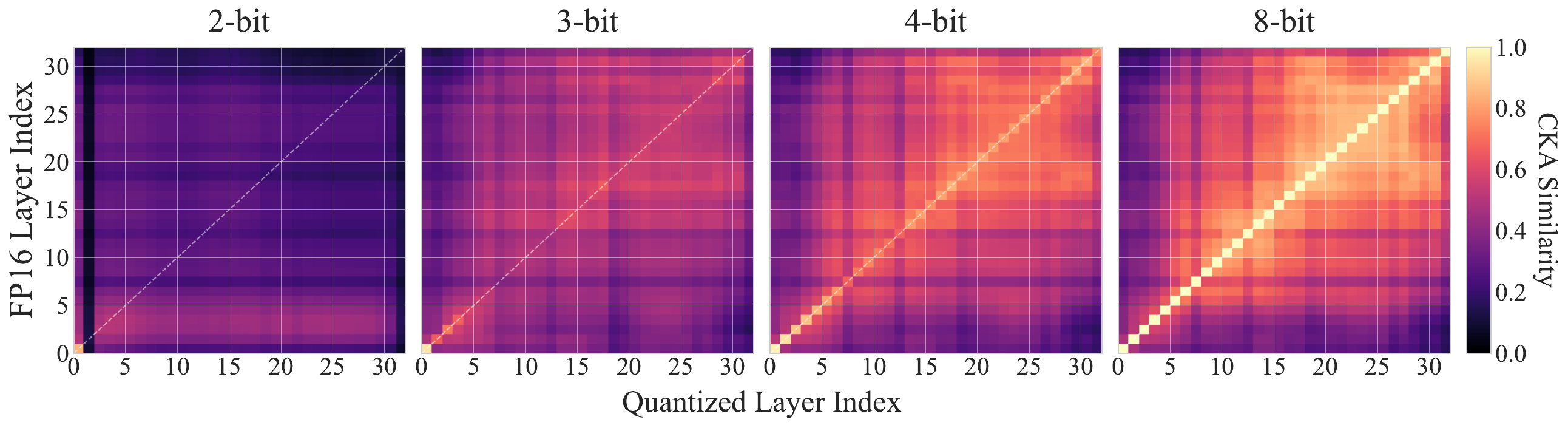}
        \caption{FFN Output (Last Subject Token)}
        \label{fig:app_cka_ffn_subject}
    \end{subfigure}    
    \begin{subfigure}[b]{0.99\textwidth}
        \centering
        \includegraphics[width=\textwidth]{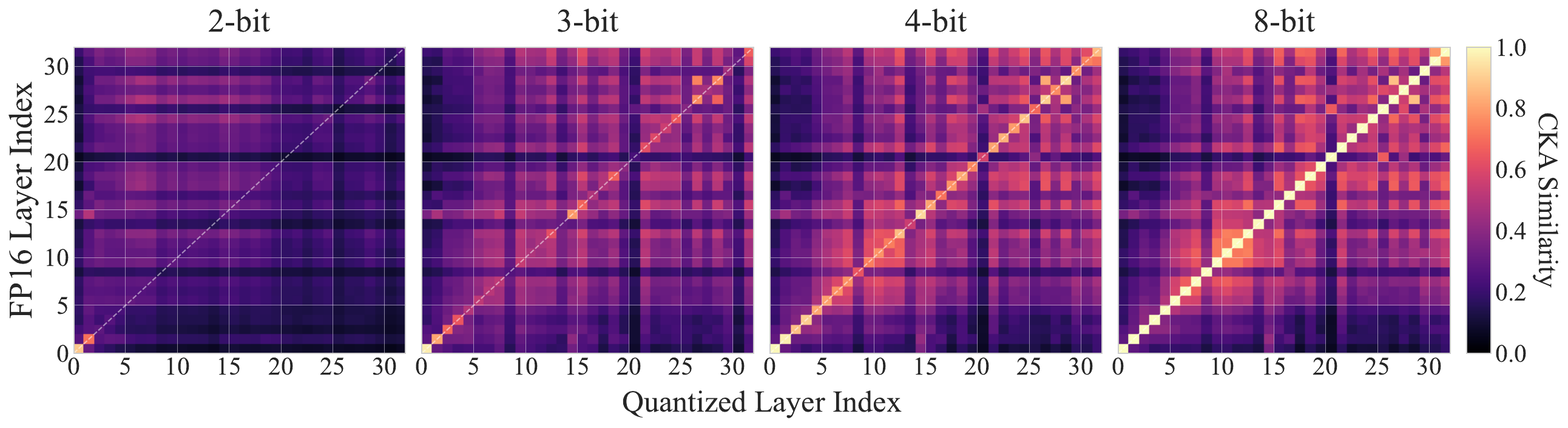}
        \caption{Attention Output (Last Subject Token)}
        \label{fig:app_cka_attn_subject}
    \end{subfigure}
    \caption{Component-wise CKA Analysis at the last subject token. The figures are stacked vertically to show the detail of FFN and Attention collapse in 2-bit models.}
    \label{fig:app_cka_subject_components}
\end{figure*}

\begin{figure*}[!t]
    \centering
    \begin{subfigure}[b]{0.99\textwidth}
        \centering
        \includegraphics[width=\textwidth]{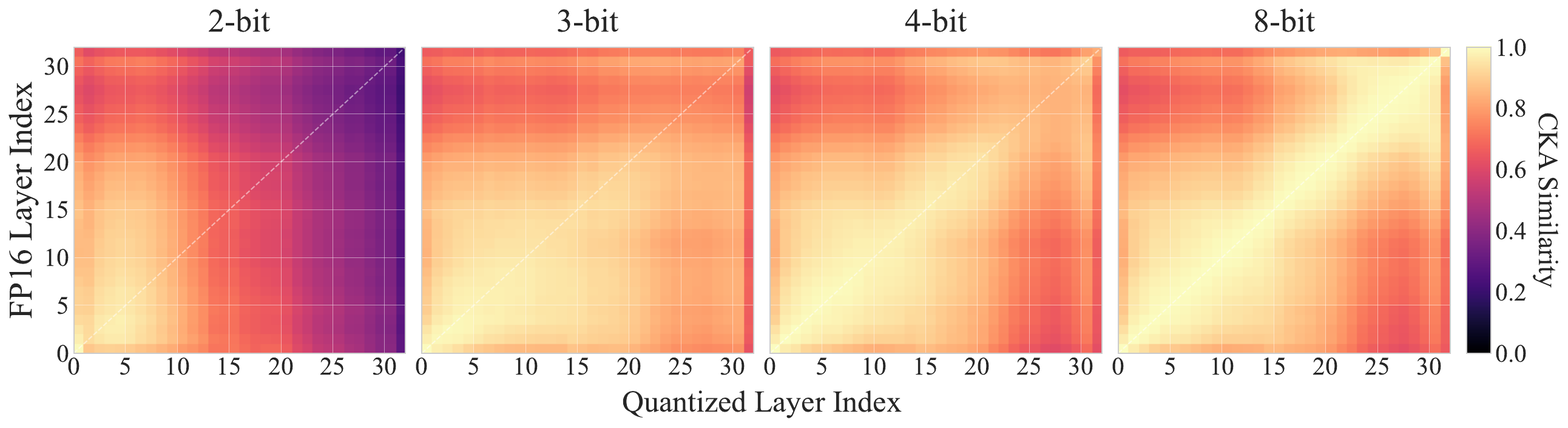}
        \caption{Layer Output (Last Token)}
        \label{fig:app_cka_layer_last}
    \end{subfigure}
    \begin{subfigure}[b]{0.99\textwidth}
        \centering
        \includegraphics[width=\textwidth]{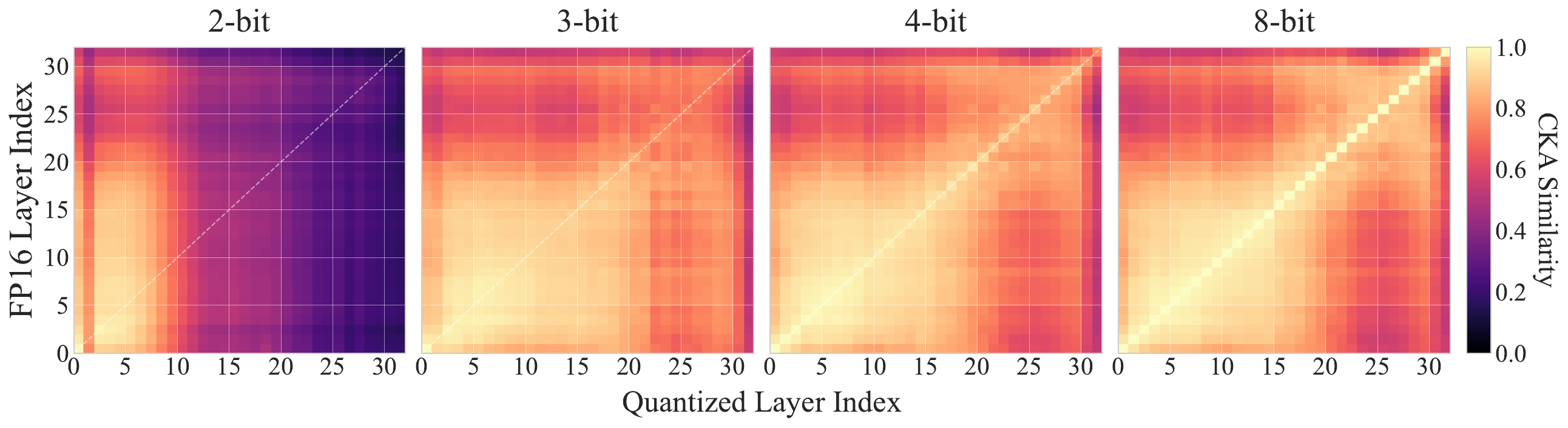}
        \caption{FFN Output (Last Token)}
        \label{fig:app_cka_ffn_last}
    \end{subfigure}
    \begin{subfigure}[b]{0.99\textwidth}
        \centering
        \includegraphics[width=\textwidth]{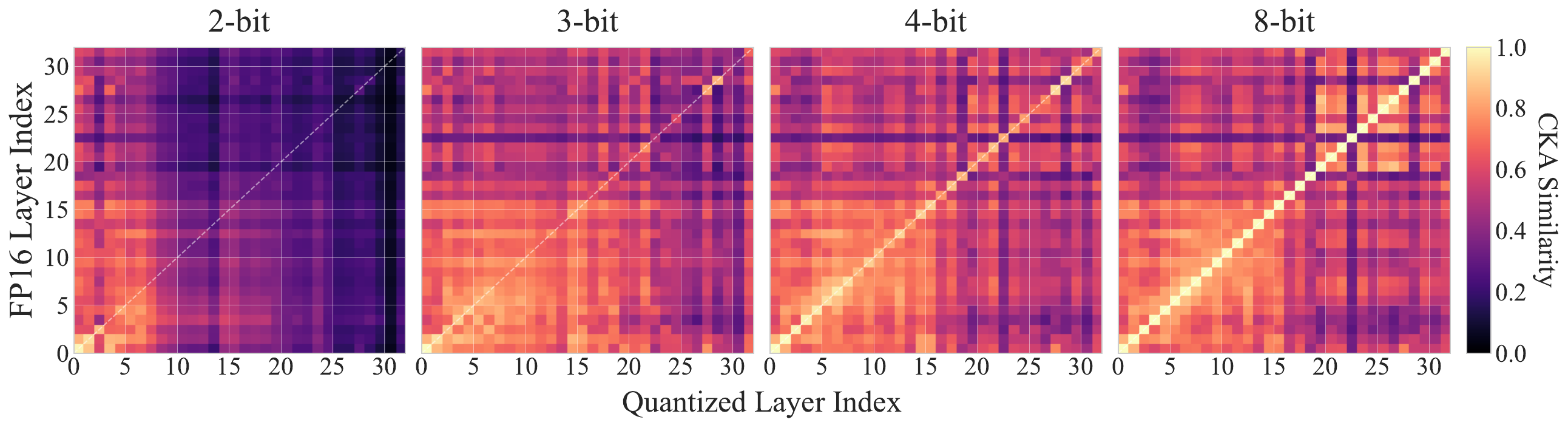}
        \caption{Attention Output (Last Token)}
        \label{fig:app_cka_attn_last}
    \end{subfigure}
    \caption{CKA Analysis at the last token. The topological collapse is consistent across all components.}
    \label{fig:app_cka_last_token}
\end{figure*}

\paragraph{CKA (Components \& Position).}
We expand the CKA analysis to specific components and different token positions.
Figure~\ref{fig:app_cka_subject_components} analyzes the internal components at the last subject token. It shows that while the layer output retains some structure due to residual connections, the internal components of the 2-bit model are completely collapsed (pitch black).
Figure~\ref{fig:app_cka_last_token} repeats the analysis at the last token. The trend remains identical: 4-bit models preserve the high-correlation block structure, while 2-bit models lose structural coherence.

\paragraph{Semantic Direction (Cosine Similarity).}

While the main text analyzes the internal structure at the subject token, here we utilize cosine similarity at the last token to verify the ultimate output of the representation.

\begin{figure*}[!t]
    \centering
    \begin{subfigure}[b]{0.48\textwidth}
        \centering
        \includegraphics[width=\textwidth]{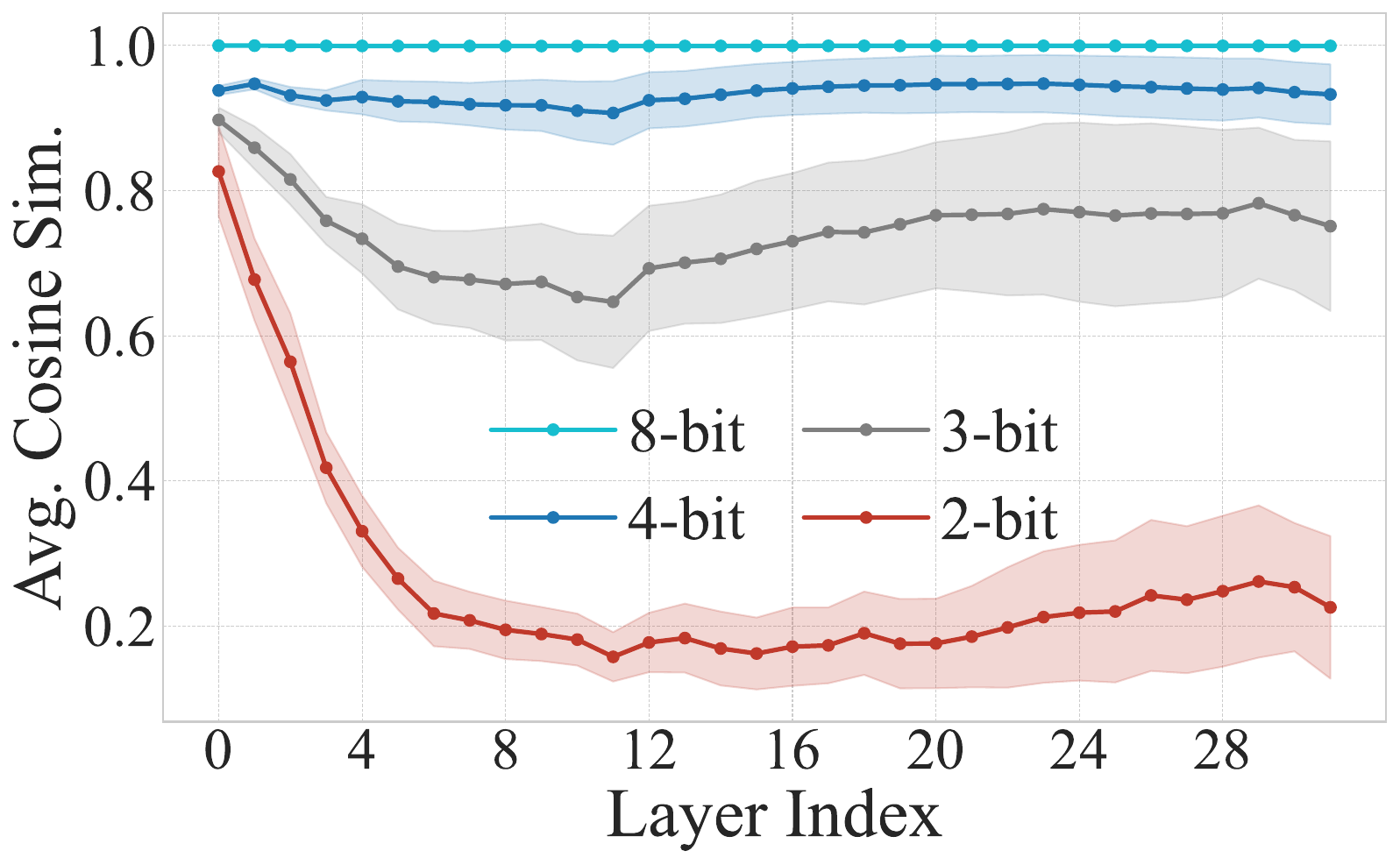} 
        \caption{Layer output on the Robust Subset}
        \label{fig:app_rep_cosine_correct}
    \end{subfigure}
    \hfill
    \begin{subfigure}[b]{0.48\textwidth}
        \centering
        \includegraphics[width=\textwidth]{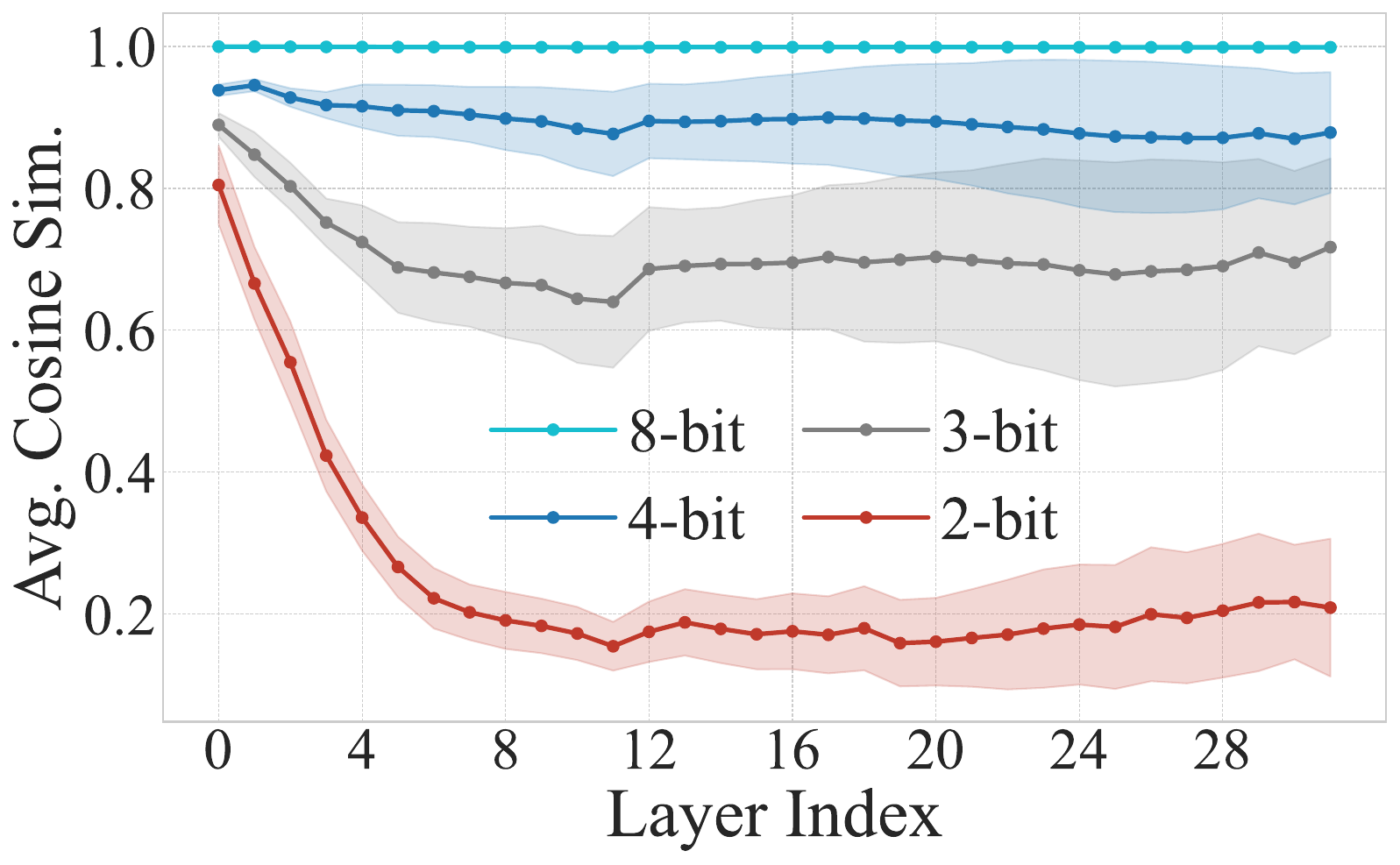} 
        \caption{Layer output on the Failure Subset}
        \label{fig:app_rep_cosine_wrong}
    \end{subfigure}
    \caption{Supplementary Cosine Similarity Analysis at the last token. Comparisons show that while 2-bit models collapse universally, 4-bit models only suffer from instability on difficult samples.}
    \label{fig:app_rep_cosine}
\end{figure*}

Figure~\ref{fig:app_rep_cosine} compares the layer output similarity. The 2-bit model suffers a complete collapse, with similarity dropping to near-zero.
The 4-bit model maintains high alignment. However, on the failure subset (Fig.~\ref{fig:app_rep_cosine_wrong}), it shows larger variance compared to the success subset (Fig.~\ref{fig:app_rep_cosine_correct}). This suggests 4-bit failures come from noise instability rather than directional error.

\paragraph{SVD Analysis.}
Figure~\ref{fig:app_svd_analysis_correct} presents the comparative SVD analysis on the Robust Subset to verify the consistency of our findings on easier samples.

\begin{figure*}[!t]
    \centering
    \begin{subfigure}[b]{0.45\textwidth}
        \centering
        \includegraphics[width=\textwidth]{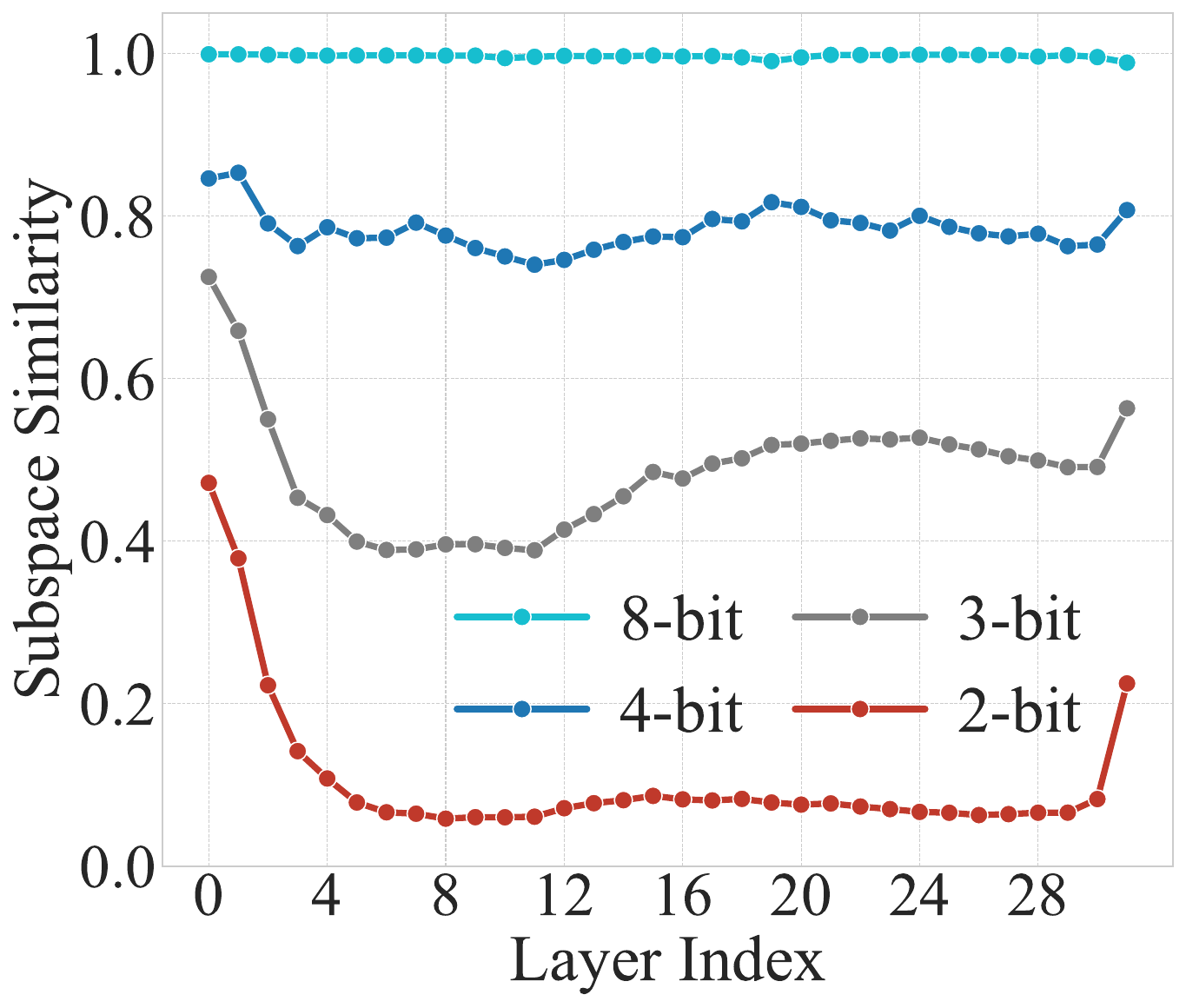}
        \caption{Activation Subspace Alignment}
        \label{fig:app_svd_activation_correct}
    \end{subfigure}
    \hspace{0.02\textwidth}
    \begin{subfigure}[b]{0.45\textwidth}
        \centering
        \includegraphics[width=\textwidth]{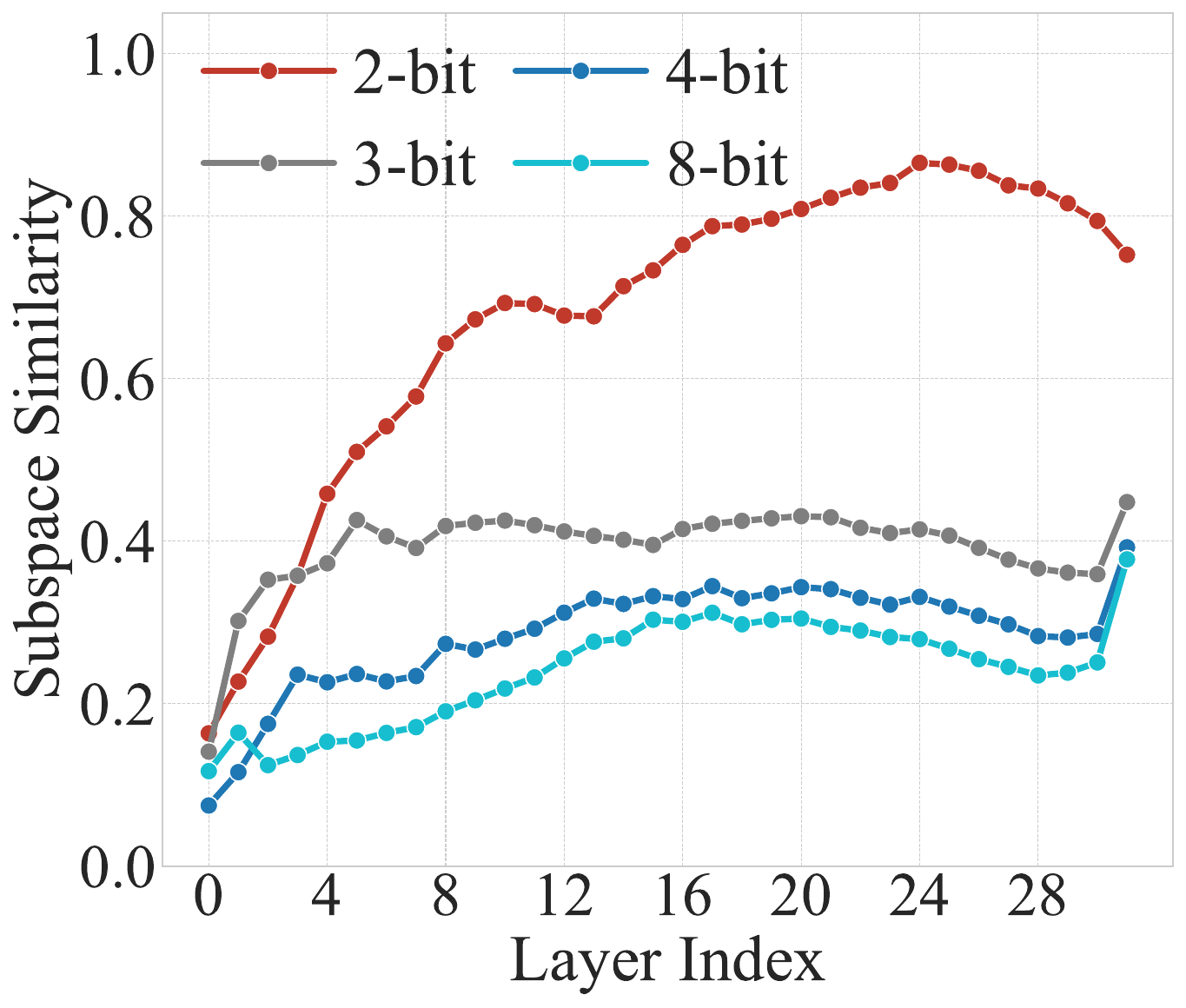} 
        \caption{Error-Signal Alignment}
        \label{fig:app_svd_error_correct}
    \end{subfigure}
    \caption{Supplementary SVD analysis on the Robust Subset. (a) 4-bit models match FP16. (b) 2-bit error remains destructive (high overlap with signal) even on easier samples.}
    \label{fig:app_svd_analysis_correct}
\end{figure*}

\begin{figure*}[!t]
    \centering
    \begin{subfigure}[b]{0.45\textwidth}
        \centering
        \includegraphics[width=\textwidth]{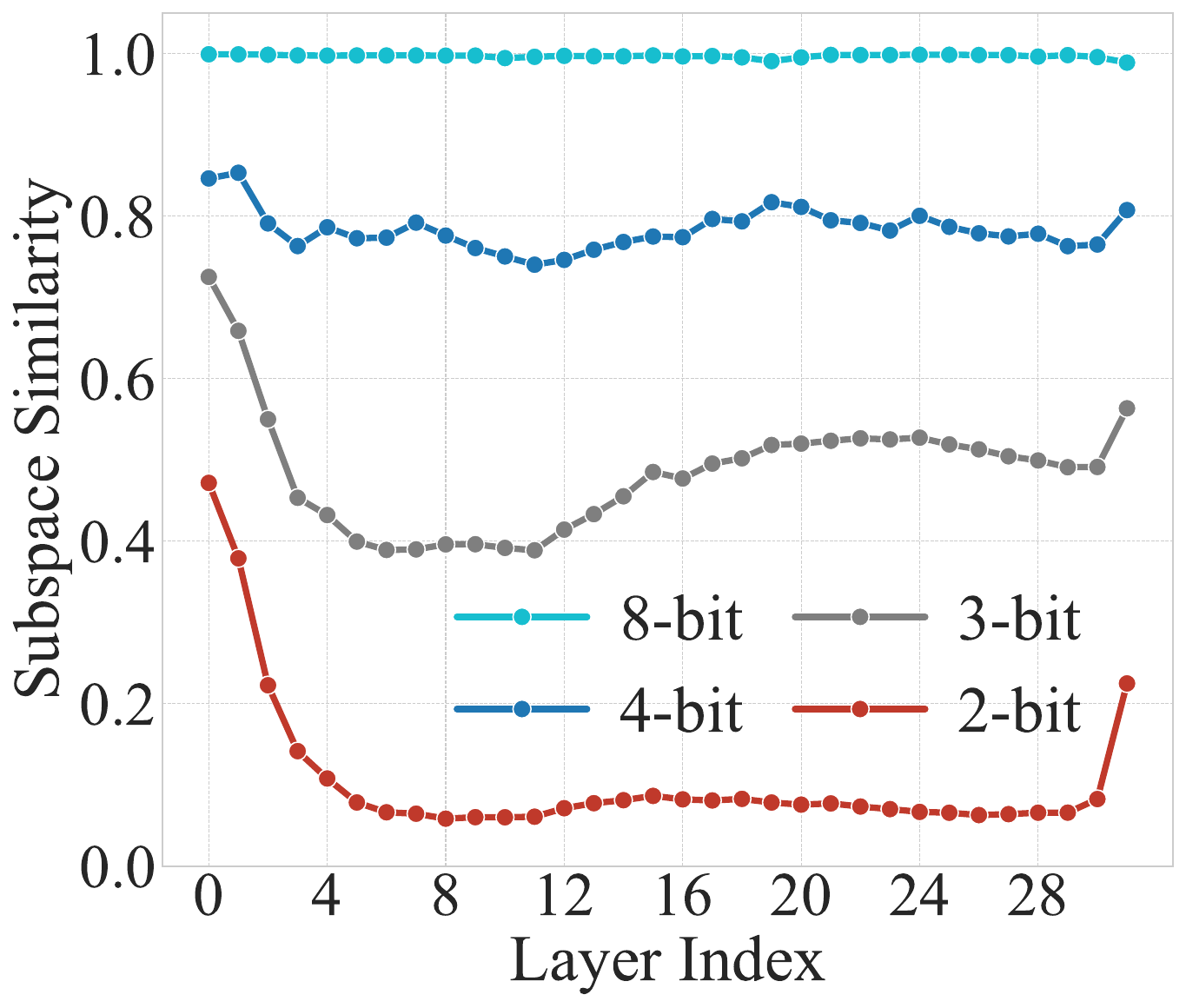} %
        \caption{Activation Subspace Alignment}
        \label{fig:app_svd_activation_correct}
    \end{subfigure}
    \hspace{0.02\textwidth}
    \begin{subfigure}[b]{0.45\textwidth}
        \centering
        \includegraphics[width=\textwidth]{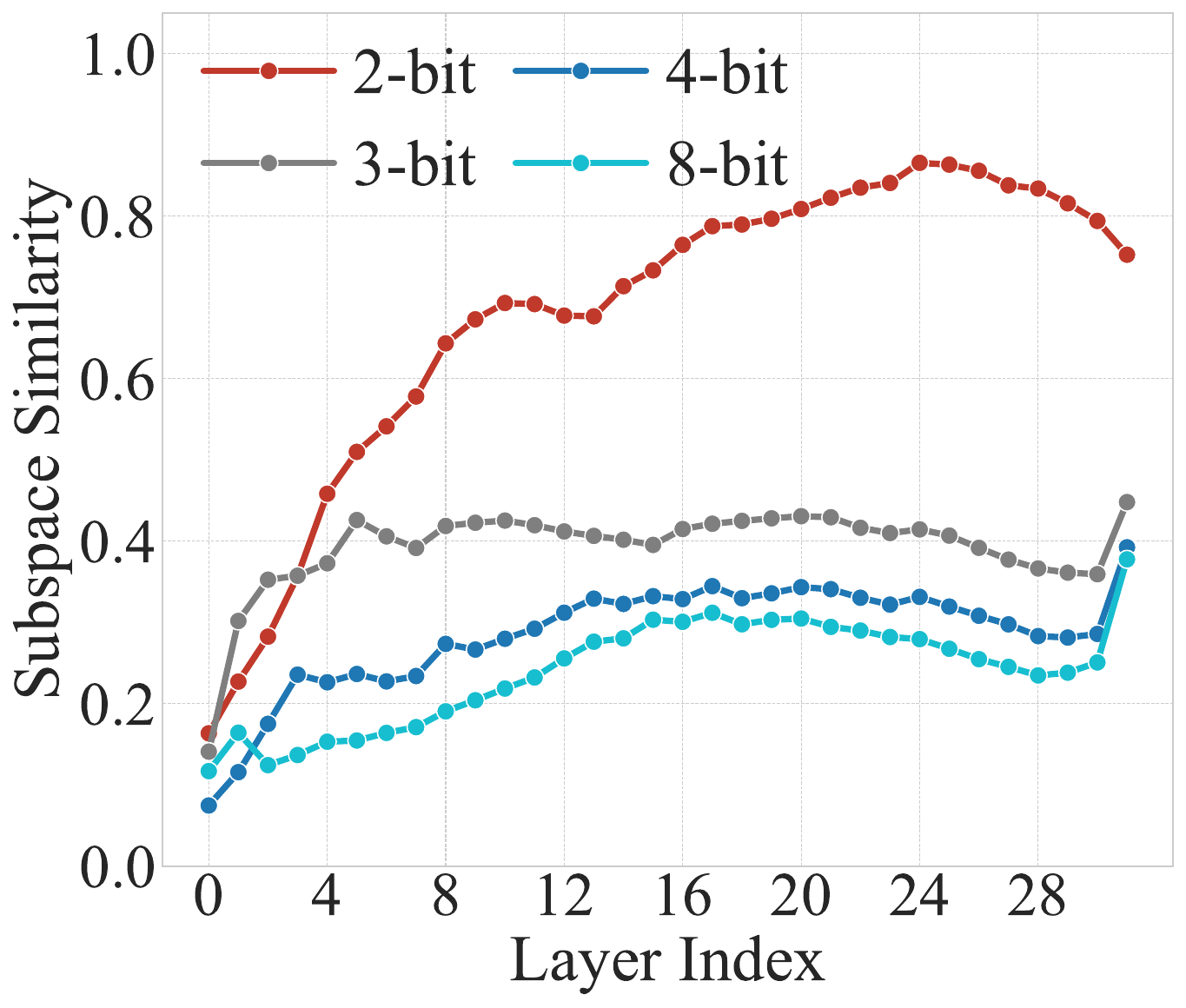} %
        \caption{Error-Signal Subspace Alignment}
        \label{fig:app_svd_error_correct}
    \end{subfigure}
    \caption{Supplementary SVD analysis on the Robust Subset. (a) Activation subspace alignment remains high for 4-bit, similar to FP16. (b) Error-signal alignment for 4-bit remains low, while 2-bit error remains highly aligned (destructive).}
    \label{fig:app_svd_analysis_correct}
\end{figure*}

\section{Intervention and Sensitivity Analysis}
\label{sec:appendix_interventions}

\subsection{Localized Sensitivity in 4-bit Models}
\label{sec:appendix_interntion_4bit}

\begin{figure*}[!t]
    \centering
    \includegraphics[width=0.66\textwidth]{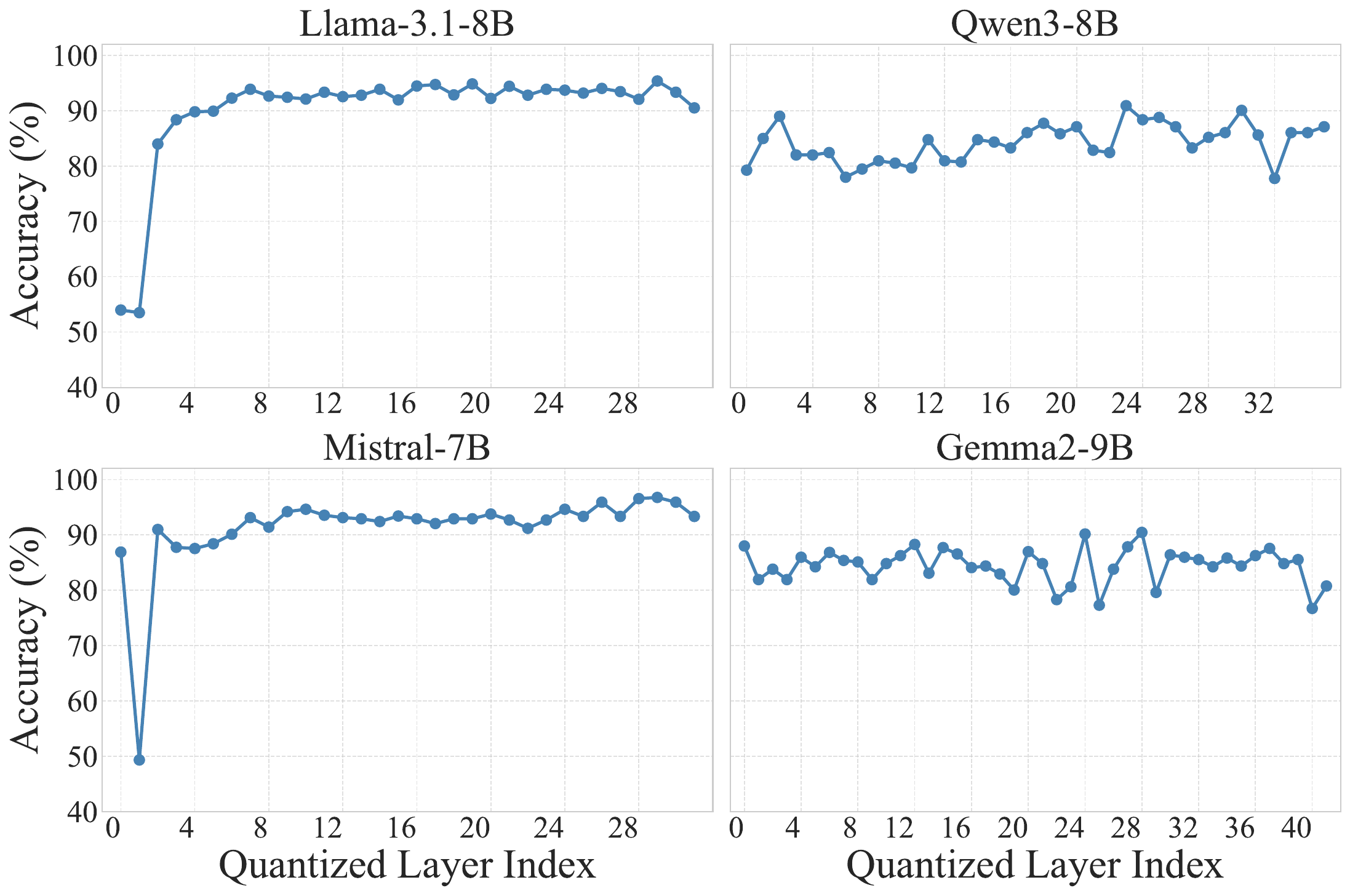}
    \caption{Single-layer 4-bit quantization sensitivity on the Failure Subset. Llama/Mistral show localized fragility, while Qwen/Gemma are balanced.}
    \label{fig:app_single_layer_4bit}
\end{figure*}

\paragraph{Layer-wise Sensitivity.}
Figure~\ref{fig:app_single_layer_4bit} complements the main text's ``domino'' analysis. We quantize only a single layer to 4-bit while keeping others in FP16. The results confirm the architecture-dependent sensitivity: Llama/Mistral show extreme sensitivity in early layers, while Qwen/Gemma show uniform sensitivity.

\paragraph{Component-wise Sensitivity.}
We analyze the sensitivity of individual components by quantizing them separately to 4-bit (Table~\ref{tab:component_sensitivity}).
\begin{itemize}[leftmargin=*]
    \item \textbf{Localized Vulnerability (Llama/Mistral)}: MLP modules are significantly more fragile than Attention modules. Specifically, the ``content generation'' weights (\texttt{down\_proj}, \texttt{v\_proj}) are far more critical than the ``routing'' weights.
    \item \textbf{Balanced Sensitivity (Qwen/Gemma)}: Degradation is uniform across MLP and Attention modules, with no single component acting as a distinct failure point.
\end{itemize}

Unlike 4-bit, where some modules remain functional, 2-bit quantization causes a universal failure. No module remains functionally robust, confirming that the failure is driven by a systemic breakdown of representational capacity rather than specific component weak points.

\begin{table*}[!t]
\centering
\small
\setlength{\tabcolsep}{7pt}
\begin{tabular}{@{}l|c|cc|cc|ccc@{}}
\toprule
\textbf{Model Family} & \textbf{All} & \textbf{MLP} & \textbf{Attn} & \textbf{MLP Gate/Up} & \textbf{MLP Down} & \textbf{QK Proj} & \textbf{V Proj} & \textbf{O Proj} \\
\midrule
Llama-3.1-8B-2bit & 0.00 & 0.00 & 9.40 & 4.82 & 4.16 & 48.91 & 21.12 & 41.45 \\
\midrule
Llama-3.1-8B-4bit & 0.00 & 38.04 & 53.88 & 70.65 & 49.57 & 89.70 & 54.44 & 88.61 \\
Mistral-7B-4bit   & 0.00 & 43.79 & 61.93 & 86.59 & 45.76 & 89.94 & 65.68 & 89.74 \\
Qwen3-8B-4bit     & 0.00 & 49.30 & 46.88 & 63.58 & 64.19 & 71.03 & 69.01 & 78.87 \\
Gemma-2-9B-4bit   & 0.00 & 54.13 & 52.33 & 50.65 & 70.03 & 76.10 & 61.11 & 81.78 \\
\bottomrule
\end{tabular}
\caption{Component-level sensitivity analysis on the Failure Subset. Values denote accuracy (\%) when only the specific component is quantized, highlighting the contrast between localized fragility (Llama/Mistral) and balanced sensitivity (Qwen/Gemma).}
\label{tab:component_sensitivity}
\end{table*}

\subsection{Systemic Collapse in 2-bit Models}
\label{sec:appendix_interntion_2bit}

\begin{figure*}[!t]
    \centering
    \includegraphics[width=0.66\textwidth]{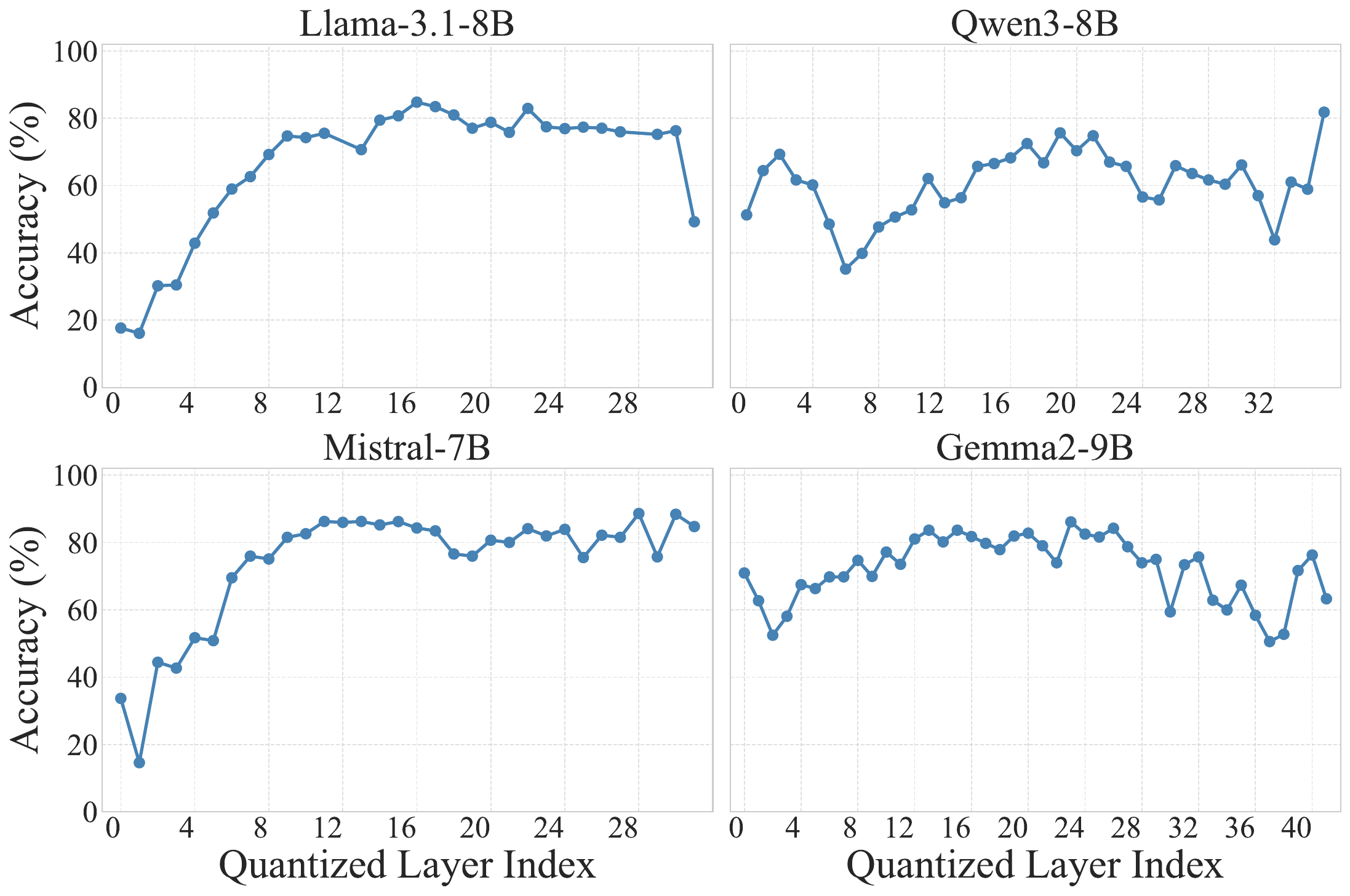}
    \caption{Single-layer 2-bit quantization sensitivity on the Failure Subset. Catastrophic drops from early layers (Llama/Mistral) are evident.}
    \label{fig:app_single_layer_2bit}
\end{figure*}

\begin{figure*}[!t]
    \centering
    \begin{subfigure}[b]{0.47\textwidth}
        \centering
        \includegraphics[width=0.95\textwidth]{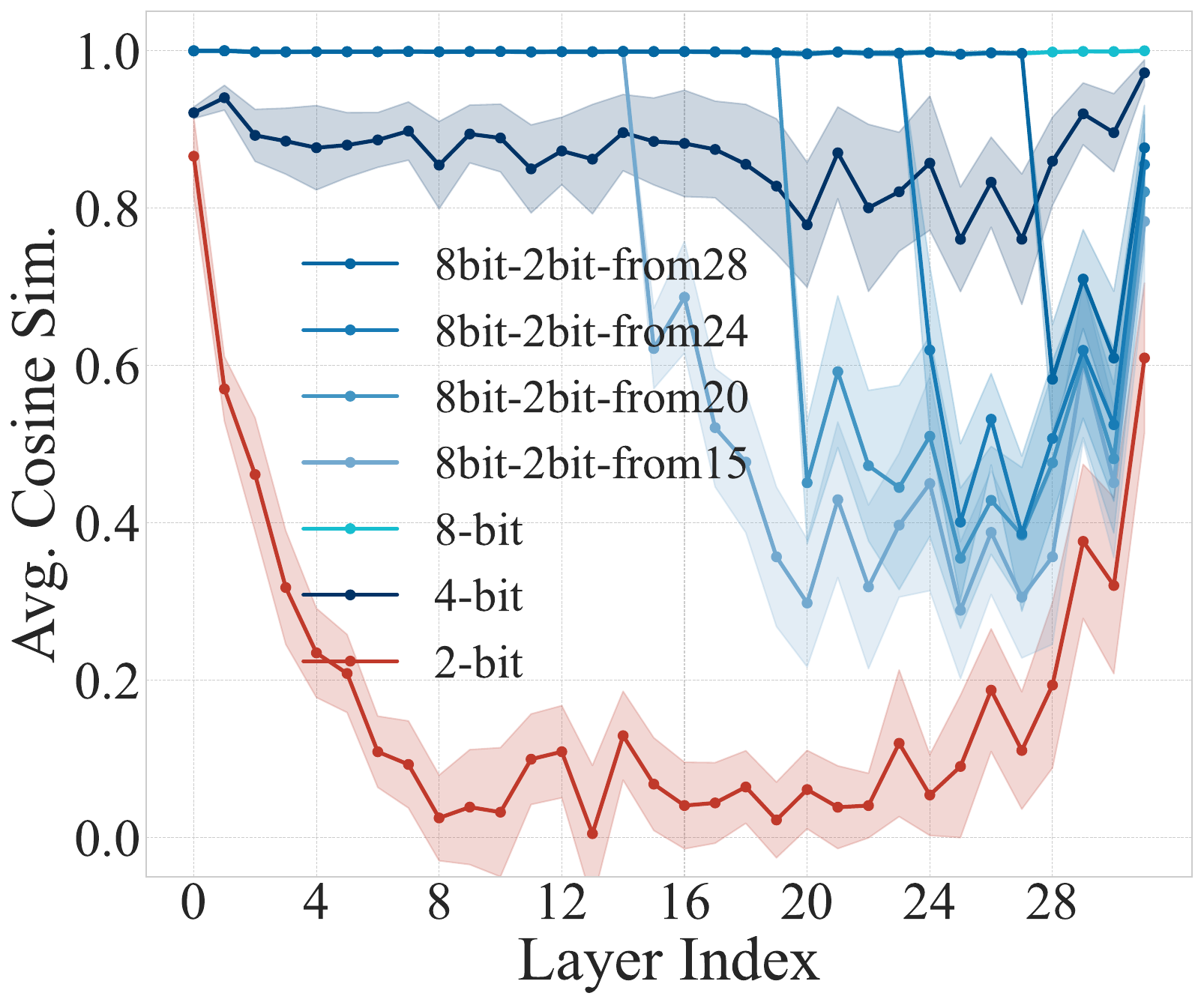}
        \caption{Attention Output}
    \end{subfigure}
    \hspace{0.02\textwidth}
    \begin{subfigure}[b]{0.47\textwidth}
        \centering
        \includegraphics[width=0.95\textwidth]{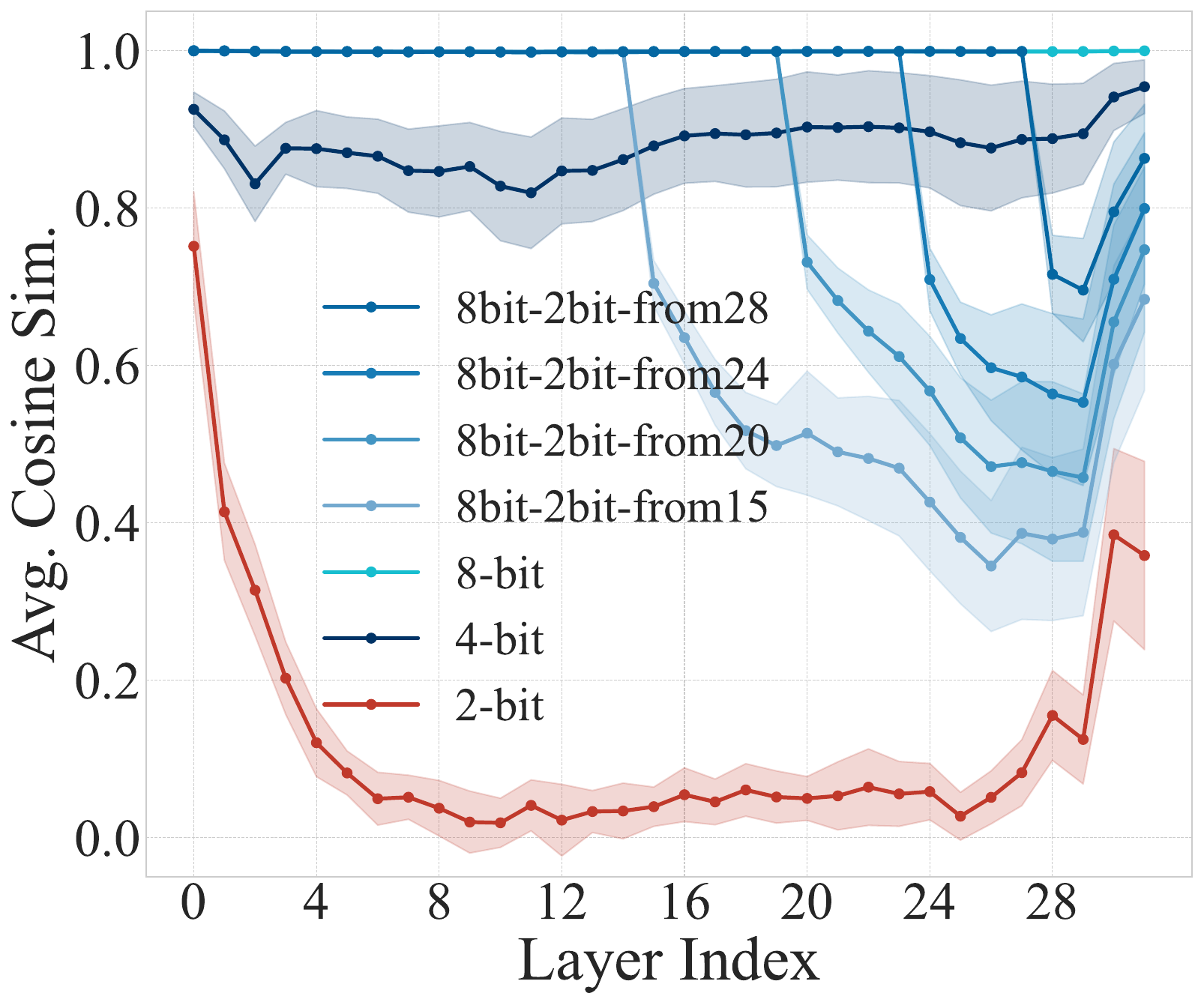}
        \caption{MLP Output}
    \end{subfigure}
    \caption{Component decomposition for high-precision signal injection on the Robust Subset. Both Attn and MLP outputs collapse upon entering 2-bit layers.}
    \label{fig:app_signal_protect_decomp}
\end{figure*}

Figure~\ref{fig:app_single_layer_2bit} shows single-layer 2-bit quantization results. Unlike 4-bit, quantizing even a single early layer (especially in Llama/Mistral) leads to catastrophic drops.
Figure~\ref{fig:app_signal_protect_decomp} decomposes the signal injection analysis. It confirms that the collapse observed in the main text (Figure~\ref{fig:signal_protect_test}) occurs simultaneously in both Attention and MLP outputs, proving the failure is systemic.

\clearpage
\onecolumn
\clearpage
\twocolumn
\section{Generalizability to AWQ Algorithm}
\label{sec:app_awq}

\makeatletter
\setlength{\@dblfptop}{0pt}       
\setlength{\@dblfpsep}{20pt}      
\setlength{\@dblfpbot}{0pt plus 1fil}
\makeatother

To verify whether our discovered failure modes generalize across quantization algorithms, we replicate the mechanistic analysis using AWQ~\cite{lin_mlsys_2024} on Llama-3.1-8B. We evaluate the models on the same Failure Subset. The macro-level accuracy strictly mirrors our GPTQ findings: AWQ 4-bit (28.17\%) $\rightarrow$ 3-bit (18.01\%) $\rightarrow$ 2-bit (0.00\%).

\subsection{Layer-wise Knowledge Probing}

\begin{figure*}[!t]
    \centering
    \begin{subfigure}[b]{0.42\textwidth}
        \centering
        \includegraphics[width=\textwidth]{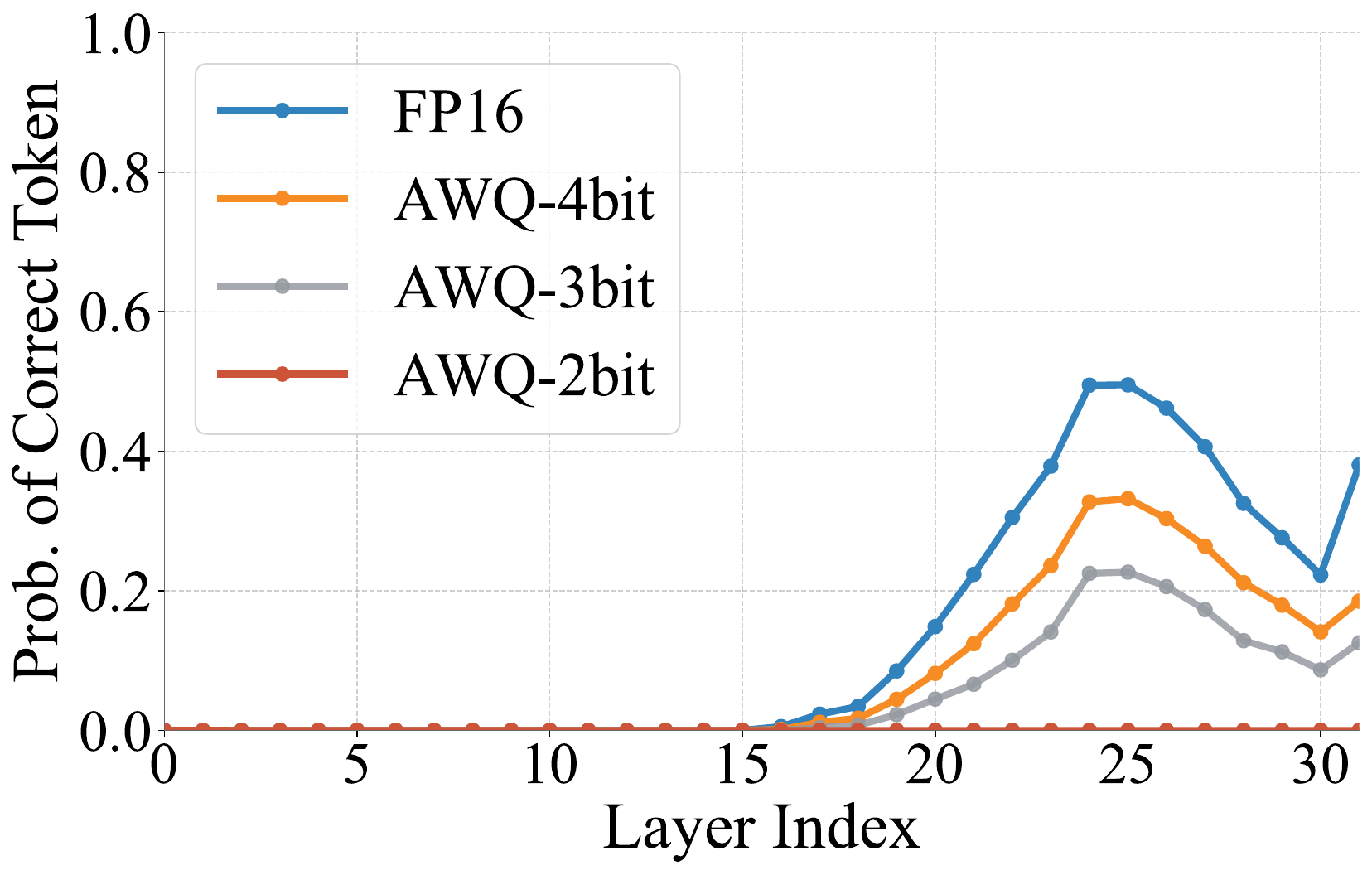}
        \caption{Probability Evolution}
    \end{subfigure}
    \hspace{0.03\textwidth}
    \begin{subfigure}[b]{0.42\textwidth}
        \centering
        \includegraphics[width=\textwidth]{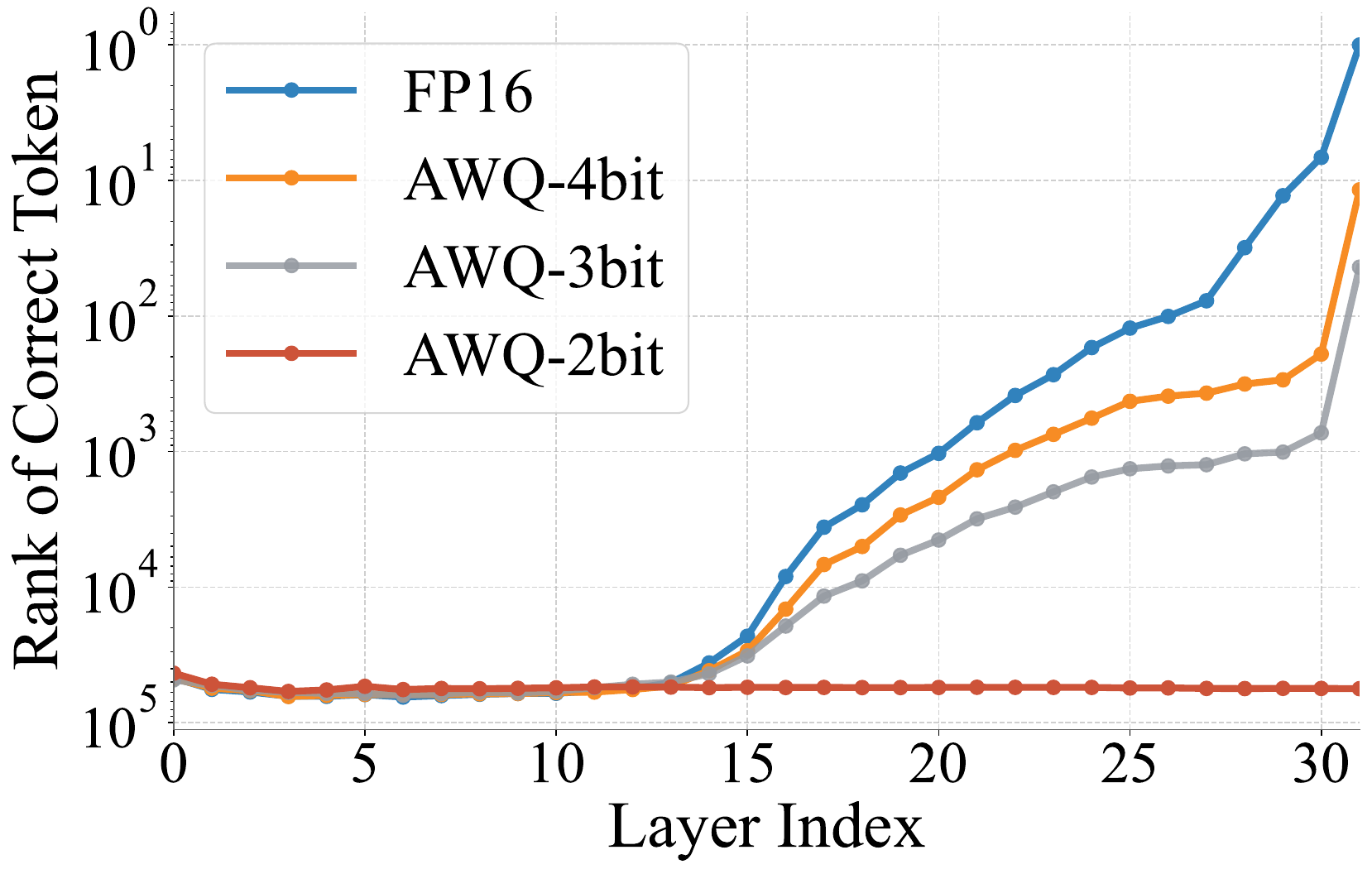}
        \caption{Rank Evolution}
    \end{subfigure}
    \caption{Layer-wise evolution of probability and rank for AWQ on the Failure Subset.}
    \label{fig:app_awq_logit_lens}
\end{figure*}

Figure~\ref{fig:app_awq_logit_lens} traces the layer-wise knowledge signals. Consistent with GPTQ, the 4-bit and 3-bit AWQ models exhibit Signal Degradation. Their target probabilities build up in deeper layers but remain lower than the FP16 baseline, accompanied by a moderate drop in target ranks. In contrast, the 2-bit model fails to recover any meaningful probability distribution, remaining completely flat at zero across all layers.

\subsection{Component-level Impairment}

\begin{figure*}[!t]
    \centering
    \begin{subfigure}[b]{0.43\textwidth}
        \centering
        \includegraphics[width=\textwidth]{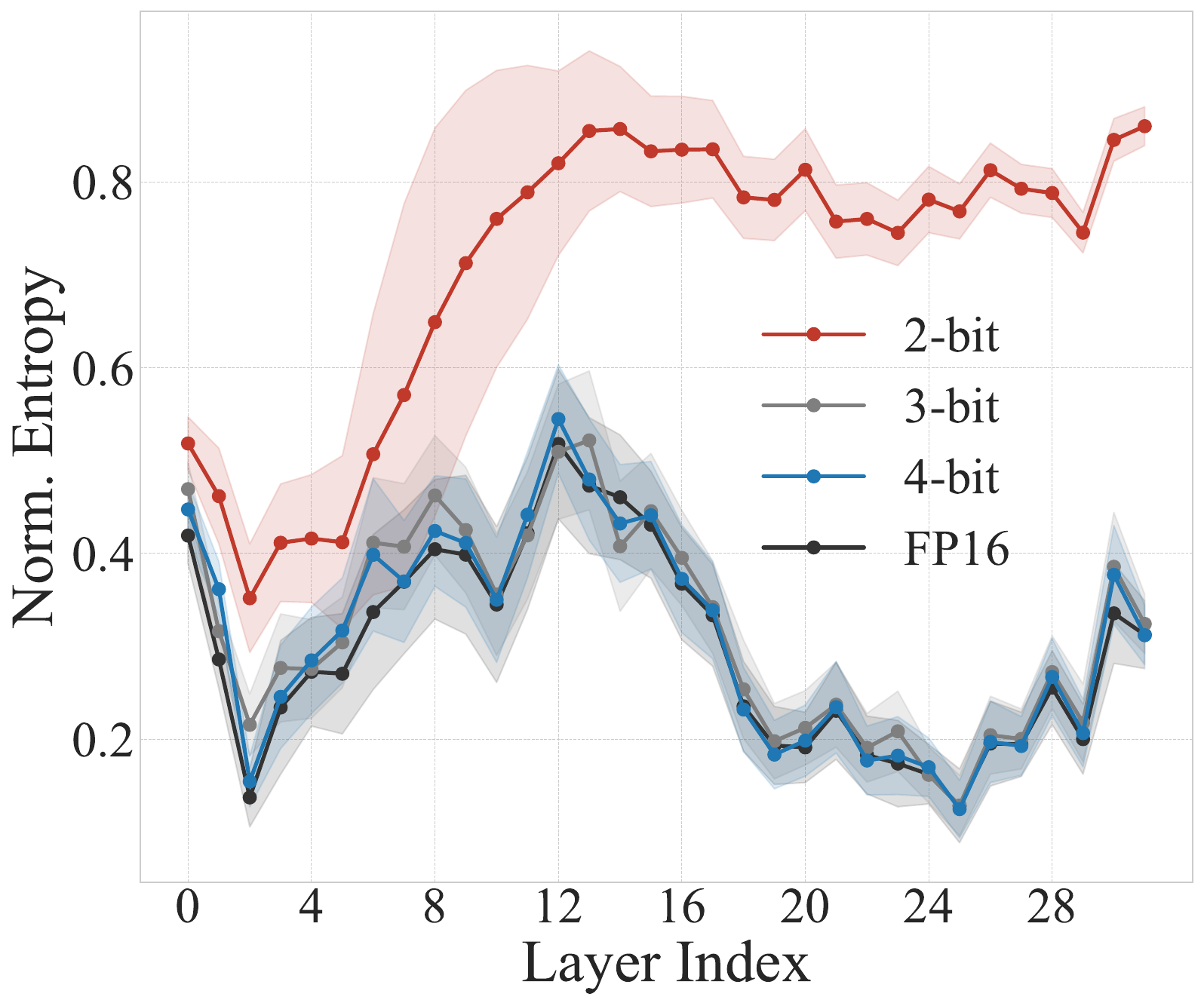}
        \caption{Normalized Attention Entropy}
    \end{subfigure}
    \hspace{0.02\textwidth}
    \begin{subfigure}[b]{0.43\textwidth}
        \centering
        \includegraphics[width=\textwidth]{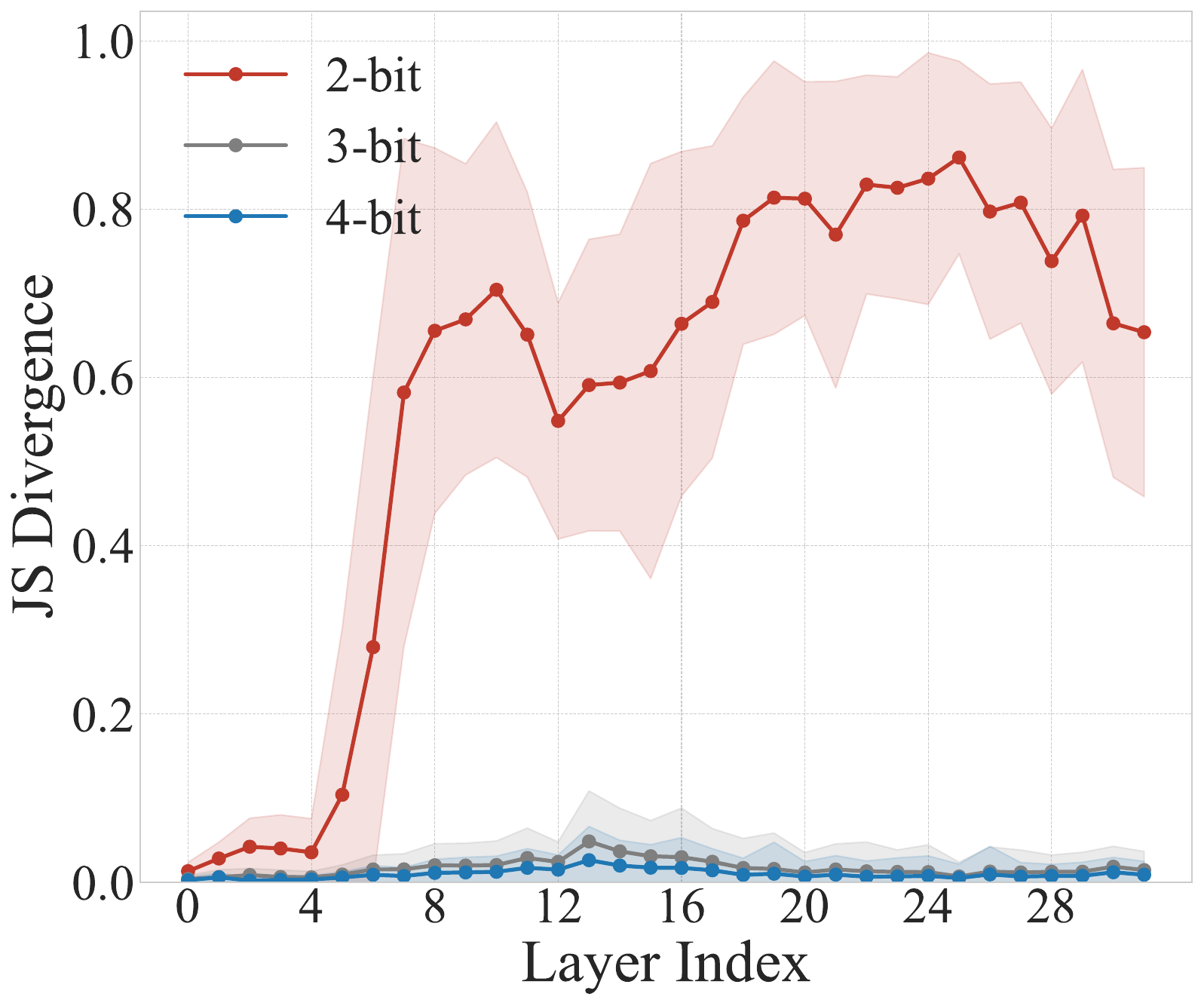}
        \caption{Focus Divergence (JSD)}
    \end{subfigure}
    \caption{Attention mechanism analysis at the last token for AWQ on the Failure Subset.}
    \label{fig:app_awq_attn}
\end{figure*}

\paragraph{Attention Mechanism.} Figure~\ref{fig:app_awq_attn} illustrates the attention patterns. While 4-bit and 3-bit models show tight alignment with the baseline, 2-bit quantization triggers a severe concentration collapse. Its normalized attention entropy exceeds 0.80 in middle-to-late layers, and its focus divergence sharply increases, indicating that the attention mechanism loses its routing capability.

\begin{figure*}[!t]
    \centering
    \begin{subfigure}[b]{0.32\textwidth}
        \centering
        \includegraphics[width=\textwidth]{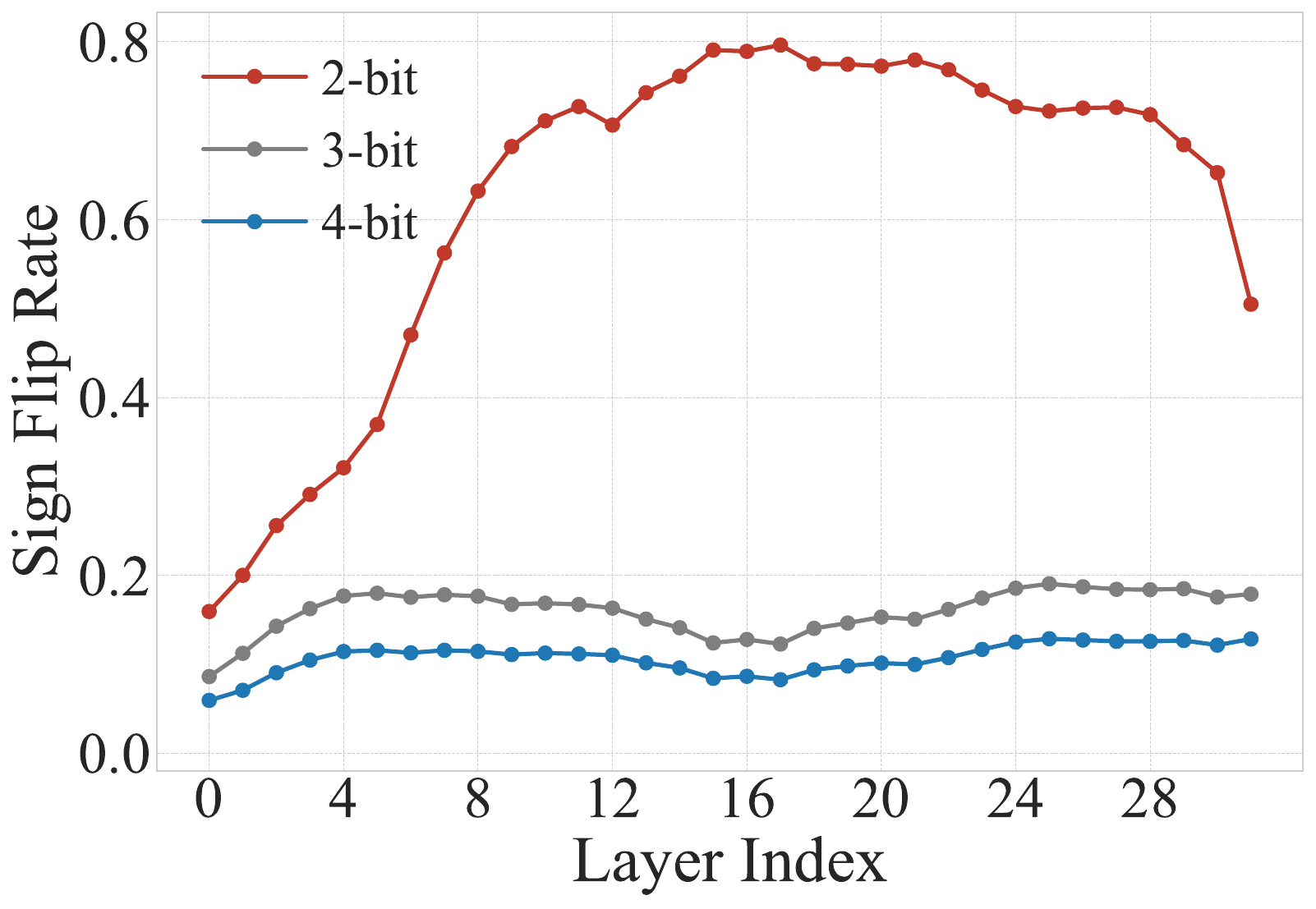}
        \caption{Gate Sign Flip Rate}
    \end{subfigure}
    \hfill
    \begin{subfigure}[b]{0.32\textwidth}
        \centering
        \includegraphics[width=\textwidth]{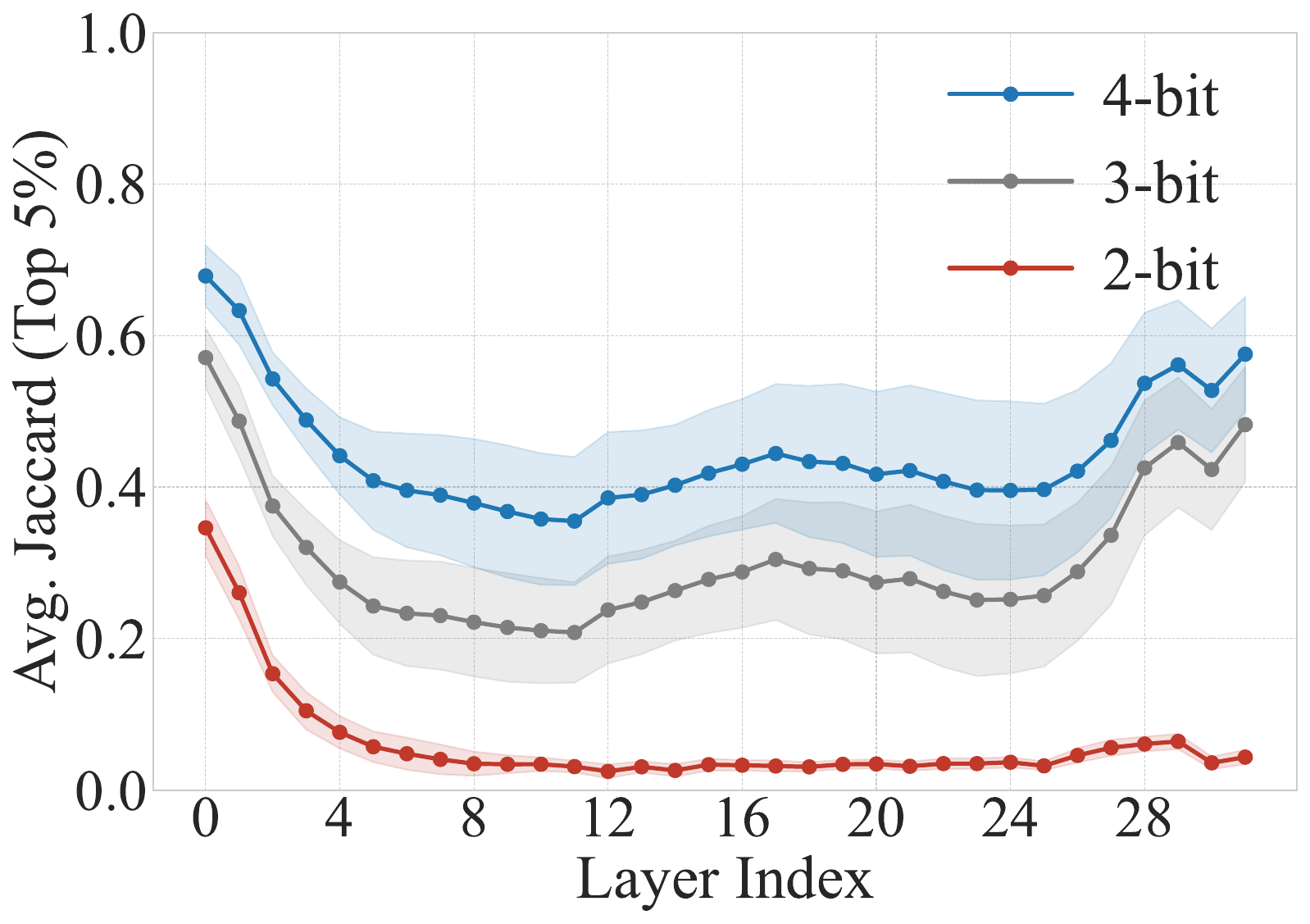}
        \caption{Expert Jaccard Similarity}
    \end{subfigure}
    \hfill
    \begin{subfigure}[b]{0.32\textwidth}
        \centering
        \includegraphics[width=\textwidth]{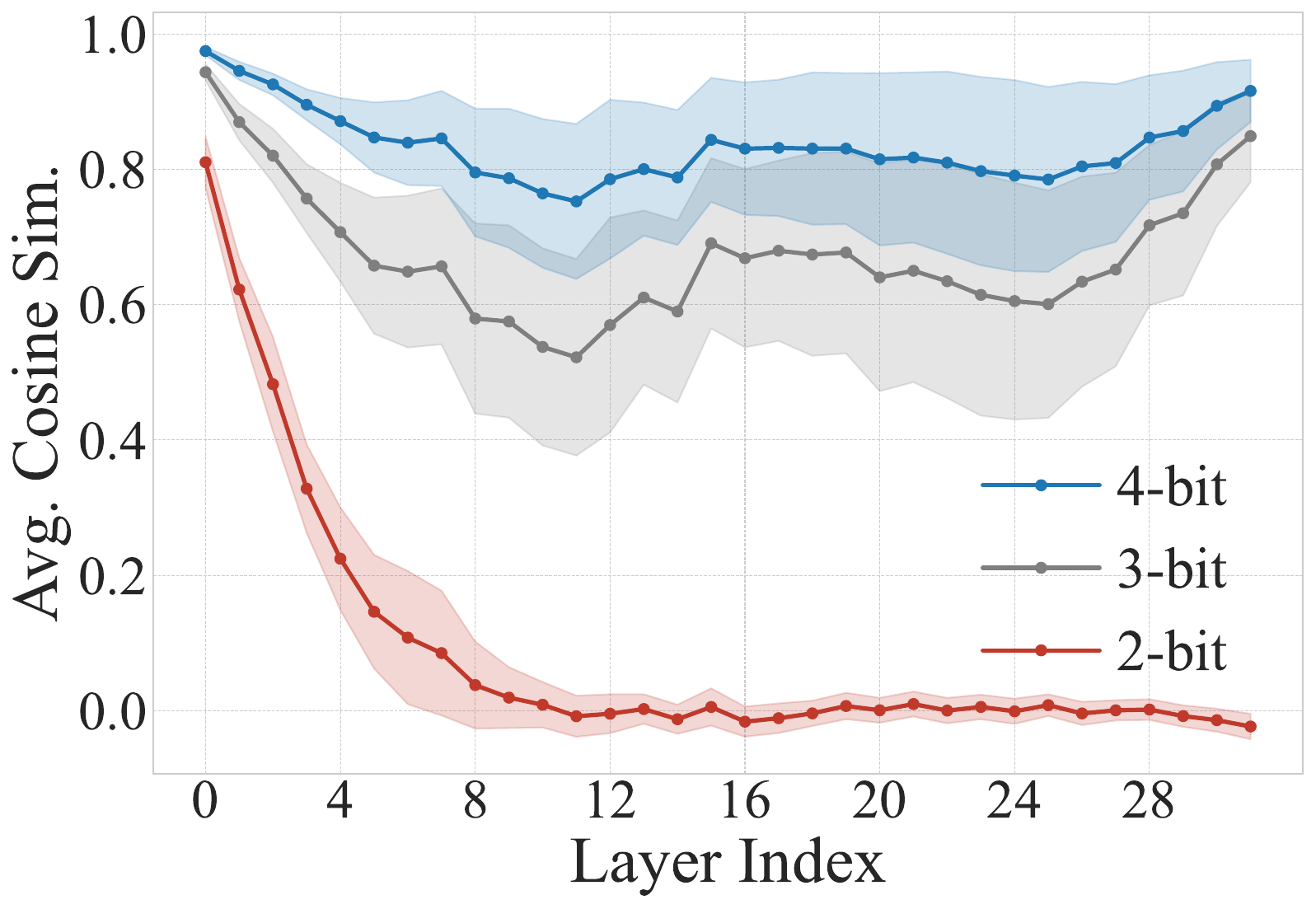}
        \caption{Value Similarity (Cosine)}
    \end{subfigure}
    \caption{Parallel indicators of FFN functionality at the last subject token for AWQ on the Failure Subset.}
    \label{fig:app_awq_ffn}
\end{figure*}

\paragraph{FFN Key-Value Memory.} Figure~\ref{fig:app_awq_ffn} presents the FFN functionality metrics. The 4-bit and 3-bit models maintain relatively stable gate flip rates and retrieve semantic values with high cosine similarity. However, the 2-bit model induces massive gate flipping, reaching nearly 80\% in middle layers. This severe disruption causes a rapid drop in expert Jaccard similarity and drives the semantic alignment of output values to near-zero.

Collectively, these results confirm that the transition from Signal Degradation to Computation Collapse is a fundamental pattern of quantization damage, rather than a GPTQ-specific artifact.

\section{Generalizability to Broader Language Tasks}
\label{sec:app_general_tasks}

\makeatletter
\setlength{\@dblfptop}{0pt}       
\setlength{\@dblfpsep}{20pt}      
\setlength{\@dblfpbot}{0pt plus 1fil}
\makeatother

To demonstrate that the discovered failure modes generalize beyond factual recall, we extend our mechanistic metrics to MMLU~\cite{hendrycks_measuring_2021} and GSM8K~\cite{cobbe_training_2021}. 

\subsection{Experimental Setup}
All evaluations are conducted in a 5-shot setup, with full-dataset accuracy reported in Table~\ref{tab:app_broad_acc}. For the mechanistic analysis, we use the complete GSM8K dataset alongside a representative MMLU subset of 1,066 samples across four diverse domains (macroeconomics, philosophy, clinical knowledge, and computer science).

\begin{table}[h]
\centering
\small
\begin{tabular}{lcccc}
\toprule
\textbf{Dataset} & \textbf{FP16} & \textbf{4-bit} & \textbf{3-bit} & \textbf{2-bit} \\
\midrule
MMLU & 63.42\% & 61.44\% & 49.11\% & 24.18\% \\
GSM8K & 56.79\% & 51.33\% & 10.16\% & 0.99\% \\
\bottomrule
\end{tabular}
\caption{Accuracy of Llama-3.1-8B on broader tasks across bit-widths.}
\label{tab:app_broad_acc}
\end{table}

\subsection{Semantic Subspace Integrity}

\begin{figure*}[!t]
    \centering
    \begin{subfigure}[b]{0.43\textwidth}
        \centering
        \includegraphics[width=\textwidth]{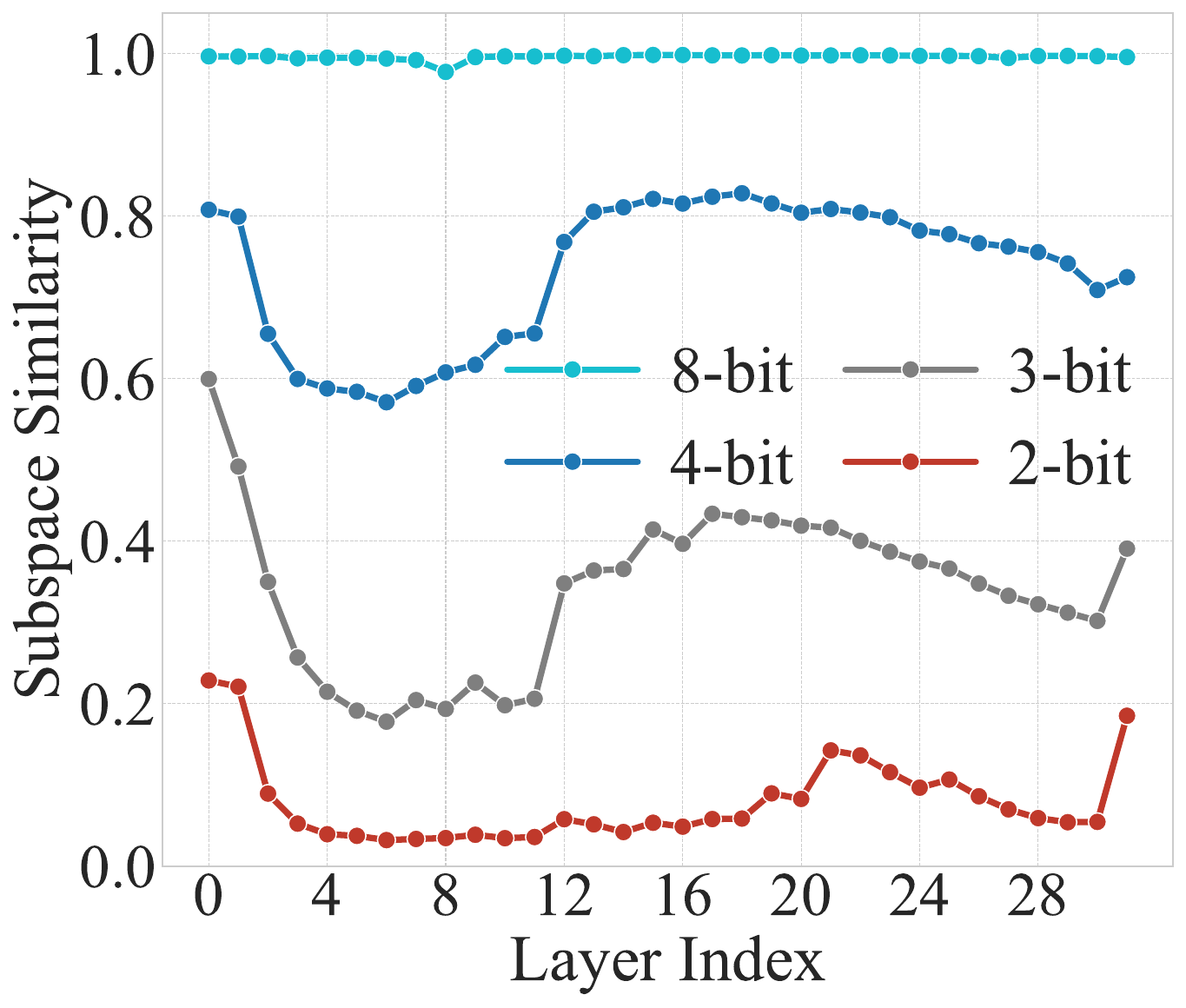}
        \caption{Subspace Similarity on MMLU}
    \end{subfigure}
    \hspace{0.02\textwidth}
    \begin{subfigure}[b]{0.43\textwidth}
        \centering
        \includegraphics[width=\textwidth]{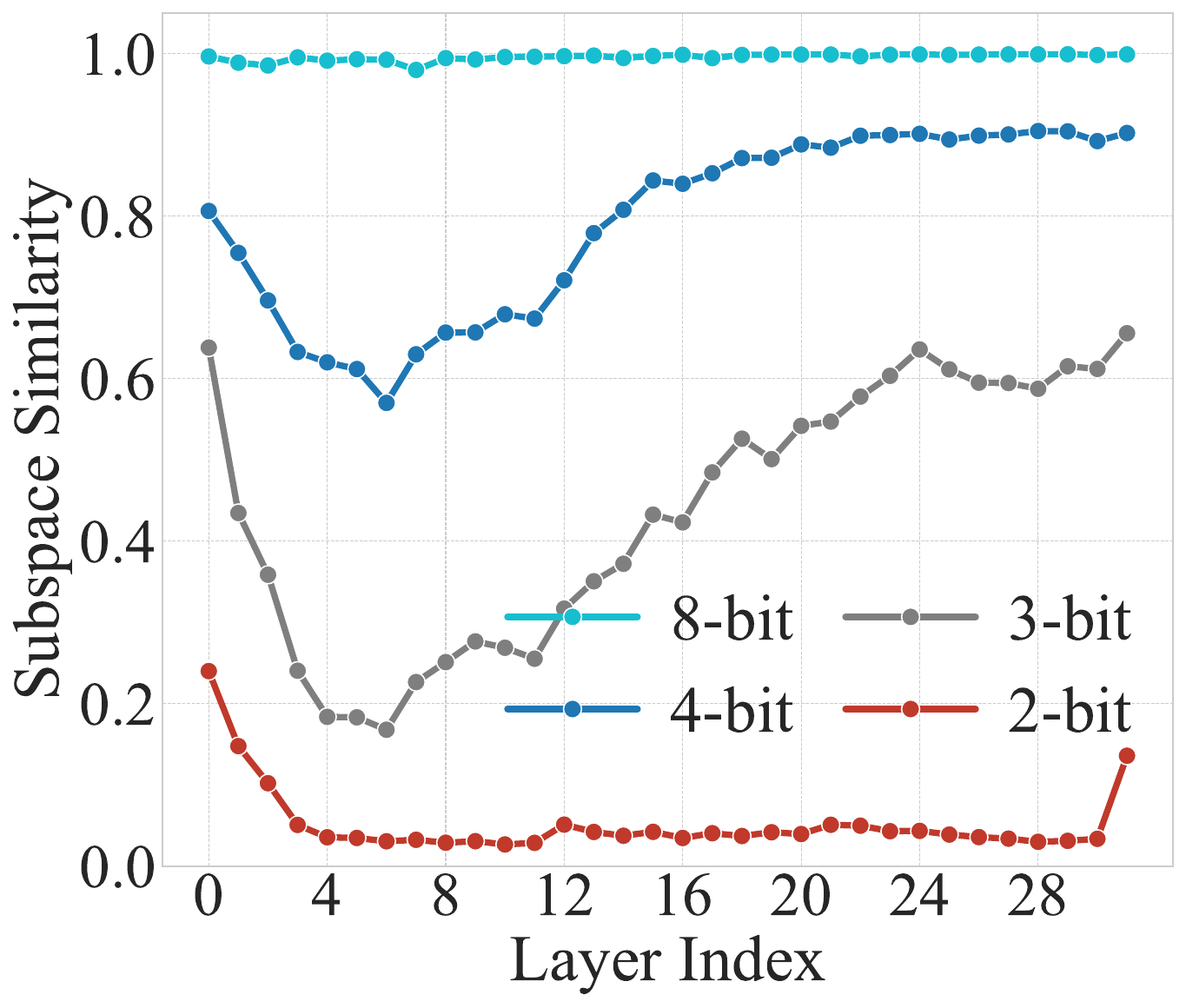}
        \caption{Subspace Similarity on GSM8K}
    \end{subfigure}
    \caption{Layer-wise SVD analysis (Top-50 dimensions) on broader tasks, calculated from a single forward pass.}
    \label{fig:app_broad_subspace}
\end{figure*}

We analyze the semantic subspace alignment using a single forward pass for both tasks. Figure~\ref{fig:app_broad_subspace} illustrates the layer-wise activation subspace similarity. Consistent with our findings on factual recall, the 2-bit trajectory plummets and remains near zero across all layers. In contrast, the 4-bit and 3-bit models initially experience a drop in similarity but subsequently recover and stabilize in deeper layers, confirming that their primary semantic directions are partially preserved despite precision loss.

\subsection{Attention and Generation Dynamics}

\begin{figure*}[!t]
    \centering
    \begin{subfigure}[b]{0.41\textwidth}
        \centering
        \includegraphics[width=\textwidth]{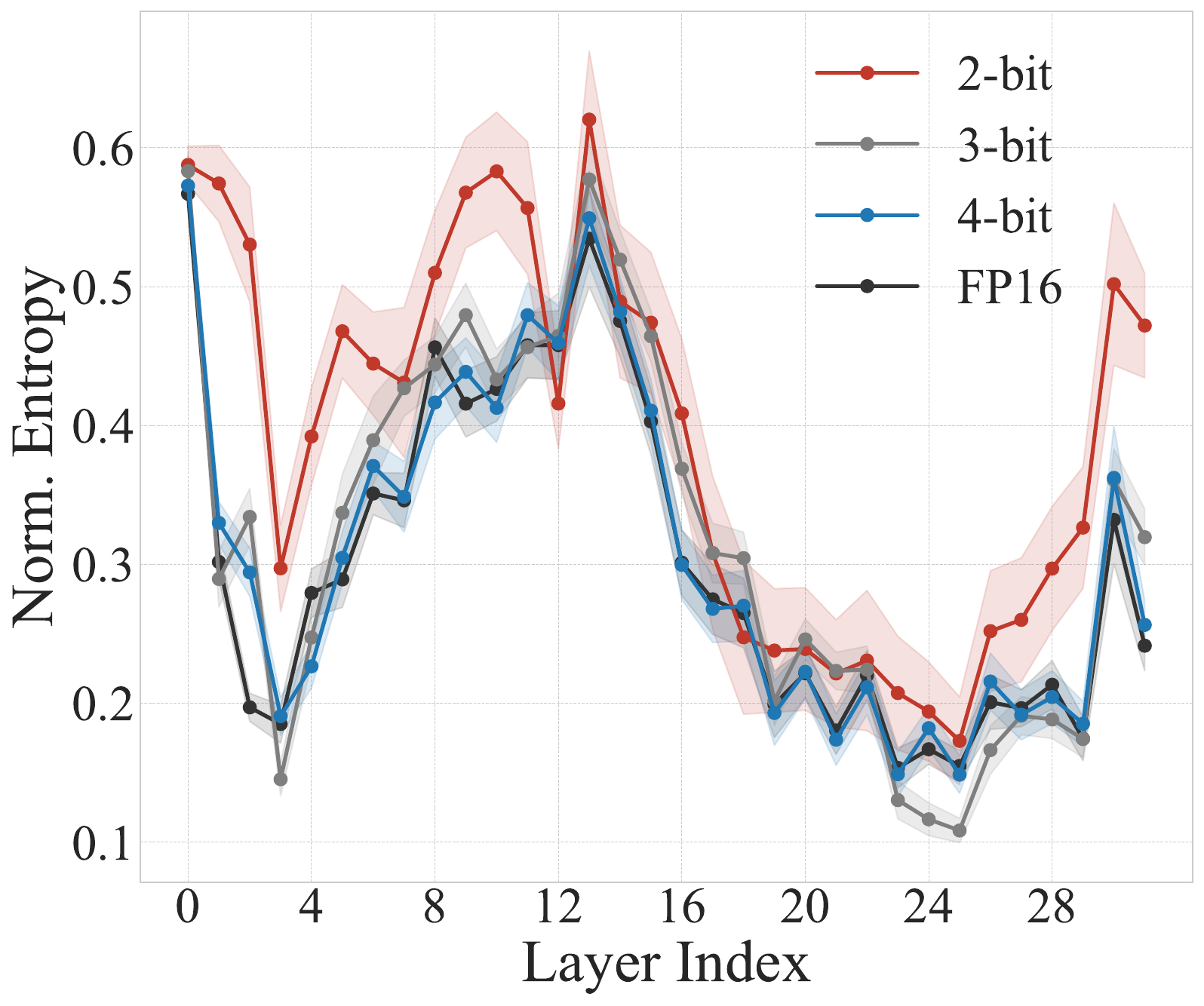}
        \caption{Layer-wise Entropy on MMLU}
    \end{subfigure}
    \hfill
    \begin{subfigure}[b]{0.55\textwidth}
        \centering
        \includegraphics[width=\textwidth]{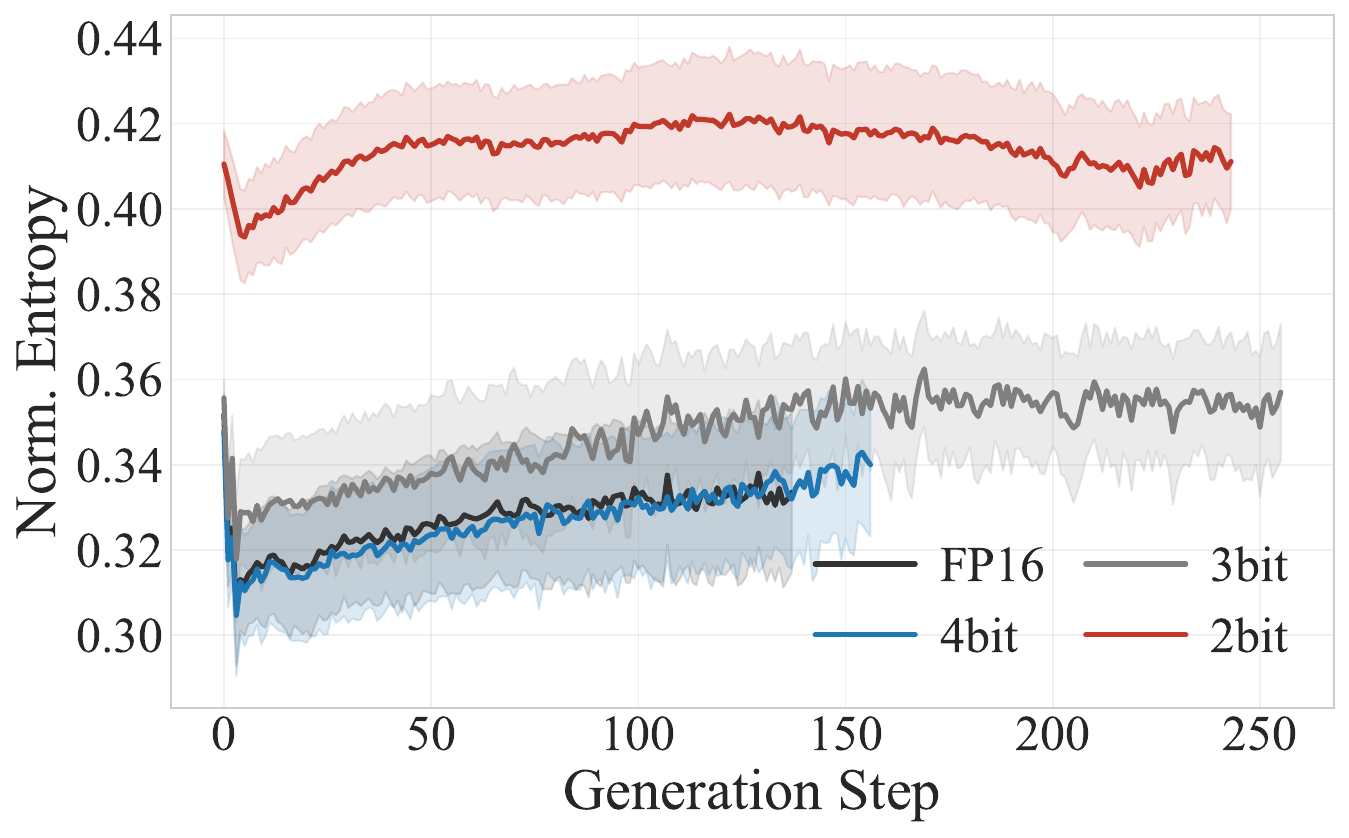}
        \caption{Temporal Entropy Dynamics on GSM8K}
    \end{subfigure}
    \caption{Attention entropy analysis. MMLU results are calculated from a single forward pass, while the GSM8K curve traces the layer-averaged entropy at each generation step (truncated when <10\% of samples remain active).}
    \label{fig:app_broad_entropy}
\end{figure*}

Given the crucial role of the attention mechanism in processing context during multi-step reasoning, attention entropy serves as an effective indicator for observing quantization-induced behavioral shifts. For MMLU (Fig.~\ref{fig:app_broad_entropy}a), the entropy is calculated from a single forward pass. For GSM8K (Fig.~\ref{fig:app_broad_entropy}b), we track the layer-averaged entropy of the last token at each generation step.

As shown in Figure~\ref{fig:app_broad_entropy}, the 2-bit model exhibits a severe deterioration of attention focus. On GSM8K, its attention entropy starts abnormally high at the beginning and persists throughout all steps, whereas the 4-bit entropy closely tracks the FP16 baseline. Because of this persistent high-entropy state, the 2-bit model fails to execute fine-grained reasoning. Qualitative inspection of failure cases reveals it generates chaotic content (e.g., meaningless numbers and repetitive loops), typically failing to halt until hitting the maximum generation length (with median generated tokens doubling from 77 in FP16 to 151 in 2-bit).

\end{document}